%% file: main.tex
\documentclass[10pt,twocolumn,letterpaper]{article}

\usepackage{times}
\usepackage{cite}
\usepackage{xspace}
\usepackage{etoolbox}
\usepackage[format=plain,labelformat=simple,labelsep=period,font=small,compatibility=false]{caption}
\usepackage[font=footnotesize,skip=3pt,subrefformat=parens]{subcaption}
\usepackage[hyphens]{url}
\usepackage{keyval}
\font\cvprtenhv  = phvb at 8pt %
\font\elvbf  = ptmb scaled 1100
\makeatletter
\def\@maketitle
   {
   \newpage
   \null
   \vskip .375in
   \begin{center}
      {{\Large \bf \@title \par}}
      {\vspace*{24pt}}
      {
      \large
      \lineskip .5em
      \begin{tabular}[t]{c}
        \@author
      \end{tabular}
      \par
      }
      \vskip .5em
      \vspace*{12pt}
   \end{center}
   }
\makeatother

\def\abstract
   {%
   \centerline{\large\bf Abstract}%
   \vspace*{12pt}%
   \it%
   }

\def\affiliation#1{\gdef\@affiliation{#1}} \gdef\@affiliation{}

\makeatletter
\DeclareRobustCommand\onedot{\futurelet\@let@token\@onedot}
\def\@onedot{\ifx\@let@token.\else.\null\fi\xspace}

\makeatother

\makeatletter
\def\cvprsection{\@startsection {section}{1}{\z@}
   {10pt plus 2pt minus 2pt}{7pt} {\large\bf}}
\def\cvprssect#1{\cvprsection*{#1}}
\def\cvprsect#1{\cvprsection{\hskip -1em.~#1}}
\def\section{\@ifstar\cvprssect\cvprsect}

\def\cvprsubsection{\@startsection {subsection}{2}{\z@}
   {8pt plus 2pt minus 2pt}{6pt} {\elvbf}}
\def\cvprssubsect#1{\cvprsubsection*{#1}}
\def\cvprsubsect#1{\cvprsubsection{\hskip -1em.~#1}}
\def\subsection{\@ifstar\cvprssubsect\cvprsubsect}
\makeatother

\usepackage{graphicx}
\usepackage[accsupp]{axessibility}
\usepackage[percent]{overpic}
\usepackage{amsmath}
\usepackage{amssymb}
\usepackage{booktabs}
\usepackage{soul}
\usepackage[dvipsnames]{xcolor}
\usepackage{enumitem}
\usepackage{float}
\usepackage{pgfplots}
\usepackage{tikz}
\usetikzlibrary{positioning,shapes}
\usepackage{threeparttable}
\usepackage{tabularx}
\usepackage{appendix}
\usepackage{array,multirow}
\usepackage[normalem]{ulem}
\usepackage{xpatch}
\usepackage{placeins}
\usepackage{siunitx}
\usepackage[pagebackref,breaklinks,colorlinks]{hyperref}
\usepackage[capitalize]{cleveref}

\newcommand*{\R}{\mathbb{R}}
\newcommand*{\ua}{$\uparrow$}
\newcommand*{\da}{$\downarrow$}
\newcommand{\twostage}{{Two-stage PIXOR}}
\newcommand*{\bw}{1.5pt}
\newcommand*{\lw}{1.5pt}
\definecolor{MP3Color}{RGB}{145, 219, 87}
\definecolor{ourmodelcolor1}{RGB}{87, 111, 219}
\definecolor{OccFlowColor}{RGB}{219, 95, 87}
\def\object{object}
\def\Object{Object}
\def\ourmodel{\textsc{ImplicitO}}
\def\reverse{backwards}
\def\Reverse{Backwards}

\begin{document}

\makeatletter
\xpatchcmd{\paragraph}{3.25ex \@plus1ex \@minus.2ex}{1pt plus 1pt minus 1pt}{\typeout{success!}}{\typeout{failure!}}
\makeatother

\title{Implicit Occupancy Flow Fields for Perception and Prediction in Self-Driving\vspace{-0.5em}}

\author{
\textbf{Ben Agro\thanks{Denotes equal contribution} , Quinlan Sykora$^*$, Sergio Casas$^*$, Raquel Urtasun} \\
Waabi, University of Toronto \\
\texttt {\{bagro, qsykora, sergio, urtasun\}@waabi.ai}
\vspace{-0.5em}
}

\maketitle

\input{sections/abstract.tex}
\vspace{-1em}
\input{sections/intro.tex}
\input{sections/background.tex}

\begin{figure*}[t]
    \centering
    \includegraphics[width = \linewidth]{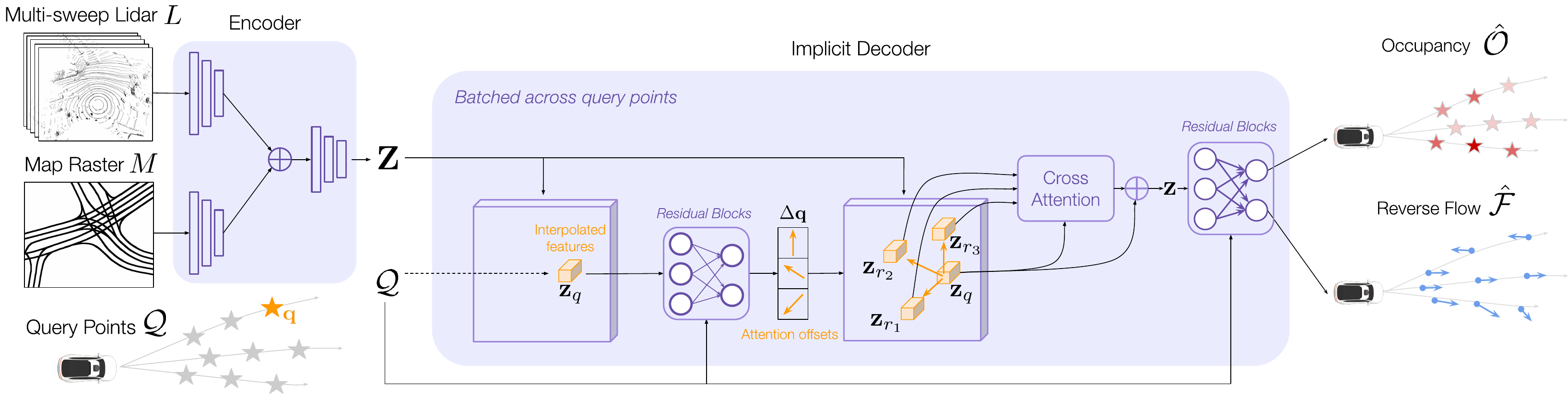}
    \caption{
      An overview of our model, \ourmodel{}. Voxelized LiDAR and
      an HD map raster are encoded by a two-stream CNN. 
      The resulting feature map $\mathbf{Z}$ is used by the decoder to obtain relevant features for the query points $\mathcal{Q}$ and eventually
      predict occupancy $\hat{\mathcal{O}}$ and reverse flow $\hat{\mathcal{F}}$.
      We illustrate the attention for {\color{YellowOrange} a single query point $\mathbf{q}$},
      but the inference is fully parallel across query points $\mathcal{Q}$.
    }
    \label{fig:architecture-diagram}
    \vspace{-1em}
\end{figure*}

\input{sections/method.tex}

\input{sections/experiments.tex}

\input{sections/conclusion.tex}

\vspace{-20pt}
\paragraph{Acknowledgements:}
We thank the Waabi team for their valuable assistance and support.

{\small
\bibliographystyle{ieee_fullname}
\bibliography{main.bib}
}

\clearpage

\renewcommand*{\w}{0.15\textwidth}

\section*{Appendix}

\appendix
In this appendix, we first describe implementation details that are relevant to \ourmodel{}, including the full computation graph of our implicit decoder architecture,
the encoder architecture, and training details.
Then, we provide implementation details for the baselines.
Last but not least, we showcase several additional results:
\begin{itemize}
  \item Introspection of $\ourmodel{}$ ($K = 4$) using visualizations of the attention offsets, 
  cross attention weights, and qualitative results.
  \item Analysis of occupancy metrics over time for both \object-based and \object-free prior art as well as \ourmodel{}.
  \item Analysis of how performance is affected by grid resolution for explicit \object-free methods.
  \item Insights into the occupancy probability calibration of different methods via reliability diagrams.
\end{itemize}

\input{sections/implementation_details.tex}
\input{sections/baseline_implementation_details.tex}
\input{sections/additional_results.tex}

\end{document}

%% file: sections/abstract.tex
\begin{abstract}
\vspace{-0.5em}
A self-driving vehicle (SDV) must be able to perceive its surroundings and predict the future behavior of other traffic participants.
Existing works either perform object detection followed by trajectory forecasting of the detected objects,
or predict dense occupancy and flow grids for the whole scene.
The former poses a safety concern as the number of detections needs to be kept low for efficiency reasons,
sacrificing object recall. The latter is computationally expensive due to the high-dimensionality of the output grid,
and suffers from the limited receptive field inherent to fully convolutional networks.
Furthermore, both approaches employ many computational resources predicting areas or objects that might never be queried by the motion planner.
This motivates our unified approach to perception and future prediction that implicitly represents occupancy and flow over time with a single neural network. Our method avoids unnecessary computation, as it can be directly queried by the motion planner at continuous spatio-temporal locations. Moreover, we design an architecture that overcomes the limited receptive field of previous explicit occupancy prediction methods by adding an efficient yet effective global attention mechanism. Through extensive experiments in both urban and highway settings, we demonstrate that our implicit model outperforms the current state-of-the-art. For more information, visit the project website: \href{https://waabi.ai/research/implicito}{https://waabi.ai/research/implicito}.

\end{abstract}

%% file: sections/intro.tex
\section{Introduction} %
\label{sec:introduction}

The goal of a self-driving vehicle is to take sensor observations of the environment
and offline evidence such as high-definition (HD) maps and execute a safe and comfortable plan towards its destination.  
Meanwhile, it is important to produce interpretable representations that explain why the vehicle performed a certain maneuver, particularly if a dangerous event were to occur.
To satisfy this, traditional autonomy stacks \cite{casas2018intentnet,Liang_2020_CVPR,weng2020ptp,weng20203d,salzmann2020trajectron,ivanovic2018the,ivanovic2020multimodal,gu2021densetnt,cui2021lookout} break down the problem into 3 tasks: perception, motion forecasting and motion planning.
Perception leverages sensor data to localize the traffic participants in the scene. 
Motion forecasting outputs the distribution of their future motion, which is typically multimodal. 
Finally, motion planning is tasked with deciding which maneuver the SDV should execute. 

\begin{figure}[t]
  \centering
  \vspace{-5pt}
  \includegraphics[width = \linewidth]{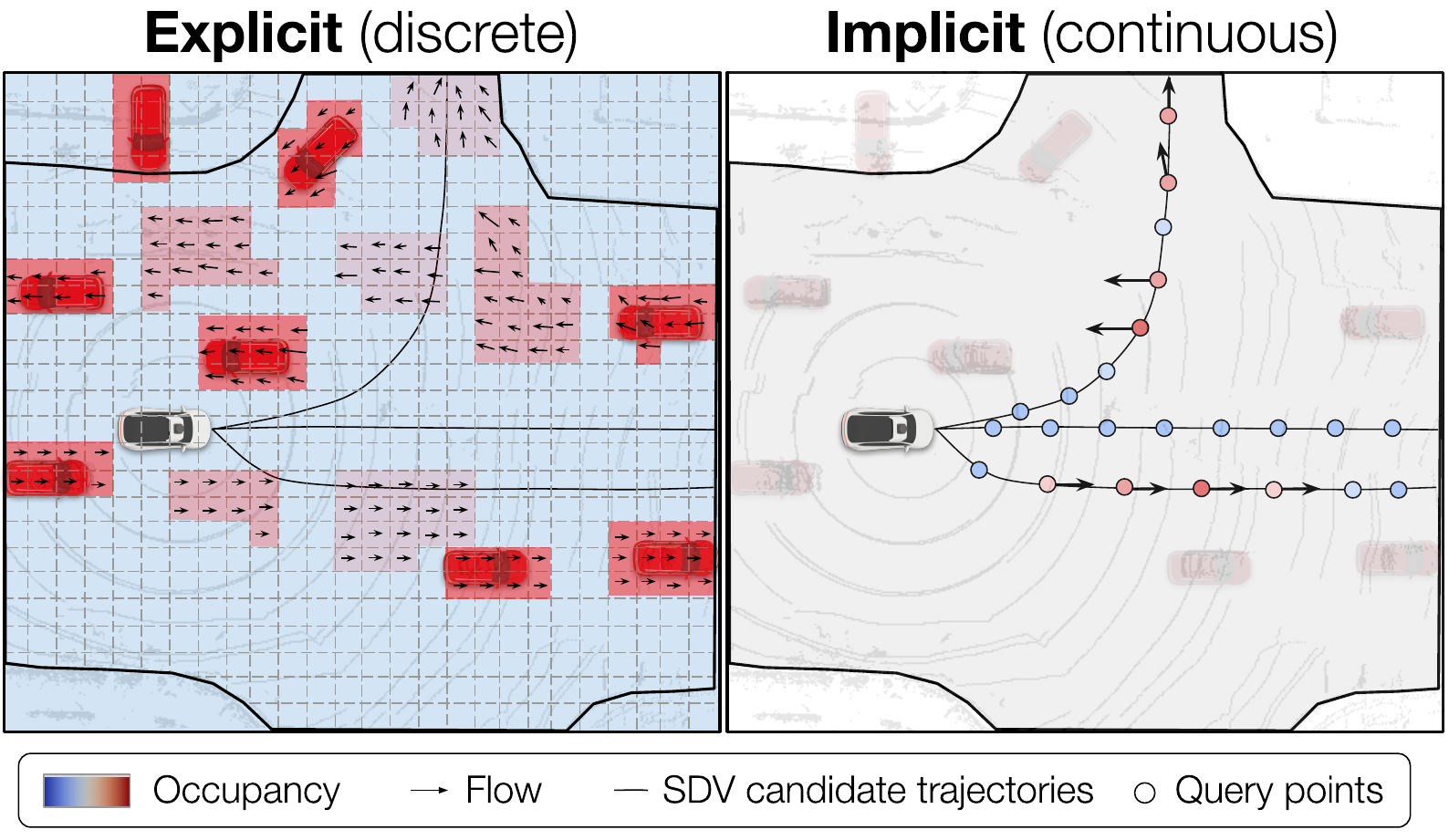}
  \vspace{-20pt}
  \caption{
  Left: Explicit approaches predict whole-scene occupancy and flow on a spatio-temporal grid. 
  Right: Our implicit approach only predicts occupancy and flow at queried continuous points, focusing on what matters for downstream planning.}
  \label{fig:hook}
  \vspace{-1em}
\end{figure}

Most autonomy systems  are \textit{\object-based}, which involves detecting the objects of interest in the scene.
To do so, object detectors threshold predicted confidence scores to determine which objects in the scene, a trade off between precision and recall.
Furthermore, object-based motion forecasting methods are limited to predict only a handful of sample trajectories or parametric distributions with closed-form likelihood for tractability, as they scale linearly with the number of objects and must run online in the vehicle.
This causes information loss that could result in unsafe situations \cite{sadat2020perceive}, e.g., if a solid object is below the detection threshold, or the future behavior of the object is not captured by the simplistic
future trajectory estimates.

In recent years, \textit{\object-free} approaches \cite{sadat2020perceive,philion2020lift,casas2021mp3,hu2021fiery} that model the presence, location and future behavior of all agents in the scene via a non-parametric distribution have emerged to address the shortcomings of \textit{\object-based} models. \Object-free approaches predict occupancy probability and motion for each cell in a spatio-temporal grid, directly from sensor data.
More concretely, the spatio-temporal grid is a 3-dimensional dense grid with two spatial dimensions representing
the bird's-eye view, and a temporal dimension from the current observation time to a future horizon of choice. All dimensions are quantized
at regular intervals.
In this paradigm, no detection confidence thresholding is required and the distribution over future motion is much more expressive, enabling the downstream motion planner to plan with consideration of low-probability objects and futures. 
Unfortunately, object-free approaches are computationally expensive as the grid must be very high-dimensional to mitigate quantization errors. 
However, most of the computation and memory employed in object-free methods is unnecessary, as motion planners only need to cost a set of spatio-temporal points  around candidate trajectories, and not a dense region of interest (RoI).  We refer the reader to \cref{fig:hook} for an illustration. 

This motivates our approach, \ourmodel{}, 
which utilizes an implicit representation to predict both occupancy probability and flow over time directly from raw sensor data and HD maps. 
This enables downstream tasks such as motion planning to efficiently evaluate a large collection of \textit{spatio-temporal query points} in parallel, focusing on areas of interest where there are potential interactions with the self-driving vehicle.  
We design an architecture that overcomes the limited receptive field of fully convolutional explicit architectures \cite{sadat2020perceive,philion2020lift,hu2021fiery,mahjourian2022occupancy} by adding an efficient yet effective global attention mechanism.
In particular, we leverage deformable convolutions \cite{dai2017deformable} and cross attention \cite{Vaswani2017Attention} to focus on a compact set of distant regions per query, giving the predictions a global context. This is useful as dynamic objects can move at very high speeds, particularly on the highway. 
For instance, when predicting in-lane occupancy 3 seconds into the future on a road where the speed limit is 30 m/s, the attention can look approximately 90 meters back along the lane to find the corresponding sensor evidence.
Extensive experiments in both urban and highway scenarios show that 
our object-free implicit approach outperforms the two prevalent paradigms in the literature on the task of occupancy-flow prediction: 
(i) object-based methods that first perform object detection to localize a finite set of objects in the scene, and then predict their future trajectory distribution 
(ii) object-free explicit methods that predict dense spatio-temporal grids of occupancy and motion.

%% file: sections/background.tex
\section{Related Work} %
\label{sec:related-work}

In this section we discuss traditional \textit{\object-based} perception and prediction as well as 
 \textit{\object-free} perception and prediction. We also outline literature in \textit{implicit
geometric reconstruction}, which inspired our approach.

\paragraph{\Object-based Perception and Motion Forecasting:} 
The majority of previous works have adopted
object-based reasoning with a 2-stage pipeline, where first object detection
\cite{yang2018pixor, lang2019pointpillars} and tracking \cite{weng20203d,
sharma2018beyond} are performed, followed by trajectory prediction from past tracks 
\cite{tang2019multiple, phan2020covernet, chai2019multipath,
zhao2019multi}. As there are multiple possible futures, these methods either
generate a fixed number of modes with probabilities and/or draw samples to
characterize the trajectory distribution.
This paradigm has three main issues
\cite{trabelsi2021drowned,sadat2020perceive}: (1) uncertainty is not propagated from
perception to downstream prediction, (2) the predicted future distributions must be overly compact in practice, as their size grows linearly with the number of objects, (3) thresholded decisions in perception make the
planner blind to undetected objects.
Several works \cite{luo2018faf,
casas2018intentnet, Liang_2020_CVPR, casas2020implicit} tackle (1) by
optimizing jointly through the multiple stages. However, (2) and (3) are
fundamentally hard to address in this object-based paradigm as it implies a finite set of
objects that can be large in crowded scenes. 
In contrast, our model is agnostic to the number of objects in the scene since it predicts the occupancy probability and flow vectors at desired spatio-temporal points.

\paragraph{\Object-Free Perception and Prediction:} 
These methods forecast future occupancy
and motion from sensor data such as LiDAR \cite{sadat2020perceive,casas2021mp3} and camera \cite{philion2020lift,hu2021fiery,hu2022st}, \textit{without considering individual
actors}. 
P3\cite{sadat2020perceive} first introduced temporal semantic occupancy grids as an interpretable intermediate representation for
motion planning. 
MP3 \cite{casas2021mp3} enhanced this representation by predicting an initial occupancy grid and warping it into the future with
a spatio-temporal grid of multi-modal flow predictions. 
Compared to fully convolutional architectures, this flow-warping increases the effective receptive field for occupancy prediction, and imposes prior on how occupancy can evolve over time.
However, forward flow struggles with dispersing occupancy over time when uncertainty increases, as pointed out in \cite{mahjourian2022occupancy}.
FIERY\cite{hu2021fiery} added instance reasoning to the \object-free paradigm  as a postprocessing, improving interpretability.
\textsc{OccFlow} \cite{mahjourian2022occupancy} introduced \textit{\reverse{} flow} 
as a representation that can capture multi-modal forward motions with just one flow
vector prediction per grid cell.
However, \textsc{OccFlow} isolates the occupancy and flow
\textit{forecasting} problem by assuming input features (e.g., position, velocity, extent etc.) from a detection and tracking module, instead of raw sensor data.

While our work belongs to the category of \object-free methods, our model only predicts occupancy-flow at select query points instead of outputting spatio-temporal occupancy-flow grids with fully convolutional networks.
We achieve this with an efficient and effective global attention mechanism. This makes the model more expressive
while improving efficiency by reducing the computation to only that which matters to downstream tasks.

\paragraph{Implicit Geometric Reconstruction:}
Geometric reconstruction refers to the task of predicting the 3D shape of an object given some incomplete representation of it, e.g., images, LiDAR, voxels.
Implicit neural geometric reconstruction methods \cite{mescheder2019occupancy,chen2019learning,Park2019deepsdf} have been shown to outperform explicit counterparts, which represent the 3D shape as a grid, set of points, voxels or a mesh. 
In contrast, implicit methods train a neural network to predict a continuous field that assigns a value to each point in 3D space, so that a shape can be extracted as an iso-surface.
More concretely, this network can predict non-linear binary
occupancy \cite{mescheder2019occupancy,chen2019learning} over 3D space $f(x): \R^3 \to [0, 1]$, or a signed distance function to the surface \cite{Park2019deepsdf}. 
Our work is motivated by these ideas, and we explore their application to the
task of occupancy and flow prediction for self-driving. Particularly, the architecture of our
implicit prediction decoder is inspired by Convolutional Occupancy Networks
\cite{peng2020convolutional}, which proposed a translation equivariant approach to accurately
reconstruct large scale scenes.

%% file: sections/method.tex
\section{Implicit Perception and Future Prediction} 
\label{sec:method}

Understanding the temporal occupancy and motion of traffic participants in the scene is critical for motion planning, allowing the self-driving vehicle (SDV) to avoid collisions, maintain safety buffers and keep a safe headway \cite{casas2021mp3}. 
Previous methods \cite{sadat2020perceive,casas2021mp3,hu2021fiery,mahjourian2022occupancy, kim2022stopnet} represent occupancy and motion in bird's-eye view (BEV) \textit{explicitly}
with a discrete spatio-temporal grid. This approach is resource inefficient, because it uses computational resources to predict in regions that are irrelevant to the SDV.
In this section, we introduce \ourmodel, an \textit{implicit} neural network that can be queried for both scene occupancy and motion at any 3-dimensional continuous point ($x$, $y$, $t$).
Here, $x$ and $y$ are spatial coordinates in BEV and $t = \bar{t} + \Delta t$
is the time into the future, where $\bar{t}$ refers to the current timestep at which we are making the predictions, and $\Delta t \geq 0$ is an offset from the current timestep into the future.   
This enables the motion planner to request the computation only at points around the candidate trajectories that are being considered. 
In the remainder of this section, we first describe the task parametrization, then the network architecture, and finally how to train our approach.

\subsection{Task Parameterization}
\label{sec:problem}
We discuss the task by defining its inputs and outputs.

\paragraph{Input parameterization:} Our model takes as input a voxelized LiDAR representation ($L$) as well as a raster of the HD map ($M$).
For the LiDAR, let $\mathcal{S}_{\bar{t}} =  \{\mathbf{s}_{\bar{t}-T_{history}+1}, \dots, \mathbf{s}_{\bar{t}}\}$ 
be the sequence of the most recent $T_{history} = 5$ sweeps. 
More precisely, $\mathbf{s}_{t'} \in \R^{P_{t'} \times 3}$  is the LiDAR sweep ending at timestep $t'$ 
containing a set of $P_{t'}$ points, each of which described by three features: $(p_x, p_y, p_h)$. $p_x$ and $p_y$ are the location of the point relative to the SDV's reference frame at the current timestep $\bar{t}$ --- centered at the SDV's current position and with the $x$-axis pointing along the direction of the its heading.
$p_h$ corresponds to the height of the point above the ground. 
Finally, $L=Voxelize(\mathcal{S}_t) \in \mathbb{R}^{T_{history} D \times H \times W}$, where we follow the multi-sweep BEV voxelization proposed in \cite{yang2018hdnet} with a discretization of $D$ depth channels normal to the BEV plane, $H$ height pixels and $W$ width pixels.
For the raster map, we take the lane centerlines $\mathcal{C}$ represented as polylines from the high-definition map and rasterize them on a single channel $M = Raster(\mathcal{C}) \in \mathbb{R}^{1 \times H \times W}$ with the same spatial dimensions.

\paragraph{Output parameterization:} 
Let $\mathbf{q}= (x, y, t) \in \mathbb{R}^3$ be a spatio-temporal point in BEV, at a future time $t$. 
The task is to predict the probability of occupancy $o: \mathbb{R}^3 \to [0, 1],$ and the flow vector $\mathbf{f}: \mathbb{R}^3 \to \R^2$
specifying the BEV motion of any agent that occupies that location.
We model the \textit{\reverse{} flow} \cite{mahjourian2022occupancy} for the flow vector $\mathbf{f}$, as it can capture multi-modal forward motions with a single reverse flow vector per grid cell.
More concretely, \reverse{} flow describes the motion at time $t$ and location $(x,y)$ as the translation vector at that location from $t-1$ to $t$, should there be an object occupying it:
\begin{align}
  &\mathbf{f}(x, y, t) = (x', y')_{t-1} - (x, y)_{t},
\end{align}
where $(x', y')$ denotes the BEV location at time $t-1$ of the point occupying  $(x, y)$ at time $t$.

\subsection{Network Architecture}
\label{sec:Architecture}

We parameterize the predicted occupancy $\hat{o}$ and flow $\mathbf{\hat{f}}$ with
a multi-head neural network $\psi$. This network takes as input the voxelized LiDAR $L$, raster map $M$, 
and a mini-batch $\mathcal{Q}$ containing $|\mathcal{Q}|$ spatio-temporal query points $\mathbf{q}$,
and estimates the occupancy $\hat{\mathcal{O}}=\{{\hat{o}}(\mathbf{q})\}_{\mathbf{q} \in \mathcal{Q}}$ and flow $\hat{\mathcal{F}}=\{\mathbf{\hat{\mathbf{f}}}(\mathbf{q})\}_{\mathbf{q} \in \mathcal{Q}}$ for the mini-batch in parallel:
\begin{align}
    \hat{\mathcal{O}}, \hat{\mathcal{F}} = \psi(L, M, \mathcal{Q})
\end{align}
The network $\psi$ is divided into a convolutional encoder that computes scene features, and an implicit decoder that outputs the occupancy-flow estimates, as shown in \cref{fig:architecture-diagram}.

Inspired by \cite{yang2018pixor}, %
our encoder consists of two convolutional stems that process the BEV LiDAR
and map raster, a ResNet \cite{resnet} that takes the
concatenation of the LiDAR and map raster features and outputs multi-resolution
feature planes, and a lightweight Feature Pyramid Network (FPN) \cite{lin2017feature}
that processes these feature planes. This results in a BEV feature map at half
the resolution of the inputs, i.e., 
$\mathbf{Z} \in \R^{C \times \frac{H}{2} \times \frac{W}{2}}$,
that contains contextual features capturing the geometry, semantics, and motion of the scene.
It is worth noting that every spatial location (feature vector) in the feature map $\mathbf{Z}$ contains spatial information
about its neighborhood (the size of the receptive field of the encoder), as well as temporal information over the past
$T_{history}$ seconds. 
In other words, each feature vector in $\mathbf{Z}$ may contain important cues regarding the motion, the local road geometry, and neighboring agents.

We design an \textit{implicit occupancy and flow decoder} motivated by the  intuition that 
the occupancy at query point $\mathbf{q} = (x, y, t) \in \mathcal{Q}$ might be caused by a distant object moving at a fast speed prior to time $t$. 
Thus, we would like to use the local features around the spatio-temporal query location to suggest where to look next. 
For instance, there might be more expressive features about an object around its original position (at times $\{(\bar{t} - T_{history} + 1), \dots, \bar{t}\}$) since that is where the LiDAR evidence is. 
Neighboring traffic participants that might interact with the object occupying the query point at time $t$ are also relevant to look for (e.g., lead vehicle, another vehicle arriving at a merging point at a similar time). 

To implement these intuitions, we first bi-linearly interpolate the feature map
$\mathbf{Z}$ at the query BEV location $\mathbf{q}_{x,y} = (x, y)$ to obtain the feature vector
$\mathbf{z}_q = Interp(\mathbf{Z}, x, y) \in \R^C$ that contains local information around the query.
We then predict $K$ reference points $\{\mathbf{r}_1, \dots, \mathbf{r}_K\}$ by offseting the initial query point $\mathbf{r}_k = \mathbf{q} + \Delta \mathbf{q}_k$, where 
the offsets $\Delta \mathbf{q}$ 
are computed by employing 
the fully connected
ResNet-based architecture proposed by Convolutional Occupancy Networks
\cite{peng2020convolutional}. 
For all offsets we then obtain the corresponding features $\mathbf{z}_{r_k} = Interp(\mathbf{Z}, \mathbf{r}_k)$. 
This can be seen as a form of deformable convolution \cite{dai2017deformable};
a layer that predicts and adds 2D offsets to the regular grid
sampling locations of a convolution, and bi-linearly interpolates for feature vectors
at those offset locations. 
To aggregate the information from the deformed sample locations, we use cross attention between learned linear projections of
$\mathbf{z}_q \in \mathbb{R}^{1 \times C}$ and
$\mathbf{Z}_r=\{\mathbf{z}_{r_1}, \dots, \mathbf{z}_{r_k}\} \in \mathbb{R}^{K \times C}$. 
The result is the aggregated feature vector $\mathbf{z}$. See \cref{fig:architecture-diagram} for a visualization of 
this feature aggregation procedure.
Finally, $\mathbf{z}$ and $\mathbf{z}_q$ are concatenated, which, along with $\mathbf{q}$, is processed by
another fully connected ResNet-based architecture with two linear layer heads to predict
occupancy logits and flow.
Please see additional implementation details and a full computational graph of our model in the supplementary.

\subsection{Training}
\label{sec:training}

We train our implicit network by minimizing  a linear combination of an occupancy loss and a flow loss
\begin{align}
  \mathcal{L} = \mathcal{L}_o + \lambda_{\mathbf{f}} \mathcal{L}_{\mathbf{f}}.
\end{align}
Occupancy is supervised with binary cross entropy loss
$\mathcal{H}$ between the predicted and the ground truth occupancy at each query point $\mathbf{q} \in \mathcal{Q}$, %
\begin{align} %
  \mathcal{L}_o = \frac{1}{|\mathcal{Q}|}  \sum_{\mathbf{q} \in \mathcal{Q}} \mathcal{H}(o(\mathbf{q}), \hat{o}(\mathbf{q})),
\end{align}
where $o(\mathbf{q})$ and $\hat{o}(\mathbf{q})$ are ground truth and predicted occupancy and query point $\mathbf{q}$,
respectively. The ground truth labels are generated by directly calculating whether or 
not the query point lies within one of the bounding boxes in the scene.
We supervised the flow only for query points that belong to foreground, i.e., points that are occupied. By doing so, the model learns to predict the motion of a query location should it be occupied.
We use the  $\ell_2$ error, where
the labels are \reverse{} flow targets from $t$ to $t-1$ computed as rigid transformations between consecutive object box annotations: 
\begin{align}
  \mathcal{L}_f = \frac{1}{|\mathcal{Q}|} \sum_{\mathbf{q} \in \mathcal{Q}} o(\mathbf{q}) || \mathbf{f}(\mathbf{q}) - \mathbf{\hat{f}}(\mathbf{q}) ||_2. \label{eq:motion-loss}
\end{align}

We train with a batch of continuous query points $\mathcal{Q}$, as opposed to points on a regular grid as previously proposed.
More concretely, for each example we sample $|\mathcal{Q}|$ 
query points uniformly across the spatio-temporal volume $[0, H] \times [0, W] \times [0, T]$, where $H \in \mathbb{R}$ and $W \in \mathbb{R}$ are the height and width of a rectangular region of interest (RoI) in BEV surrounding the SDV,
and $T \in \mathbb{R}$ is the future horizon we are forecasting.

%% file: sections/experiments.tex
\section{Experiments} 
\label{sec:experiments}

In this section, we introduce the datasets and metrics used to benchmark occupancy-flow perception and prediction, 
and show that \ourmodel{} outperforms the state-of-the-art in both urban and highway settings. Further, we conduct two ablations studies to understand the effect of our contributions to the decoder architecture, and an analysis of the inference time of our implicit decoder compared to explicit alternatives.

\paragraph{Datasets:}
\label{sec:datasets}

We conduct our experiments using two datasets: Argoverse 2 Sensor \cite{Argoverse2} (urban), and HighwaySim (highway).
The Argoverse 2 Sensor (AV2) dataset is collected in U.S. cities and 
consists of 850 fifteen-second sequences 
with sensor data from two 32-beam LiDARs at a frequency of 10 Hz, high-definition maps with lane-graph and ground-height data, and bounding box annotations. We split the set into 700 sequences for training and 150 for validation, and break the sequences into examples that include 5 frames of LiDAR history and a prediction time horizon of 5 seconds. 
In our experiments, we only consider the occupancy and flow of vehicles, which we define
as the union of the following AV2 annotation classes: regular vehicle, large vehicle, wheeled device, box truck, truck, vehicular trailer, truck cab, school bus, articulated bus, message-board trailer and railed vehicle.
Query points are labeled with occupancy by checking if they intersect with the annotated bounding boxes.
Occupied query points are labeled with flow vectors using finite differences between the current query point and where that point was in the previous frame.
Incomplete tracks caused by missing annotations were filled-in using the constant turn rate and acceleration (CTRA) motion model\cite{CTRA},
so the models learn the prior that occupancy is persistent.
HighwaySim (HwySim) is a dataset generated with a state-of-the-art simulator, containing realistic highway traffic scenarios including on-ramps, off-ramps, and curved roads. %
A Pandar64 LiDAR is realistically simulated, and maps with lane-graph and ground-height are provided. 
700 sequences of around 15 seconds each are split 80/20 into training/validation.
Sequences are cut into examples, each with a history of 5 past LiDAR frames and
a 5 s future horizon.

\input{figures/main_table.tex}

\paragraph{Metrics:}
\label{sec:metrics}
To be fair with the baselines, we evaluate all models
with query points on a regular spatio-temporal grid. 
Temporally, we evaluate a prediction horizon of 5 seconds with a resolution of 0.5 seconds for both datasets.
In AV2, we employ a rectangular RoI of 80 by 80 meters centered around the SDV position at time $\bar{t}$ with a spatial grid resolution of 0.2 m.
In HwySim, we use an asymmetric ROI with 200 meters ahead and 40 meters back and to the sides of the SDV at time $\bar{t}$ with a grid resolution of 0.4 m. This is to evaluate on highway vehicles moving fast (up to 30 m/s) in the direction of the SDV over the full prediction horizon.
For simplicity, we refer to the grid cell centroids as ``query-points''. %
We evaluate the ability of the models to recover the ground-truth occupancy-flow. In particular, we utilize metrics to measure the precision, recall, accuracy and calibration of the occupancy, the flow errors, and the consistency between the occupancy and flow predictions.

\textit{Mean average precision (mAP):} 
mAP captures if the model correctly predicts a higher occupancy probability in occupied regions 
relative to unoccupied regions, i.e., an accurate ranking of occupancy probability. mAP is computed as the area
under the precision recall curve averaged across all timesteps in the prediction horizon.

\textit{Soft-IoU:} We follow prior works \cite{mahjourian2022occupancy,waymo_open_dataset,kim2022stopnet} in the
use of \textit{soft intersection over union} for assessing occupancy predictions:
\begin{align}
  \text{Soft-IoU} =  \frac{\sum_{\mathbf{q} \in \mathcal{Q}} o(\mathbf{q}) \hat{o}(\mathbf{q})}{\sum_{\mathbf{q} \in \mathcal{Q}} (o(\mathbf{q}) + \hat{o}(\mathbf{q}) - o(\mathbf{q}) \hat{o}(\mathbf{q}))}.
\end{align}
Unlike mAP, Soft-IoU also considers the magnitude of predicted occupancy probability instead of just the predicted probability ranking. 

\textit{Expected Calibration Error (ECE):}
ECE measures the expected difference between model confidence and accuracy.
This is desirable because the occupancy outputs may be used by downstream planners in a probabilistic way ---e.g., to compute the expected collision cost \cite{casas2021mp3}. Thus, we need to understand if the predicted probabilities are poorly calibrated, i.e., suffering from over-confidence or under-confidence \cite{guo2017on,luo2021safety}.

\textit{Foreground mean end-point-error (EPE):} This metric measures the average L2 flow error at
each occupied query point:
\begin{align}
    \text{EPE} = \frac{1}{\sum_{\mathbf{q} \in \mathcal{Q}} o(\mathbf{q})} \sum_{\mathbf{q} \in \mathcal{Q}} o(\mathbf{q})|| \mathbf{f}(\mathbf{q}) - \hat{\mathbf{f}}(\mathbf{q}) ||_2.
\end{align}

\textit{Flow Grounded Metrics:} 
Let $O_t$, $\hat{O}_t$, and $\hat{F}_t$ denote the occupancy labels, predicted occupancy, and predicted flow on a 
spatio-temporal grid at time $t$, respectively.
The flow grounded occupancy grid at timestep $t > \bar{t}$, 
is obtained by warping the ground truth occupancy grid at the previous timestep $O_{t - 1}$ 
with the predicted flow field $\hat{F}_{t}$, and multiplying it element-wise with the predicted occupancy at the current timestep $\hat{O}_{t}$ \cite{waymo_open_dataset}.
We report flow-grounded Soft-IoU and mAP by comparing the flow-grounded occupancy to the
occupancy ground truth.
The flow grounded metrics are useful for evaluating the consistency between the occupancy and flow predictions, as you can only achieve a high score if (1) the flow is accurate and (2) the warped ground-truth occupancy aligns well with the predicted occupancy for the next time step. %

\textit{Inference Time:} When measuring inference time, all methods were implemented with vanilla PyTorch code 
(no custom kernels) and run on a single Nvidia GeForce GTX 1080 Ti. This metric is sensitive to 
implementation, hardware, and optimizations, and thus should not be compared across different works.

\input{figures/qualitative_comparison_v2.tex}
\input{figures/introspection.tex}

\paragraph{Baselines:} 
\label{sec:sota-comp}
We compare against five baselines that cover the different perception and prediction paradigms outlined in the \cref{sec:introduction}.
\textsc{MultiPath} \cite{chai2019multipath}, \textsc{LaneGCN} \cite{lanegcn},
and \textsc{GoRela} \cite{cui2022gorela} are \object-based trajectory prediction models.
Following \cite{luo2021safety,kim2022stopnet}, 
to evaluate these \object-based models on the task of occupancy-flow prediction,
we rasterize the trajectory predictions to generate occupancy and flow fields.
For occupancy, we rasterize the multi-modal trajectory predictions weighted by the mode probabilities.
For flow, we generate a multi-modal spatio-temporal flow field, where for each mode, a grid cell predicted to
be occupied by an \object{} contains the forward-flow rigid-transformations defined by the trajectory of that \object{}.
\textsc{OccFlow} uses the occupancy-flow prediction architecture and
\textit{flow-traced loss} from Mahjourian et al. \cite{mahjourian2022occupancy},
using input features from a pre-trained detection and tracking module.
More information on the
detector can be found in the supplementary.
MP3 \cite{casas2021mp3} is an end-to-end trained perception and prediction method that predicts multi-modal forward-flow vectors and associated
probabilities, and uses these to warp a predicted initial occupancy grid forward in time.
We compute EPE on the expected motion vector (the probability-weighted sum of
modes) when evaluating MP3.

\paragraph{Benchmark against state-of-the-art:} 
\cref{tab:hook-comparison} presents our results on AV2 and HwySim
against the state-of-the-art baselines described above. 
For this experiment, our model \ourmodel{} predicts $K = 1$ attention offset.
Our method outperforms all others across all metrics and both datasets, showing
the suitability of \ourmodel{} in both urban and highway settings.
\cref{fig:argo-qualitative} displays qualitative results of these models on AV2.
Notice that all the \object-based models generally under-perform relative to the \object-free approaches.
This is likely because these models are not optimized for occupancy-flow directly, rather they are
trained to predict accurate trajectories.
The qualitative results of \textsc{GoRela} in \cref{fig:argo-qualitative} show
that thresholding to produce instances can result in missed detections (Scene 4). Further,
the trajectory parameterization results in rasterized occupancy that is more often inconsistent with the map
(Scenes 1 and 2), or inconsistent due to apparent collisions with other actors (Scenes 1 and 3).
This agrees with the results from \cite{kim2022stopnet}, and reaffirms the utility of the \object-free parameterization.
Interestingly, on AV2, the \object-based approaches have a high Soft-IoU despite their inaccurate occupancy ranking.
We find this is because these models are overconfident (reflected in their high ECE), which is heavily rewarded
by Soft-IoU on the many ``easy'' examples in AV2 with stationary vehicles (in the evaluation set, 64.4\% of
actors are static within the prediction horizon). %
This is supported by the worse relative Soft-IoU of these \object-based models on HwySim, which has a much higher
proportion of dynamic actors.
Interestingly \textsc{MP3} outperforms \textsc{OccFlow}
in the joint perception and prediction setting, contrary to the results under perfect perception assumption reported by \cite{mahjourian2022occupancy}. We hypothesize this is because MP3 is trained 
end-to-end and does not have the intermediate \object-based detection representation. 
We can see in \cref{fig:argo-qualitative} that
\textsc{OccFlow} hallucinates occupancy at the initial timestep (Scenes 1 and 4), and misses the detection of 
a vehicle in Scene 4, both of which are artifacts of training with input from a pre-trained detection model. %

\paragraph{Flow and Attention visualization:}
\cref{fig:argo-introspection} plots the reverse flow vectors as well as the attention offsets on 
two of the scenes from \cref{fig:argo-qualitative} (the middle two rows).
The first observation is that the flow vectors and attention offsets rely very heavily on the map raster, as expected.
The second observation is that the direction of the backward flow vectors and the attention offsets are very heavily correlated. This shows that the model has learned to ``look backwards''
along the lanes to gather relevant features despite the offsets being unsupervised.
We hypothesize that \ourmodel{} outperforms the others because of its larger effective
receptive field.
\cref{fig:argo-qualitative} shows that \ourmodel{} maintains occupancy into the future
more accurately than MP3. We attribute this to the attention offsets being a more general and
expressive mechanism than MP3's forward flow warping. 
To illustrate this further, in the supplementary we plot occupancy metrics as a function of prediction time $\Delta t$.

\input{figures/decoder_ablation_table.tex}

\paragraph{Influence of the decoder architecture:}
\label{sec:decoder-ablation}

In this section, we compare various occupancy-flow decoders, all trained end-to-end from LiDAR input
with the same encoder architecture as \ourmodel{} (described in \cref{sec:Architecture}).
This allows us to isolate the effect of our implicit decoder architecture design. 
\textsc{ConvNet} implements the decoder from Mahjourian et al. \cite{mahjourian2022occupancy}, but
it takes as input a feature map from the encoder, instead of hand-crafted detection features.
\textsc{ConvNetFT} denotes this same decoder architecture trained with the auxiliary supervision of flow trace loss \cite{mahjourian2022occupancy}.
Note that MP3 and \ourmodel{} from \cref{tab:hook-comparison} already use this encoder and are trained end-to-end, 
so the same results are presented for this ablation study.
As shown in \cref{tab:decoder-comparison} our implicit decoder \ourmodel{} outperforms all
the other decoders across all metrics except for ECE, on both HwySim and AV2. 
Notice that \textsc{ConvNet} and \textsc{ConvNetFT} outperform
their detection-based counterpart \textsc{OccFlow} in \cref{tab:hook-comparison} by a significant margin.
This highlights the utility of end-to-end training in the
\object-free paradigm for occupancy-flow prediction. Evidently, thresholding to
produce detections and hand-crafted features limits the information available for occupancy-flow perception and prediction.
Again, we hypothesize that \ourmodel{} outperforms the others due to its offset mechanism 
increasing the effective receptive field. Even with the powerful encoder and flow warping mechanism,
\textsc{ConvNet} and \textsc{MP3} fail to match this. This is supported by the relatively close
performance of \textsc{ConvNet} to \ourmodel{} on AV2, but not HwySim. On HwySim most vehicles travel larger
fraction of the ROI, so a larger effective receptive field is necessary. On AV2 more vehicles
are stationary or move slowly, so a large receptive field is less important for occupancy-flow
prediction.

\input{figures/num_offset_ablation.tex}

\paragraph{Influence of the number of offsets ($K$):}
\label{sec:offset-ablation}
Based on the attention offset visualizations in \cref{fig:argo-introspection}, we have
conjectured that the predicted attention offsets of our implicit decoder are
responsible for its state-of-the-art performance. In this section, we ablate the number of
predicted offsets of \ourmodel{} to investigate this further.
\cref{tab:head-ablation-comparison} reports results for implicit decoders with a different number of attention
offsets.
$K = 0$ denotes no attention offset, predicting
occupancy-flow from $\mathbf{z}_{\mathbf{q}}$ alone without cross-attention (see \cref{fig:architecture-diagram}).
We first note that $K = 1$ clearly outperforms $K = 0$, particularly on HwySim.
This aligns with our intuition that the main function of the attention offsets is to expand the receptive field of a query point. Since vehicles travel at much lower speeds in urban than highway, AV2 has a lower effective receptive field
requirement than HwySim and thus the improvements are not as pronounced.
We observe fairly close and mixed results between one and four attention offsets. $K=1$ has the best occupancy prediction metrics, while $K=4$ is the best in some flow metrics. 
Under the assumption that the predicted offsets look back to where occupancy could have
come from in the past, $K > 1$ would only improve performance over $K = 1$ when occupancy could come
from more than one past location (e.g., complex intersections, Scene 2 of \cref{fig:argo-qualitative}).
These examples are rare in the training and evaluation datasets, and having redundant offsets in the simple cases where one offset suffices could introduce noise, explaining why $K=4$ does not outperform
$K=1$.
See the supplementary for visualizations of the attention offsets when $K=4$.

\begin{figure}[t]
  \centering
  \input{figures/inference_time_v2.tex}
  \caption{Decoder inference time as a function of the number of query points for
    the \object-free decoders presented in \cref{tab:hook-comparison} on HwySim. \ourmodel{} uses $K = 1$.
  }
  \label{fig:inference-time}
  \vspace{-1em}
\end{figure}

\paragraph{Inference Time Comparison:}
In this section we compare the decoder inference time
of explicit \object-free methods in the literature (from \cref{tab:hook-comparison})
against the decoder of \ourmodel{} with $K = 1$. \cref{fig:inference-time} presents the inference time
as a function of the number of query points. For the explicit models (\textsc{MP3}, \textsc{OccFlow}),
this includes the time to bi-linearly interpolate occupancy probability at the continuous query points.
The plot evaluates on query points in a range $(1, 2 \cdot 10^{5})$,
that most planners will operate within. For instance, $2,000$ candidate trajectories $\times$ 
$10$ timesteps per trajectory is well aligned with the literature \cite{sadat2019joint,zeng2019end,sadat2020perceive,philion2020lift,casas2021mp3}.
With $200,000$ trajectories, this allows for $10$ queries per timestep to integrate
occupancy over the volume of the SDV, which should provide a good estimation of collision probability.
We notice that for $\lesssim 20,000$ query points, \ourmodel{} has a constant inference time
because it is batched over query points. Once the GPU is saturated, the operations are run sequentially
so inference time increases approximately linearly. The explicit decoders have approximately constant
inference times (the only increase is due to bilinear interpolation), but are significantly slower
than \ourmodel{} in this ``planner-relevant'' range.

%% file: figures/main_table.tex
\begin{table*}[t]
    \setlength\tabcolsep{5pt} %
    \footnotesize
    \centering
    \begin{tabular}{@{}lcccccccccccc@{}}
    \toprule
                                                    & \multicolumn{6}{c}{\textbf{AV2}} & \multicolumn{6}{c}{\textbf{HwySim}}  \\ 
                                                    \cmidrule(l{5pt}r{5pt}){2-7} \cmidrule(l{5pt}){8-13}
                                                    & mAP \ua        & Soft-IoU \ua     & ECE \da        & EPE \da            & \multicolumn{2}{c}{Flow Grounded}  & mAP \ua        & Soft-IoU \ua    & ECE \da        & EPE \da            & \multicolumn{2}{c}{Flow Grounded}   \\
                                                    \cmidrule(r{5pt}){6-7} \cmidrule{12-13}
                                                    &                &                 &                &                    & mAP \ua        & Soft-IoU \da       &                &                &                &                    & mAP \ua        & Soft-IoU \da        \\ 
                                                    \cmidrule(l{5pt}r{5pt}){2-7} \cmidrule(l{5pt}){8-13}
    \textsc{MultiPath} \cite{chai2019multipath}     & 0.625          & 0.398           & 0.916          & 0.982              & 0.803          & 0.321             & 0.299          & 0.231          & 0.433          & 4.227              & 0.463          & 0.154              \\
    \textsc{LaneGCN} \cite{lanegcn}                 & 0.620          & 0.449           & 1.138          & 0.709              & 0.778          & 0.350             & 0.472          & 0.283          & 0.337          & 2.951              & 0.636          & 0.194               \\ 
    \textsc{GoRela} \cite{cui2022gorela}            & 0.609          & 0.453           & 1.161          & 0.671              & 0.813          & 0.355             & 0.548          & 0.259          & 0.288          & 2.206              & 0.722          & 0.166               \\ 
    \textsc{OccFlow} \cite{mahjourian2022occupancy} & 0.675          & 0.356           & 0.348          & 0.390              & 0.886          & 0.493             & 0.597          & 0.370          & 0.260          & 0.842              & 0.841          & 0.330               \\
    \textsc{MP3} \cite{casas2021mp3}                & 0.774          & 0.422           & 0.201          & 0.472              & 0.902          & 0.466             & 0.637          & 0.246          & 0.208          & 1.172              & 0.833          & 0.193               \\
    \ourmodel                                       & \textbf{0.799} & \textbf{0.480}  & \textbf{0.193} & \textbf{0.267}     & \textbf{0.936} & \textbf{0.597}    & \textbf{0.716} & \textbf{0.415} & \textbf{0.076} & \textbf{0.510}     & \textbf{0.886} & \textbf{0.492}      \\ \bottomrule
    \end{tabular}
    \caption{
      Comparing our proposed model \ourmodel{} to state-of-the-art perception and prediction models on AV2 and HwySim. 
      The first three rows are \object-based models, while the others are \object-free.
    }
    \label{tab:hook-comparison}
    \vspace{-1em}
  \end{table*}

%% file: figures/qualitative_comparison_v2.tex
\newcommand*{\w}{0.19\textwidth}

\begin{figure*}[t]
    \centering
    \definecolor{hallucination-color}{RGB}{219, 94, 86}
    \definecolor{fading-color}{RGB}{184, 219, 86}
    \definecolor{map-inconsistent-color}{RGB}{86, 219, 147}
    \definecolor{actor-inconsistent-color}{RGB}{86, 131, 219}
    \definecolor{miss-detection-color}{RGB}{200, 86, 219}
    \begin{tikzpicture}
        \node[inner sep=0pt, outer sep=0, anchor=north, draw=black, line width=\lw] (104-45-gt)        at (0,0)    {\includegraphics[width=\w,trim={0cm, 2cm, 0cm, 0cm}, clip]{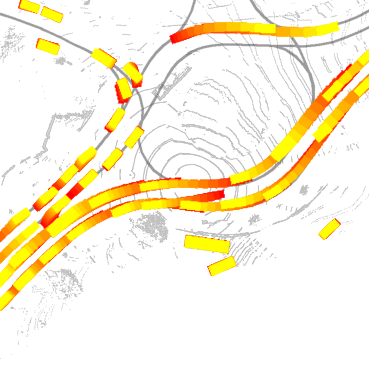}};
        \node[inner sep=0pt, outer sep=0, anchor=north, draw=black, line width=\lw] (104-45-gorela)    at (\w,0)   {\includegraphics[width=\w,trim={0cm, 2cm, 0cm, 0cm}, clip]{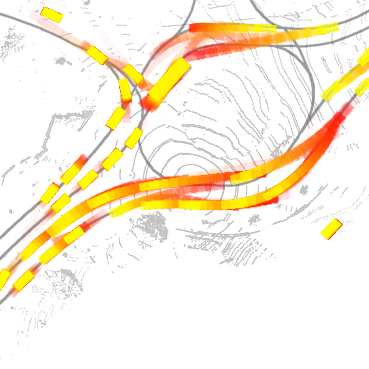}};
        \node[inner sep=0pt, outer sep=0, anchor=north, draw=black, line width=\lw] (104-45-occflow)   at (2*\w,0) {\includegraphics[width=\w,trim={0cm, 2cm, 0cm, 0cm}, clip]{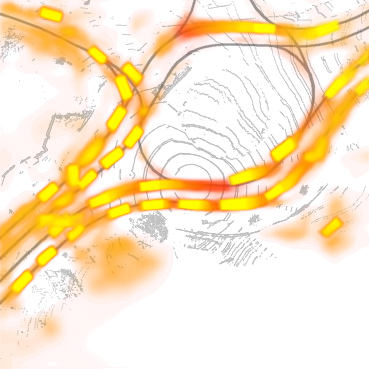}};
        \node[inner sep=0pt, outer sep=0, anchor=north, draw=black, line width=\lw] (104-45-mp3)       at (3*\w,0) {\includegraphics[width=\w,trim={0cm, 2cm, 0cm, 0cm}, clip]{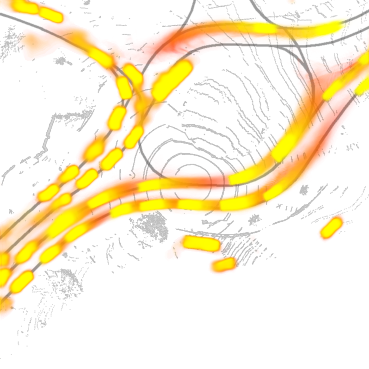}};
        \node[inner sep=0pt, outer sep=0, anchor=north, draw=black, line width=\lw] (104-45-implicito) at (4*\w,0) {\includegraphics[width=\w,trim={0cm, 2cm, 0cm, 0cm}, clip]{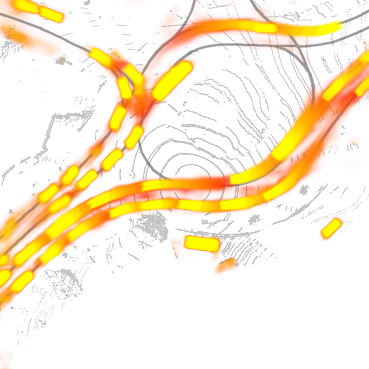}};

        \node[anchor = south] at (104-45-gt.north) {Ground Truth};
        \node[anchor = south] at (104-45-gorela.north) {\textsc{GoRela}};
        \node[anchor = south] at (104-45-occflow.north) {\textsc{OccFlow}};
        \node[anchor = south] at (104-45-mp3.north) {\textsc{MP3}};
        \node[anchor = south] at (104-45-implicito.north) {\ourmodel{}};

        \node[inner sep=0pt, outer sep=0, anchor=north, draw=black, line width=\lw] (31-45-gt)         at (104-45-gt.south)        {\includegraphics[width=\w,trim={0cm, 1.5cm, 0cm, 0.5cm}, clip]{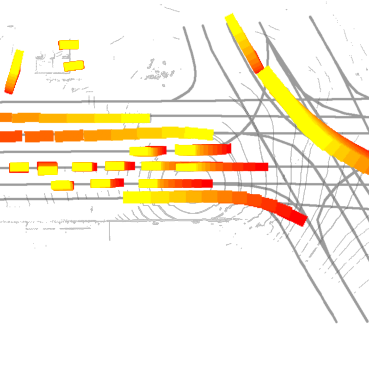}};
        \node[inner sep=0pt, outer sep=0, anchor=north, draw=black, line width=\lw] (31-45-gorela)     at (104-45-gorela.south)    {\includegraphics[width=\w,trim={0cm, 1.5cm, 0cm, 0.5cm}, clip]{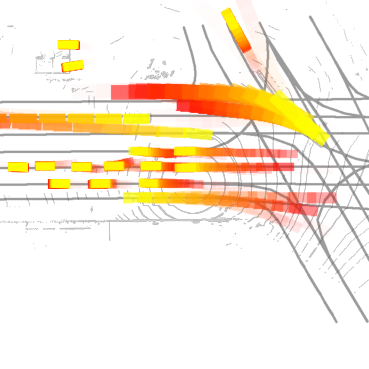}};
        \node[inner sep=0pt, outer sep=0, anchor=north, draw=black, line width=\lw] (31-45-occflow)    at (104-45-occflow.south)   {\includegraphics[width=\w,trim={0cm, 1.5cm, 0cm, 0.5cm}, clip]{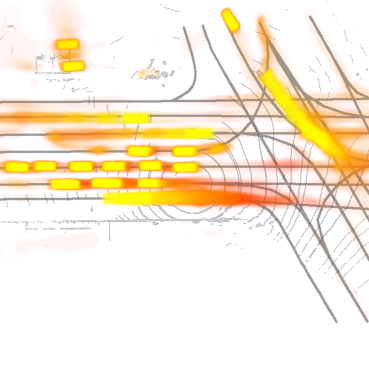}};
        \node[inner sep=0pt, outer sep=0, anchor=north, draw=black, line width=\lw] (31-45-mp3)        at (104-45-mp3.south)       {\includegraphics[width=\w,trim={0cm, 1.5cm, 0cm, 0.5cm}, clip]{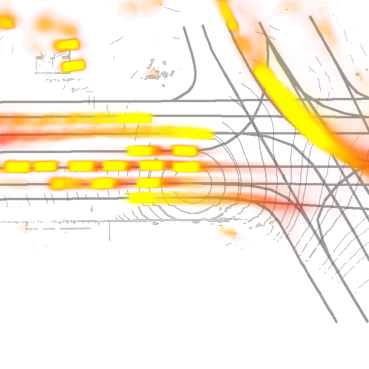}};
        \node[inner sep=0pt, outer sep=0, anchor=north, draw=black, line width=\lw] (31-45-implicito)  at (104-45-implicito.south) {\includegraphics[width=\w,trim={0cm, 1.5cm, 0cm, 0.5cm}, clip]{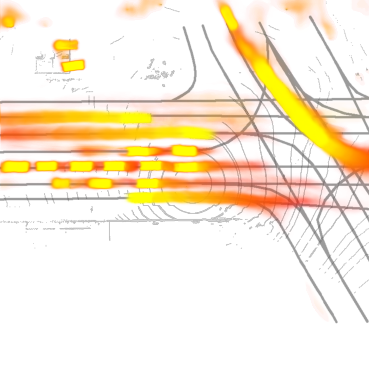}};

        \node[inner sep=0pt, outer sep=0, anchor=north, draw=black, line width=\lw] (24-75-gt)         at (31-45-gt.south)        {\includegraphics[width=\w, trim={0cm, 1cm, 0cm, 1cm}, clip]{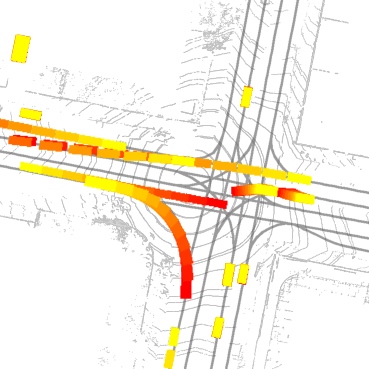}};
        \node[inner sep=0pt, outer sep=0, anchor=north, draw=black, line width=\lw] (24-75-gorela)     at (31-45-gorela.south)    {\includegraphics[width=\w, trim={0cm, 1cm, 0cm, 1cm}, clip]{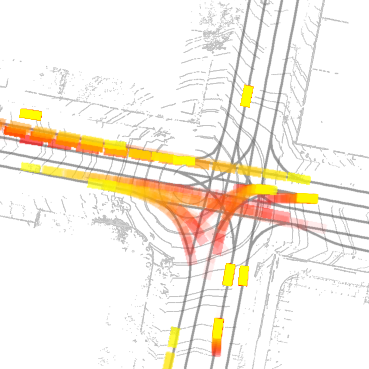}};
        \node[inner sep=0pt, outer sep=0, anchor=north, draw=black, line width=\lw] (24-75-occflow)    at (31-45-occflow.south)   {\includegraphics[width=\w, trim={0cm, 1cm, 0cm, 1cm}, clip]{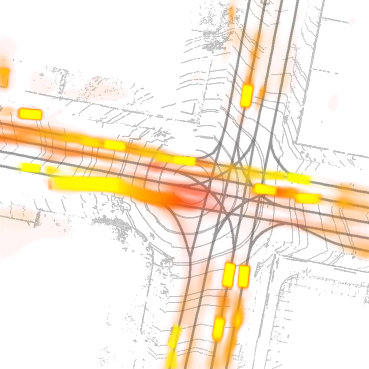}};
        \node[inner sep=0pt, outer sep=0, anchor=north, draw=black, line width=\lw] (24-75-mp3)        at (31-45-mp3.south)       {\includegraphics[width=\w, trim={0cm, 1cm, 0cm, 1cm}, clip]{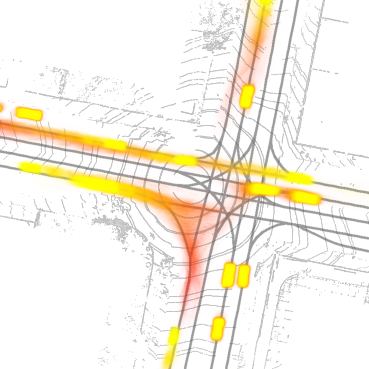}};
        \node[inner sep=0pt, outer sep=0, anchor=north, draw=black, line width=\lw] (24-75-implicito)  at (31-45-implicito.south) {\includegraphics[width=\w, trim={0cm, 1cm, 0cm, 1cm}, clip]{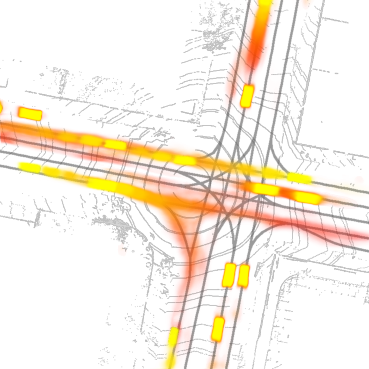}};

        \node[inner sep=0pt, outer sep=0, anchor=north, draw=black, line width=\lw] (14-105-gt)        at (24-75-gt.south)        {\includegraphics[width=\w, trim={0cm, 1cm, 0cm, 1cm}, clip]{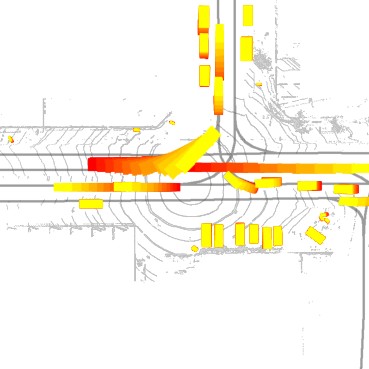}};
        \node[inner sep=0pt, outer sep=0, anchor=north, draw=black, line width=\lw] (14-105-gorela)    at (24-75-gorela.south)    {\includegraphics[width=\w, trim={0cm, 1cm, 0cm, 1cm}, clip]{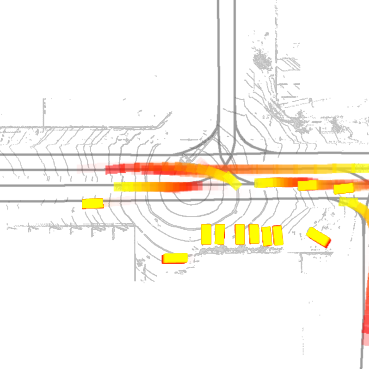}};
        \node[inner sep=0pt, outer sep=0, anchor=north, draw=black, line width=\lw] (14-105-occflow)   at (24-75-occflow.south)   {\includegraphics[width=\w, trim={0cm, 1cm, 0cm, 1cm}, clip]{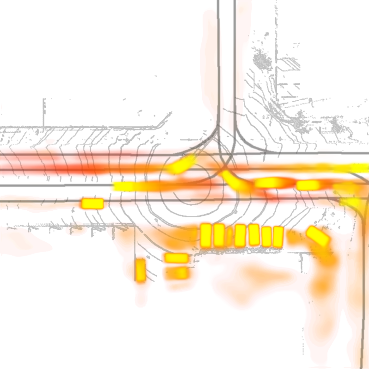}};
        \node[inner sep=0pt, outer sep=0, anchor=north, draw=black, line width=\lw] (14-105-mp3)       at (24-75-mp3.south)       {\includegraphics[width=\w, trim={0cm, 1cm, 0cm, 1cm}, clip]{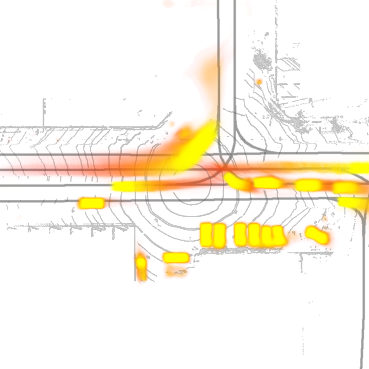}};
        \node[inner sep=0pt, outer sep=0, anchor=north, draw=black, line width=\lw] (14-105-implicito) at (24-75-implicito.south) {\includegraphics[width=\w, trim={0cm, 1cm, 0cm, 1cm}, clip]{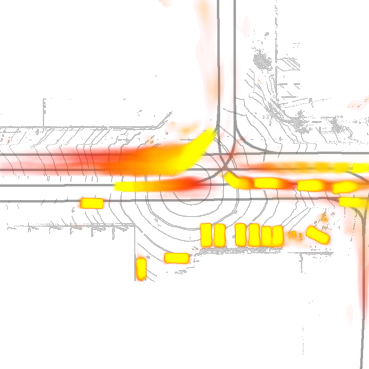}};

        \node[anchor = south, rotate=90] at (104-45-gt.west) {Scene 1};
        \node[anchor = south, rotate=90] at (31-45-gt.west) {Scene 2};
        \node[anchor = south, rotate=90] at (24-75-gt.west) {Scene 3};
        \node[anchor = south, rotate=90] at (14-105-gt.west) {Scene 4};

        \node[anchor=west] at (31-45-implicito.south east) {\includegraphics[width=0.056\textwidth]{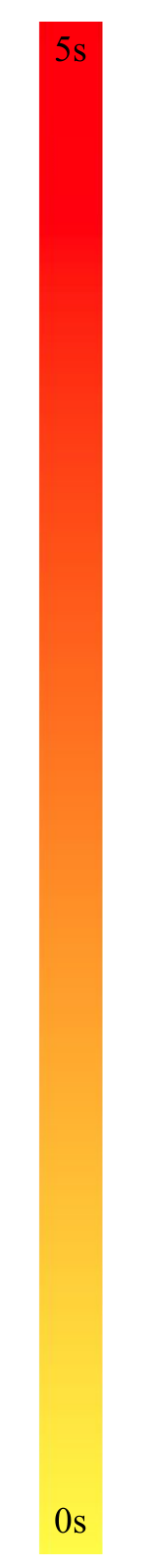}};

        \node at(104-45-occflow.center)[draw, color=hallucination-color,line width=\bw, minimum width=0.05\textwidth, minimum height=0.04\textwidth,yshift=-0.05\textwidth,xshift=-0.04\textwidth]{}; 
        \node at(104-45-occflow.center)[draw, color=hallucination-color,line width=\bw, minimum width=0.05\textwidth, minimum height=0.04\textwidth,yshift=+0.05\textwidth,xshift=-0.06\textwidth]{}; 
        \node at(14-105-occflow.center)[draw, color=hallucination-color,line width=\bw, minimum width=0.12\textwidth, minimum height=0.05\textwidth,yshift=-0.04\textwidth,xshift=0.03\textwidth]{}; 

        \node at(104-45-mp3.center)[draw, color=fading-color,line width=\bw, minimum width=0.08\textwidth, minimum height=0.04\textwidth,yshift=-0.03\textwidth,xshift=-0.02\textwidth]{}; 
        \node at(31-45-mp3.center)[draw, color=fading-color,line width=\bw, minimum width=0.12\textwidth, minimum height=0.03\textwidth,yshift=-0.02\textwidth,xshift=0.01\textwidth]{}; 
        \node at(24-75-mp3.center)[draw, color=fading-color,line width=\bw, minimum width=0.06\textwidth, minimum height=0.08\textwidth,yshift=-0.03\textwidth,xshift=-0.00\textwidth]{}; 
        \node at(14-105-mp3.center)[draw, color=fading-color,line width=\bw, minimum width=0.1\textwidth, minimum height=0.05\textwidth,yshift=0.01\textwidth,xshift=-0.025\textwidth]{}; 
        \node at(31-45-occflow.center)[draw, color=fading-color,line width=\bw, minimum width=0.12\textwidth, minimum height=0.03\textwidth,yshift=0.02\textwidth,xshift=-0.03\textwidth]{}; 
        \node at(24-75-occflow.center)[draw, color=fading-color,line width=\bw, minimum width=0.04\textwidth, minimum height=0.07\textwidth,yshift=-0.03\textwidth,xshift=-0.00\textwidth]{}; 

        \node at(14-105-gorela.center)[draw, color=miss-detection-color,line width=\bw, minimum width=0.06\textwidth, minimum height=0.04\textwidth,yshift=0.01\textwidth,xshift=-0.00\textwidth]{}; 
        \node at(14-105-occflow.center)[draw, color=miss-detection-color,line width=\bw, minimum width=0.06\textwidth, minimum height=0.04\textwidth,yshift=0.01\textwidth,xshift=-0.00\textwidth]{}; 

        \node at(31-45-gorela.center)[draw, color=map-inconsistent-color,line width=\bw, minimum width=0.12\textwidth, minimum height=0.04\textwidth,yshift=0.03\textwidth,xshift=0.02\textwidth]{}; 
        \node at(104-45-gorela.center)[draw, color=map-inconsistent-color,line width=\bw, minimum width=0.08\textwidth, minimum height=0.04\textwidth,yshift=0.05\textwidth,xshift=0.03\textwidth]{}; 

        \node at(24-75-gorela.center)[draw, color=actor-inconsistent-color,line width=\bw, minimum width=0.08\textwidth, minimum height=0.08\textwidth,yshift=-0.02\textwidth,xshift=0.00\textwidth]{}; 
        \node at(104-45-gorela.center)[draw, color=actor-inconsistent-color,line width=\bw, minimum width=0.08\textwidth, minimum height=0.05\textwidth,yshift=-0.00\textwidth,xshift=0.05\textwidth]{}; 
        
    \end{tikzpicture}
    \caption{
        Occupancy predictions of various models (columns) across four scenes (rows) in AV2.
        Opacity denotes occupancy probability, 
        and the colormap indicates prediction time $\Delta t$ (from current to future horizon, as shown on the right).
        Failure modes are highlighted with colored boxes:
        \textbf{\textcolor{hallucination-color}{occupancy hallucination}},
        \textbf{\textcolor{fading-color}{fading/missing occupancy}}, 
        \textbf{\textcolor{map-inconsistent-color}{inconsistent with map}}, 
        \textbf{\textcolor{actor-inconsistent-color}{inconsistent with actors}}, 
        \textbf{\textcolor{miss-detection-color}{miss-detection}}.
    }
    \vspace{-10pt}
    \label{fig:argo-qualitative}    
\end{figure*}

%% file: figures/introspection.tex
\begin{figure}[t]

    \begin{tikzpicture}
        \node[inner sep=0pt, outer sep=0, anchor=north, draw=black, line width=1.5pt] (104-45-motion)        at (0,0)    {
            \includegraphics[width=0.47\linewidth,trim={0cm, 2cm, 0cm, 0cm}, clip]{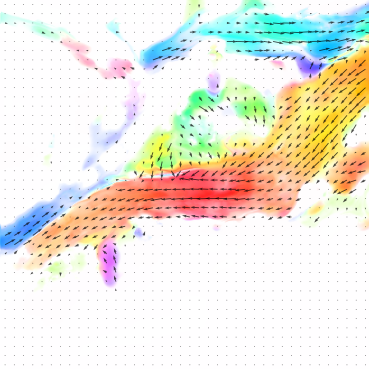}
        };
        \node[inner sep=0pt, outer sep=0, anchor=south west,line width=1.5pt] at (104-45-motion.south west)    {
            \includegraphics[width = 0.06\linewidth]{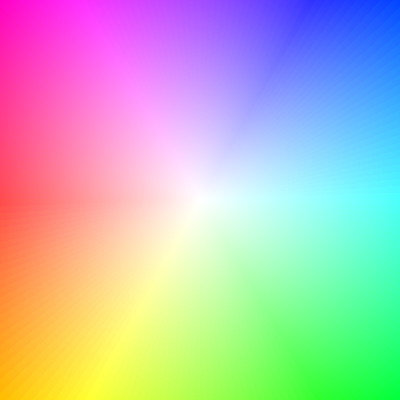}
        };
        \node[inner sep=0pt, outer sep=0, anchor=north, draw=black, line width=1.5pt] (104-45-attention)    at (0.47\linewidth,0)   {
            \includegraphics[width = 0.47\linewidth,trim={0cm, 2cm, 0cm, 0cm}, clip]{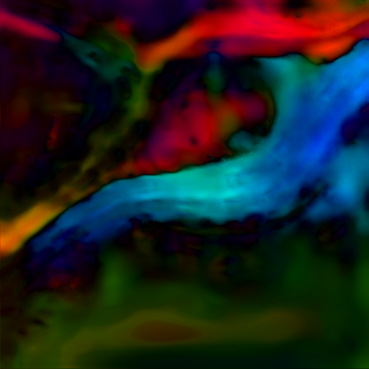}
        };
        \node[inner sep=0pt, outer sep=0, anchor=south west, line width=1.5pt] at (104-45-attention.south west)    {
            \includegraphics[width = 0.06\linewidth]{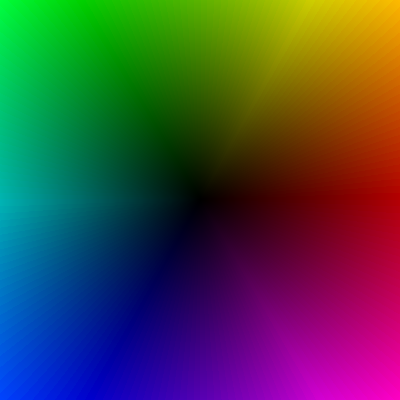}
        };
        \node[inner sep=0pt, outer sep=0, anchor=north,  draw=black, line width=1.5pt] (31-45-motion)   at (104-45-motion.south) {
            \includegraphics[width = 0.47\linewidth, trim={0cm, 1cm, 0cm, 1cm}, clip]{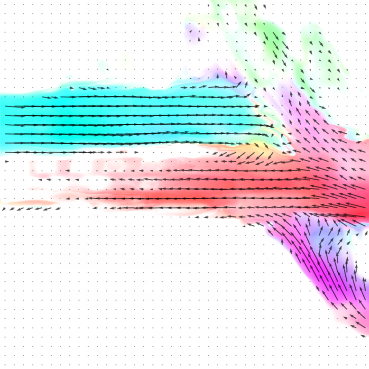}
        };
        \node[inner sep=0pt, outer sep=0, anchor=south west, line width=1.5pt] at (31-45-motion.south west)    {
                \includegraphics[width = 0.06\linewidth]{figures/raw-images/argo_color_wheel_inverted.png}
        };
        \node[inner sep=0pt, outer sep=0, anchor=north, draw=black, line width=1.5pt] (31-45-attention)   at (104-45-attention.south) {
            \includegraphics[width = 0.47\linewidth, trim={0cm, 1cm, 0cm, 1cm}, clip]{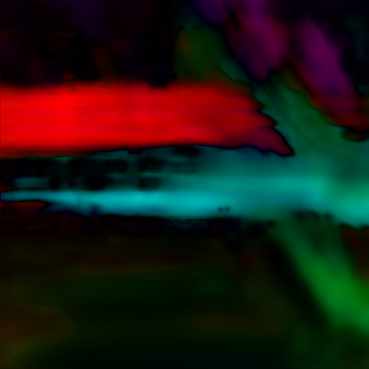}
        };
        \node[inner sep=0pt, outer sep=0, anchor=south west, line width=1.5pt] at (31-45-attention.south west)    {
            \includegraphics[width = 0.06\linewidth]{figures/raw-images/argo_color_wheel.png}
        };
        \node[anchor=south] at (104-45-motion.north) {\Reverse{} Flow};
        \node[anchor=south] at (104-45-attention.north) {Attention Offsets};
        \node[anchor=south, rotate=90] at (104-45-motion.west) {Scene 1};
        \node[anchor=south, rotate=90] at (31-45-motion.west) {Scene 2};
    \end{tikzpicture}
    \caption{Visualizations of the \reverse{} flow field predictions and attention offset predictions of \ourmodel{}
    at the last timestep of the prediction horizon on Scene 1 and Scene 2 from \cref{fig:argo-qualitative}.}
    \vspace{-10pt}
    \label{fig:argo-introspection}
\end{figure}

%% file: figures/decoder_ablation_table.tex
\begin{table}[t]
    \scriptsize
    \centering
    \begin{threeparttable}
    \setlength\tabcolsep{2.5pt} %
    \begin{tabularx}{\columnwidth}{l | l  c c c c c c}
    \toprule
    \multicolumn{1}{c}{}  &   \multicolumn{1}{c}{}                                                                 & mAP \ua        & Soft-IoU \ua    & ECE \da         &  EPE \da           & \multicolumn{2}{c}{Flow Grounded}  \\
    \cmidrule(r{2pt}){7-8}                                                                                                                                  
    \multicolumn{1}{c}{}  &   \multicolumn{1}{c}{}                                                                 &                &                &                 &                    & mAP \ua        & Soft-IoU \ua         \\ \midrule
    \parbox[t]{2mm}{\multirow{4}{*}{\rotatebox[origin=c]{90}{\textbf{AV2}}}}                                       &   
            \textsc{MP3} \cite{casas2021mp3}                                                                       & 0.774          & 0.422          & 0.201           & 0.472              & 0.902          & 0.466              \\
    {}  &   \textsc{ConvNet}                                                                                       & 0.796          & 0.466          & \textbf{0.135}  & 0.312              & 0.929          & 0.581              \\
    {}  &   \textsc{ConvNetFT} \cite{mahjourian2022occupancy}                                                      & 0.796          & 0.475          & 0.198           & 0.302              & 0.929          & 0.582              \\
    {}  &   \ourmodel                                                                                              & \textbf{0.799} & \textbf{0.480} & 0.193           & \textbf{0.267}     & \textbf{0.936} & \textbf{0.597}     \\ 
    \midrule                                                                                                                                                                                                 
    \parbox[t]{2mm}{\multirow{4}{*}{\rotatebox[origin=c]{90}{\textbf{HwySim}}}}  &
         \textsc{MP3} \cite{casas2021mp3}                                                                          & 0.637          & 0.246          & 0.208           & 1.172              & 0.833          & 0.193          \\
    {}  &\textsc{ConvNet}                                                                                          & 0.648          & 0.344          & \textbf{0.024}  & 0.657              & 0.859          & 0.408          \\
    {}  &\textsc{ConvNetFT} \cite{mahjourian2022occupancy}                                                         & 0.654          & 0.351          & \textbf{0.024}  & 0.657              & 0.860          & 0.416          \\
    {}  &\ourmodel                                                                                                 & \textbf{0.716} & \textbf{0.415} & 0.076           & \textbf{0.510}     & \textbf{0.886} & \textbf{0.492} \\ \bottomrule
    \end{tabularx}
    \end{threeparttable}
    \caption{Comparing the performance of occupancy-flow decoders, trained end-to-end, with the same encoder.}
    \label{tab:decoder-comparison}
    \vspace{-1em}
  \end{table}

%% file: figures/num_offset_ablation.tex
\begin{table}[t]
  \scriptsize
  \centering
  \begin{threeparttable}
  \setlength\tabcolsep{3.5pt} %
  \begin{tabularx}{\columnwidth}{l | c c c c c c c}
  \toprule
  \multicolumn{1}{c}{}  &   \multicolumn{1}{c}{Num. offsets}                                            & mAP \ua           & Soft-IoU \ua        & ECE \da             & EPE \da                 & \multicolumn{2}{c}{Flow Grounded}   \\
  \cmidrule{7-8}                                                                                                                                  
  \multicolumn{1}{c}{}  &   \multicolumn{1}{c}{}                                                                 &                &                &                 &                & mAP \ua             & Soft-IoU \ua            \\ \midrule
  \parbox[t]{2mm}{\multirow{3}{*}{\rotatebox[origin=c]{90}{\textbf{AV2}}}}  &   $K = 0$                          & 0.790          & 0.456          & \textbf{0.128}  & 0.300          & 0.930          & 0.583                    \\
  {}  &   $K = 1$                                                                                                & \textbf{0.799} & \textbf{0.480} & 0.193           & 0.267          & \textbf{0.936} & \textbf{0.597}           \\
  {}  &   $K = 4$                                                                                                & 0.797          & 0.478          & 0.257           & \textbf{0.252} & \textbf{0.936} & 0.570                    \\ \midrule
  \parbox[t]{2mm}{\multirow{3}{*}{\rotatebox[origin=c]{90}{\textbf{HwySim}}}}  & $K = 0$                         & 0.649          & 0.359          & 0.052           & 0.686          & 0.857          & 0.421                     \\[1pt]
  {}  &   $K = 1$                                                                                                & \textbf{0.716} & \textbf{0.415} & 0.076           & 0.510          & 0.886          & \textbf{0.492}            \\
  {}  &   $K = 4$                                                                                                & 0.714          & 0.404          & \textbf{0.051}  & \textbf{0.509} & \textbf{0.890} & 0.487                     \\[1pt] \bottomrule
  \end{tabularx}
  \end{threeparttable}
  \caption{Ablation study on the effect of the number of predicted attention offsets on the performance of \ourmodel{}.}
  \label{tab:head-ablation-comparison}
  \vspace{-2em}
\end{table}
\setlength{\extrarowheight}{0pt}

%% file: figures/inference_time_v2.tex
\begin{tikzpicture}

\definecolor{darkgray176}{RGB}{176,176,176}
\definecolor{green01270}{RGB}{0,127,0}
\definecolor{lightgray204}{RGB}{204,204,204}

\begin{axis}[
width=\linewidth,
height=0.6\linewidth,
legend cell align={left},
legend style={
  fill opacity=0.6,
  draw opacity=1,
  text opacity=1,
  at={(0.97, 0.03)},
  anchor=south east ,
  draw=lightgray204,
  font=\scriptsize,
  nodes={scale=0.8, transform shape}, 
},
legend image post style={scale=0.4},
tick align=outside,
tick pos=left,
label style={font=\footnotesize},
tick label style={font=\scriptsize},
x grid style={darkgray176},
xlabel={Number of Query Points $|\mathcal{Q}|$},
xmin=-9949, xmax=208951,
xtick style={color=black},
y grid style={darkgray176},
ylabel={Inference Time (ms)},
ymin=-0.0649032767613735, ymax=37,
ytick style={color=black},
mark size=1pt,
]
\addplot [thick, MP3Color, opacity=1.0]
table {%
1 35.2025743624522
1001 35.2005380862423
2001 35.1968530488135
3001 35.1922212352337
4001 35.187519667223
5001 35.1836473615725
6001 35.1813892969438
7001 35.1813103233864
8001 35.1837043681784
9001 35.1886122440105
10001 35.195853319297
11001 35.2050724474835
12001 35.2158046375354
13001 35.2275090376815
14001 35.2396165790995
15001 35.2515512620425
16001 35.2627264768751
17001 35.2725842648965
18001 35.2805738699969
19001 35.2861961841617
20001 35.2890123706155
21001 35.2887265268437
22001 35.2852295441916
23001 35.2786630189698
24001 35.2695225621315
25001 35.258660855883
26001 35.2472773413511
27001 35.2368118788712
28001 35.2287700918653
29001 35.2245360514925
30001 35.2251213351977
31001 35.2309896882691
32001 35.2419566900435
33001 35.2571697255684
34001 35.2752463482359
35001 35.294468190206
36001 35.3130612305425
37001 35.3294577737954
38001 35.3425185956885
39001 35.3516562703211
40001 35.3568783939927
41001 35.3586842281878
42001 35.357923092325
43001 35.3556025613446
44001 35.3526946369305
45001 35.3499757176005
46001 35.3479339432132
47001 35.3467138853169
48001 35.3461373737168
49001 35.3457680848237
50001 35.3449849368217
51001 35.3431052588811
52001 35.3394996326038
53001 35.3336864707781
54001 35.3254185348503
55001 35.3147301960763
56001 35.3019275168066
57001 35.2875822911401
58001 35.2724151925854
59001 35.2572259829786
60001 35.2427782339135
61001 35.229722438177
62001 35.2185338449731
63001 35.2094994832973
64001 35.2027197368188
65001 35.1981489416601
66001 35.1956317632854
67001 35.1949507286713
68001 35.1958403054616
69001 35.1980360015224
70001 35.2012542752053
71001 35.205226619294
72001 35.2097011376347
73001 35.2144436379104
74001 35.2192672909248
75001 35.2240126169132
76001 35.2285840768396
77001 35.2329307149021
78001 35.2370640857739
79001 35.2410299505403
80001 35.2449260470818
81001 35.2488720697091
82001 35.2530126285169
83001 35.2574984928728
84001 35.2624667602814
85001 35.268026774719
86001 35.2742371426573
87001 35.2810999137316
88001 35.2885472007731
89001 35.2964460344399
90001 35.3046038952809
91001 35.3127902743329
92001 35.3207590532459
93001 35.3282693021725
94001 35.3351226596492
95001 35.3411737843373
96001 35.3463556459309
97001 35.3506883583174
98001 35.3542683655874
99001 35.3573179286843
100001 35.3600302967274
101001 35.3627150936809
102001 35.365664596916
103001 35.3691890858131
104001 35.3735718815461
105001 35.3790328549636
106001 35.3857680809176
107001 35.3938844162971
108001 35.4033594754076
109001 35.4140619325179
110001 35.4256228613004
111001 35.4374631373688
112001 35.4487725310641
113001 35.4585607118165
114001 35.4656858639919
115001 35.4691493479727
116001 35.4681935035051
117001 35.4626286476018
118001 35.4529845876837
119001 35.4405155019275
120001 35.4272148811273
121001 35.4154938669117
122001 35.4078607855805
123001 35.4065205468225
124001 35.412992756237
125001 35.4278583704425
126001 35.4506896347461
127001 35.4800815949546
128001 35.5138920194153
129001 35.5495026812015
130001 35.5842032590153
131001 35.6155026863402
132001 35.6414807920522
133001 35.6609411894801
134001 35.6736161265452
135001 35.6800927650755
136001 35.6816631882156
137001 35.6800172266273
138001 35.6768996262517
139001 35.6737232499783
140001 35.6713900586161
141001 35.6702545584083
142001 35.6700319498447
143001 35.6701973157653
144001 35.6701320371522
145001 35.6693895853314
146001 35.6677949223291
147001 35.6654082462963
148001 35.6624831943669
149001 35.6592080526061
150001 35.6556446565664
151001 35.6517031082567
152001 35.6471248165645
153001 35.6416152705539
154001 35.6350148808314
155001 35.6274937491344
156001 35.6195043726487
157001 35.6119258907327
158001 35.6058376266173
159001 35.6024286059066
160001 35.6028111237136
161001 35.607726749842
162001 35.6174719158619
163001 35.6317645699139
164001 35.6497585443167
165001 35.6701478099344
166001 35.6913594081309
167001 35.7117488809262
168001 35.7299148588648
169001 35.7448468261922
170001 35.7560702679265
171001 35.7636602274693
172001 35.7681835643526
173001 35.7704787325906
174001 35.7715486073311
175001 35.7722953070763
176001 35.7734410845591
177001 35.7754576186769
178001 35.7785692563475
179001 35.7828034465909
180001 35.7880839174559
181001 35.7942306701472
182001 35.8010688552128
183001 35.8084329695949
184001 35.8160939545519
185001 35.82377763682
186001 35.8311182390657
187001 35.837665991483
188001 35.8428739081757
189001 35.846208864372
190001 35.8472800917102
191001 35.8457935998192
192001 35.8417406103672
193001 35.8354223699441
194001 35.8274254991094
195001 35.8185797849614
196001 35.8098479047695
197001 35.8022258295488
198001 35.7965892212404
199001 35.793597331362
};
\addlegendentry{MP3}
\addplot [thick, OccFlowColor, opacity=1.0]
table {%
1 28.9891264745273
1001 28.9762710172617
2001 28.9521444664965
3001 28.9196128579657
4001 28.882302226401
5001 28.8439819633815
6001 28.8080008105101
7001 28.7768855342267
8001 28.7521570564519
9001 28.7343569899086
10001 28.7232349782742
11001 28.7180081482941
12001 28.7176310073223
13001 28.7210254324718
14001 28.7272279718183
15001 28.7354668644504
16001 28.7451798147402
17001 28.755993068529
18001 28.767665798426
19001 28.7800449160633
20001 28.7929850469757
21001 28.8064164585861
22001 28.8202269601401
23001 28.8342153630931
24001 28.8482206070791
25001 28.8620801733534
26001 28.8756500156224
27001 28.8888179564019
28001 28.901523287331
29001 28.9137509648545
30001 28.9255236969206
31001 28.9368898004244
32001 28.9479047210576
33001 28.9586126502148
34001 28.9690467254986
35001 28.9792093044493
36001 28.9890900362538
37001 28.9986681213258
38001 29.0079268213401
39001 29.01685818158
40001 29.0254803339124
41001 29.0338384790374
42001 29.0419991129405
43001 29.0500460218636
44001 29.0580705188314
45001 29.0661591368823
46001 29.0743901733987
47001 29.082819804124
48001 29.0914809183588
49001 29.1003759744538
50001 29.1094827945228
51001 29.1187415777383
52001 29.1280665765075
53001 29.1373504952328
54001 29.1464662627966
55001 29.1552929294439
56001 29.1637279967251
57001 29.1717061857004
58001 29.1792152829498
59001 29.1863034355926
60001 29.1930686553462
61001 29.1996446419227
62001 29.2061750053228
63001 29.2127850987784
64001 29.2195606620608
65001 29.226532273673
66001 29.2336794387932
67001 29.2409414947418
68001 29.248239604476
69001 29.2555000226397
70001 29.2626792985628
71001 29.2697657249135
72001 29.2767804810087
73001 29.2837751975412
74001 29.2908142218182
75001 29.2979522906903
76001 29.3052460505042
77001 29.3127376097372
78001 29.3204610650567
79001 29.3284539001924
80001 29.3367568490764
81001 29.3454090624068
82001 29.3544438836485
83001 29.3638782879799
84001 29.3736983783159
85001 29.3838549372157
86001 29.39425779596
87001 29.404784370032
88001 29.4152883645528
89001 29.4256192652377
90001 29.4356357455498
91001 29.4452174775229
92001 29.4542852904555
93001 29.462801334624
94001 29.4707768699209
95001 29.4782667319252
96001 29.4853620469671
97001 29.4921693181692
98001 29.4987972589786
99001 29.5053359433582
100001 29.511832646502
101001 29.5182907214331
102001 29.5246701459666
103001 29.5308854844798
104001 29.5368362098343
105001 29.542417634791
106001 29.5475431319382
107001 29.5521633795458
108001 29.556274127736
109001 29.5599198949787
110001 29.5632013139915
111001 29.5662595093203
112001 29.5692660738023
113001 29.5724084296586
114001 29.57586528635
115001 29.5797925091185
116001 29.5843141800404
117001 29.5895183939481
118001 29.5954573329209
119001 29.6021701348111
120001 29.6096813824563
121001 29.6180119591477
122001 29.62717384861
123001 29.6371576675915
124001 29.647913043888
125001 29.6593406691326
126001 29.6712615138016
127001 29.6834460773244
128001 29.6956087465503
129001 29.7074607006465
130001 29.7187288255346
131001 29.729202656888
132001 29.7387716908724
133001 29.7474534453269
134001 29.7554183045238
135001 29.7629821924634
136001 29.770602750312
137001 29.7788553763577
138001 29.7883759296292
139001 29.7997942391696
140001 29.813663083712
141001 29.8303939706214
142001 29.8501871908226
143001 29.8729849448947
144001 29.8984847088081
145001 29.9261403795468
146001 29.9552147080184
147001 29.9848847352797
148001 30.0143245387537
149001 30.0428123294999
150001 30.0698096930741
151001 30.0950141207483
152001 30.1183815798137
153001 30.1401202320895
154001 30.1605933116486
155001 30.1802196069082
156001 30.1993450215513
157001 30.2181575155362
158001 30.2365886897921
159001 30.2543136969013
160001 30.2708055399832
161001 30.2854437799394
162001 30.2976647236533
163001 30.3070999281479
164001 30.3137002240004
165001 30.3177949175124
166001 30.3200334159362
167001 30.3213056458644
168001 30.3225729417482
169001 30.3247087523842
170001 30.3283114737268
171001 30.3336401242485
172001 30.3406147907853
173001 30.3488648824718
174001 30.3578420523861
175001 30.366958367293
176001 30.3757349409323
177001 30.3838789738177
178001 30.3912772872268
179001 30.397975484403
180001 30.4040663557394
181001 30.4095985916573
182001 30.4145152617036
183001 30.4186284569636
184001 30.4216720375276
185001 30.423331812054
186001 30.4233384928906
187001 30.4215304929303
188001 30.4178601974112
189001 30.4124055970813
190001 30.4053100126429
191001 30.3967743395411
192001 30.3870652407026
193001 30.3765305877458
194001 30.3655999044223
195001 30.3548345358349
196001 30.3449245305735
197001 30.3366178667603
198001 30.3306030249039
199001 30.3274414977773
};
\addlegendentry{\textsc{OccFlow}}
\addplot [thick, ourmodelcolor1, opacity=1.0]
table {%
1 1.68252466285563
1001 1.68347799603241
2001 1.68532477266525
3001 1.68796811521959
4001 1.69132311343685
5001 1.6953699968282
6001 1.7001932076185
7001 1.70599886440323
8001 1.71309934555353
9001 1.72187549452109
10001 1.73271122532948
11001 1.74593843442452
12001 1.76179373975356
13001 1.78040094545309
14001 1.80177887762743
15001 1.82587101072197
16001 1.85259141837466
17001 1.88187888811842
18001 1.91371776819065
19001 1.94816767273812
20001 1.98536020216374
21001 2.02548768932049
22001 2.06880300061092
23001 2.11561781371384
24001 2.16628927273229
25001 2.22121080799279
26001 2.28078520812127
27001 2.34535430943874
28001 2.4151314980928
29001 2.49011602355307
30001 2.57002081287233
31001 2.6542467812023
32001 2.74190836771587
33001 2.83190009821449
34001 2.92300733102507
35001 3.01401905380682
36001 3.10383459554332
37001 3.1915335445765
38001 3.27641389227575
39001 3.35796536548985
40001 3.43583400828078
41001 3.509763344598
42001 3.57954283912619
43001 3.64496813027368
44001 3.70584181691058
45001 3.76199035311404
46001 3.81332359816749
47001 3.85989791307899
48001 3.90201158078489
49001 3.94028293755492
50001 3.97570109929862
51001 4.00964945686529
52001 4.04385486863854
53001 4.08028582156389
54001 4.1209983303442
55001 4.16794416276368
56001 4.22277447870646
57001 4.28666386667121
58001 4.36018593134763
59001 4.44325908873117
60001 4.53516612268343
61001 4.63463895138077
62001 4.73998788689979
63001 4.84925516409768
64001 4.96037682668374
65001 5.07132535144438
66001 5.1802306234697
67001 5.28548786449726
68001 5.38580316637227
69001 5.48023117535972
70001 5.56816981619667
71001 5.64933021370127
72001 5.72367872671759
73001 5.79137686235824
74001 5.85273200740086
75001 5.90815963714465
76001 5.95819168593191
77001 6.0035002768354
78001 6.04495400797855
79001 6.08366200239365
80001 6.12101205463568
81001 6.15864262624979
82001 6.19836407384973
83001 6.242022842027
84001 6.29132518822243
85001 6.34764231580937
86001 6.41185271268846
87001 6.4842317212417
88001 6.56443685209566
89001 6.65156017377494
90001 6.74426532045902
91001 6.84095602083647
92001 6.93996133544391
93001 7.03968111640776
94001 7.13869631579579
95001 7.23581787378701
96001 7.33008581429057
97001 7.42073842237601
98001 7.50716373619735
99001 7.58886482038075
100001 7.66543600151256
101001 7.73656425941511
102001 7.80205749805292
103001 7.86188154588615
104001 7.91621158335207
105001 7.96547463996274
106001 8.01038742774692
107001 8.05197798502973
108001 8.09155618837691
109001 8.13072143364349
110001 8.17124291152575
111001 8.21498752346264
112001 8.26375579997599
113001 8.31917215543961
114001 8.38251550929892
115001 8.4546524269575
116001 8.53601150283502
117001 8.62656489510732
118001 8.72595575913978
119001 8.83361009106169
120001 8.94882095767244
121001 9.0709088895903
122001 9.19923717040971
123001 9.33315347475107
124001 9.47194547757631
125001 9.61466432730685
126001 9.759960845533
127001 9.9059976206767
128001 10.0504077907708
129001 10.1904001730685
130001 10.3229599383809
131001 10.4452168543016
132001 10.5548207358379
133001 10.650373506805
134001 10.7316959196653
135001 10.8000309569293
136001 10.8579655969436
137001 10.909198570173
138001 10.9581100671327
139001 11.0092468192377
140001 11.0668386117378
141001 11.134374250527
142001 11.2143128366544
143001 11.307991356355
144001 11.4156948465557
145001 11.5368250217321
146001 11.6701355469581
147001 11.8139347327143
148001 11.9662959254631
149001 12.125199130239
150001 12.288596167798
151001 12.4545409593817
152001 12.6212306208752
153001 12.7871029169628
154001 12.9508548207482
155001 13.1114704088567
156001 13.2681597476289
157001 13.4202689599804
158001 13.5672141192761
159001 13.7083831717465
160001 13.8431411492245
161001 13.9709110587658
162001 14.0912730882001
163001 14.2041712947642
164001 14.3100856296375
165001 14.4100548998532
166001 14.5057062290874
167001 14.5991302901894
168001 14.6926173177066
169001 14.7884217763559
170001 14.8884796174875
171001 14.9942083791841
172001 15.1063574203482
173001 15.2249938693023
174001 15.3495402873172
175001 15.4789789569402
176001 15.6119817296102
177001 15.7471325499092
178001 15.8830935747914
179001 16.018754885869
180001 16.1532793007041
181001 16.2860863455393
182001 16.4168345729176
183001 16.545319713985
184001 16.6713549387067
185001 16.7946911189433
186001 16.9149438048368
187001 17.0315528793108
188001 17.1438171284624
189001 17.2509251494213
190001 17.3520193813764
191001 17.446257002264
192001 17.5328268113443
193001 17.6109558397221
194001 17.6799085485048
195001 17.7389511989988
196001 17.7873604891432
197001 17.8244357884466
198001 17.8495549719148
199001 17.8622510994295
};
\addlegendentry{\ourmodel{}}
\end{axis}

\end{tikzpicture}

%% file: sections/conclusion.tex
\section{Conclusion} 
\label{sec:conclusion}

In this paper, we have proposed a unified approach to joint perception and prediction
for self-driving that implicitly represents occupancy and flow over time with a
neural network. This queryable implicit representation can
provide information to a downstream motion planner more effectively and efficiently.
We showcased that our implicit architecture predicts occupancy and flow more accurately
than contemporary explicit approaches in both urban and highway settings.
Further, this approach outperforms more traditional object-based perception and prediction paradigms.
In the future, we plan to assess the impact of our improvements on the downstream task of motion planning.

%% file: sections/implementation_details.tex
\section{Training Details}
\label{sec:training-details}

\paragraph{Initialization:}
We found that training was more consistent if we initialized the weights with a small standard deviation of $0.01$
and a bias of $0$ for the linear layer that predicts the attention offsets $\Delta \mathbf{q}$.
This ensures that the initial offset predictions are small enough such that $\mathbf{r} = \mathbf{q} + \Delta\mathbf{q}$
still represents a point within the feature map $\mathbf{Z}$, but large enough such that the predicted offsets do not get stuck
at $\Delta \mathbf{q} = \mathbf{0}$. 

\paragraph{Loss hyperparameters:}
We use a flow regression loss weighting of $\lambda_{\mathbf{f}} = 0.1$ (downweighted by a factor of 10 relative to occupancy loss).

\paragraph{Optimizer:}
We use an initial learning rate of $1.0 \times 10^{-3}$, which is multiplied by a factor of $0.25$ every $6$ training epochs. We
train for a total of 20 epochs, using the AdamW optimizer \cite{loshchilov2017decoupled} with a weight decay of $1.0 \times 10^{-4}$.

\paragraph{Query points used during training:}
We use $|\mathcal{Q}| = 220,000$ query points per example in the batch, sampled uniformly on the spatio-temporal volume. 
Future work may investigate more effective sampling methods for training.

%% file: sections/baseline_implementation_details.tex
\section{Baseline Implementation Details}

\paragraph{Object detector:}
For a fair comparison with object-based and hybrid baselines (\textsc{MultiPath, LaneGCN, GoRela} and \textsc{OccFlow}), we use the same state-of-the-art object detector, \twostage{}, in all of their implementations. This detector is a variant of the original PIXOR \cite{yang2018pixor}, with the following improvements:
\begin{itemize}
    \item We use multi-scale deformable self-attention~\cite{zhu2020deformable} instead of the upsampling deconvolutional layers after the ResNet backbone to aggregate information from different scales and enhance the feature extraction. Following PIXOR, we output the feature maps with $1/4$ resolution. %
    \item We use the output from the dense detector header as the first-stage results. The top 500 bounding boxes after non-maximum supression (NMS) are used as region proposal for the second stage. The 2D IoU threshold for NMS is set to 0.7.
    \item We use RotatedROIAlign~\cite{wu2019detectron2} to extract $3\times 3$ ROI features from the feature map in the second-stage header, and apply two self-attention layers: one on features within each ROI and another between features of different ROIs.
    \item Two MLPs are used to predict the classification score and box refinement for the region proposals.
    \item We use DETR-like set-based loss \cite{carion2020end} with bipartite matching for both stages. An additional IoU loss is used for bounding box regression besides the smooth L1 loss.
\end{itemize}

\paragraph{Trajectory prediction baselines:}
For the methods that perform trajectory prediction (\textsc{MultiPath, LaneGCN, GoRela}), we train \twostage{} jointly with the trajectory prediction model in an end-to-end fashion, as proposed in SpAGNN\cite{casas2019spatially}.

\paragraph{OccFlow:}
This hybrid method performs object-based perception (object detection) and object-free prediction (occupancy-flow).
Since it requires kinematic information in the raster passed from perception to prediction, we train the detector to additionally output longitudinal, lateral and angular velocity and acceleration.
We then postprocess this information into a raster with five channels: occupancy probability based on detector confidence, velocity in $x$ and $y$ directions, and acceleration in $x$ and $y$ directions, for each pixel.
Given this raster, we employ the same encoder architecture as our model for a fair comparison.
Finally, a fully convolutional header decodes both occupancy and backward flow on a spatio-temporal grid.

\paragraph{MP3:}
This baseline also uses the same encoder as our model for a fair comparison. It then decodes initial occupancy and forward flow over time with a fully convolutional header, and the future occupancy is obtained by warping the initial occupancy with the temporal flow as described in the original paper.

%% file: sections/additional_results.tex
\input{figures/multihead_qualitative_v2.tex}

\section{Additional Results}

\subsection{Qualitative Results for \ourmodel{} with $K = 4$ offsets}

\cref{fig:multihead-visualization} presents the occupancy predictions over time, reverse flow predictions,
attention offsets,
and cross attention weights of \ourmodel{} with $K = 4$ across four different scenes in AV2. The reverse flow predictions, attention offsets, and cross attention weights are all visualized at the final timestep in the prediction horizon.
For clarification on the different flow visualizations used throughout main manuscript and supplementary, please see \cref{fig:flow-types}.
The cross-attention weights are those used in the weighted sum of the $K$ interpolated feature vectors
in the feature aggregation module.

First, we notice that the offsets learn to look mainly backward and forward in the direction of the lanes.
Second, we find that the decoder pays more attention to those offsets that look backwards along the lanes.
This is particularly apparent in Offset 1; it is weighted heavily in the lane regions where it points
against the flow of traffic. This aligns with our intuition that the offset sampling mechanism helps the model attend to the
where occupancy was in the past --- where the LiDAR evidence is --- in order to predict the future.

We hypothesize that \ourmodel{} with $K = 4$ does not outperform \ourmodel{} with $K = 1$, because, assuming the offsets
look backwards in time and space, $K = 1$ can already model multi-modal futures for occupancy,
in the same way that the \reverse{}-flow formulation does. Then, multiple offset predictions 
would only provide more information in areas where occupancy could have come from multiple directions in the past,
like the intersection in the Scene 3 of \cref{fig:multihead-visualization}. Indeed, focusing in on this example
in \cref{fig:side-by-side}, we see that \ourmodel{} with $K = 4$ is qualitatively able to predict future occupancy
from both the left turning car (\textbf{A}) and the cars entering the intersection (\textbf{B}), as highlighted in the blue box,
while \ourmodel{} with $K = 1$ only predicts occupancy from \textbf{B}.
\input{figures/side-by-side.tex}
However, these examples are rare in the training and evaluation data, and $K > 1$ adds noise in cases
where it is not needed, explaining the slightly worse metrics of $K = 4$ relative
to $K = 1$.
\input{figures/flow_types.tex}

\subsection{Measuring Occupancy Predictions Over Time}

This section presents an analysis of the occupancy prediction accuracy as a function of prediction time $\Delta t$.
We want to investigate (1) how do the occupancy metrics of the \object-free baselines from the literature and
\ourmodel{} evolve over time, and (2) how well do the explicit models (those that
predict occupancy on a spatio-temporal grid) predict occupancy at timesteps that are not aligned with the grid.
\begin{figure*}
    \centering
    \begin{subfigure}[b]{0.49\textwidth}
        \centering
        \input{figures/mAP-vs-time-hwysim.tex}
        \label{fig:mAP-vs-time-hwysim}
    \end{subfigure}
    \hfill
    \begin{subfigure}[b]{0.49\textwidth}
        \centering
        \input{figures/iou-vs-time-hwysim.tex}
        \label{fig:iou-vs-time-hwysim}
    \end{subfigure}
    \caption{mAP and Soft IoU as functions of prediction time for \ourmodel{} and various \object{}-free models from the literature on HwySim.}
    \label{fig:metrics-over-time}
\end{figure*}

\begin{figure*}
    \centering
    \begin{subfigure}[b]{0.49\textwidth}
        \centering
        \input{figures/mAP-vs-time-argo.tex}
        \label{fig:mAP-vs-time-argo}
    \end{subfigure}
    \hfill
    \begin{subfigure}[b]{0.49\textwidth}
        \centering
        \input{figures/iou-vs-time-argo.tex}
        \label{fig:iou-vs-time-argo}
    \end{subfigure}
    \caption{mAP and Soft IoU as functions of prediction time for various \object{}-free models from the literature and
    \ourmodel{} on AV2.}
    \label{fig:metrics-over-time-argo}
\end{figure*}

Expanding on (2): Due to memory constraints,
all explicit models (e.g., MP3 and \textsc{OccFlow}) output an occupancy map with temporal dimension
$(5 / 0.5) + 1 = 11$, where the $+1$ comes from the prediction at the initial observation time ($\Delta t = 0$).
However, typically motion planners may need higher temporal resolution 
query points, because, for example, a highway vehicle can travel \SI{15}{m} in \SI{0.5}{s}.
\ourmodel{} by-passes this problem because it is trained on continuous query points at no additional cost.
Thus, in this section we evaluate all models at a time discretization of 0.1s (the period of LiDAR the sweeps). For the
explicit models, this requires linear interpolation on the predicted occupancy map.

The results for HwySim are presented in \cref{fig:metrics-over-time}.
The first trend to note is that while all models have a similar mAP and Soft-IoU at the initial timestep,
\ourmodel{} has a higher mAP and Soft-IoU at later timesteps. We attribute this to its expressive attention offset mechanism,
which increases the effective receptive field, allowing accurate occupancy predictions far away from the LiDAR evidence
information in $\mathbf{Z}$.
Secondly, we observe that at prediction times that require interpolation (e.g., those not at $\{0, 0.5, \dots, 5\} s$),
the explicit models have worse mAP and Soft-IoU than \ourmodel. This is because this interpolation inherently
fails to model how occupancy moves; e.g., interpolation at a spatio-temporal point between grid timesteps
can result in no predicted occupancy even if occupancy had passed through that point along its trajectory.
\ourmodel{} does not suffer from this problem because it can produce predictions at any continuous time. We observe this
in the smooth mAP and Soft-IoU curves over time of \ourmodel.
We notice that while MP3 has a more accurate occupancy ranking (mAP) over time than \textsc{OccFlow}, it
has a worse Soft-IoU. The better occupancy ranking is attributed to the flow-warping mechanism of MP3,
which imposes a realistic prior on the evolution of occupancy, and increases the effective receptive field.
However, as described by Mahjourian et al. \cite{mahjourian2022occupancy}, MP3 suffers from
lost occupancy over time, as reflected in its poor Soft-IoU. 

We also include occupancy metrics over time on AV2, presented in \cref{fig:metrics-over-time-argo}.
While the conclusions are similar here, they are less obvious, likely due to the
fact that the vehicles move a smaller proportion of the ROI than in HwySim on average,
making interpolation more accurate.
To illustrate the advantages of \ourmodel{} we compare it qualitatively against baselines on a single 
scene from AV2 in \cref{fig:unrolled-occupancy}. \textsc{GoRela}
predicts many agents entering the intersection at once, which is unrealistic. \textsc{OccFlow}
has a detection with incorrect size and orientation of a vehicle at $\Delta t = \SI{0.0}{s}$, and suffers
from spreading / inaccurate occupancy predictions at later timesteps, likely due to the limited
receptive field of its convolutional decoder. MP3 has disjoint-pixel occupancy predictions,
which is caused by its forward-flow warping mechanism and was first observed by
Mahjourian et al. \cite{mahjourian2022occupancy}.
In contrast, \ourmodel{} exhibits accurate occupancy predictions and realistic multi-modality
(vehicle going straight or turning left, and a lane-change). %

\subsection{Spatial Resolution Evaluation}

In this experiment, we investigate how the performance of an explicit
\object-free model is affected by its output grid resolution. We train three MP3 models,
each with a different output grid resolution; \SI{0.2}{m}, \SI{1.0}{m}, and \SI{2.0}{m}.
The rasterized input to the encoder is kept the same,
and the models are evaluated with a spatial resolution of \SI{0.2}{m} by bi-linearly interpolating
the output grid. The results are presented in \cref{tab:spatial-resolution}, along with the
metrics of \ourmodel{} for comparison.
We notice that at an output resolution of \SI{2.0}{m}, the performance of MP3 is much worse
than with a \SI{0.2}{m} output resolution. However, \SI{1.0}{m} is only slightly worse.
This is likely because the ground-truth labels are rectangular bounding boxes, so as long
as the discretization resolution is less than the smallest bounding box dimension,
bi-linear interpolation on the output grid will be accurate. Most vehicles have a smallest
side length greater than \SI{1.0}{m}, but less than \SI{2.0}{m}, explaining the results.

\begin{table*}
    \small
    \centering
    \begin{threeparttable}
    \setlength\tabcolsep{4pt} %
    \begin{tabularx}{0.6\textwidth}{c c c c c c c}
    \toprule
                              & mAP \ua        & Soft-IoU \ua   & ECE \da         & EPE \da            & \multicolumn{2}{c}{Flow Grounded}   \\
    \cmidrule{6-7}                                                                                                                
    \multicolumn{1}{c}{}      &                &                &                 &                    & mAP \ua        & Soft-IoU \ua   \\
    \midrule
    MP3 (\SI{2.0}{m})         & 0.635          & 0.286          & 0.731           & 0.547              & 0.986          & 0.280          \\
    MP3 (\SI{1.0}{m})         & 0.769          & 0.416          & 0.307           & 0.458              & 0.927          & 0.460          \\
    MP3 (\SI{0.2}{m})         & 0.774          & 0.422          & 0.201           & 0.472              & 0.902          & 0.466          \\
    \ourmodel{}               & \textbf{0.799} & \textbf{0.480} & \textbf{0.193}  & \textbf{0.267}     & \textbf{0.936} & \textbf{0.597} \\ 
    \bottomrule
    \end{tabularx}
    \end{threeparttable}
    \caption{Comparing MP3 \cite{casas2021mp3} with different spatial output resolutions, evaluated at a spatial
    resolution of \SI{0.2}{m} on AV2. The results for \ourmodel{} are presented for reference.}
    \label{tab:spatial-resolution}
\end{table*}

\input{figures/unrolled_occupancy.tex}

\input{figures/reliability_diagram.tex}

\subsection{Analyzing Occupancy Calibration with Reliability Diagrams}
\label{sec:reliability-diagrams}

\cref{fig:argo-reliability-plots} displays the reliability plots for all baselines from the literature on AV2. 
All the methods in the top row are \object-based, while those in the bottom row are \object-free.
We notice that the \object-based methods are poorly calibrated on the task of occupancy prediction
relative to the \object-free methods. \Object-based methods predict discrete multi-modal 
trajectories which poorly capture the uncertainty in occupancy, and result in overconfidence.
On the contrary, \object-free methods can capture non-parametric occupancy distributions without
thresholding to produce instances, resulting in better calibration.

\subsection{Query Point Training Procedure}

As described in Sec.~3.3, we train \ourmodel{} with a batch of continuous query points $\mathcal{Q}$ sampled uniformly from the future spatio-temporal volume.
This section ablates this querying procedure by training \ourmodel{} on AV2 by querying points on a regular spatio-temporal grid
with 0.5 s temporal resolution and varying spatial resolutions. 
Our evaluation procedure is the same as described in Sec. 4, and our results are presented in \cref{tab:resolution-ablation}.
We observe a slight disadvantage of training with points from a regular grid with respect to our proposed method.
\begin{table*}
  \centering
  \begin{threeparttable}
  \begin{tabularx}{0.5\textwidth}{l c c c c}
  \toprule
     & \multicolumn{4}{c}{Spatial Resolution} \\
          & 0.2 m & 0.5 m  & 1.25 m & 2.0 m \\[-0.2em]
  \midrule
  mAP \ua     & $-0.19\%$ & $-0.25\%$ & $-0.83\%$ & $-0.90\%$ \\
  Soft-IoU \ua & $-0.46\%$ & $-1.86\%$ & $-1.66\%$ & $-2.24\%$ \\[-0.2em]
  \bottomrule
  \end{tabularx}
  \end{threeparttable}
    \caption{Comparing the performance of \ourmodel{} when trained with query points sampled from a regular spatio-temporal grid
    against the proposed continuous query point training procedure.
    We report the relative change in evaluation metrics.}
    \label{tab:resolution-ablation}
\end{table*}

\subsection{LiDAR History Ablation}

In this section we measure the effect of LiDAR input frequency and history duration
on the performance of \ourmodel{}.
First, we highlight that the LiDAR frequency and history duration as well as encoder architecture
were held constant for all models so changes in performance due to different LiDAR inputs are expected to be similar across all methods.
When we train  \ourmodel{} with 10 past LiDAR frames at 10 Hz and see a minor relative improvement of
$0.24\%$ in mAP and $0.98\%$ in Soft-IoU.
For reference, MP3 improves similarly ($0.45\%$ mAP and $0.22\%$ Soft-IoU).

\subsection{Inferring Agent Identity}

In this section, we provide a proof of concept for inferring agent identity (ID) with appropriate post-processing, similar to \cite{hu2021fiery,mahjourian2022occupancy}.
Inferring agent ID requires some form of instance prediction (like the ``centerness'' field from \cite{hu2021fiery}, or
object detections in \textsc{OccFlow} \cite{mahjourian2022occupancy}). 

Without any additional training required, we pair \ourmodel{} $K = 1$ with an object detector
(the same one used by the \object{}-based baselines).
Then, for any query point $\mathbf{q}$, we add the attention offset $\mathbf{\Delta q}$ to find the reference point $\mathbf{r} = \mathbf{q} + \Delta \mathbf{q}$.
Finally, we assign the ID of occupancy at $\mathbf{q}$ to that of detection with the closest centroid to $\mathbf{r}$.
Recall that despite the attention offset $\mathbf{\Delta q}$ being unsupervised, we found  $\mathbf{r}$ to point to the occupancy at the current time $\Delta t = 0$ (i.e., where the LiDAR evidence for that object is; see L721-737, Fig. 4 in the main paper).

\begin{figure*}
    \centering
    \begin{tikzpicture}
        \node[inner sep=0pt, outer sep=0, anchor=north, draw=black, line width=2pt] (zero)        at (0,0)    {
            \includegraphics[width=0.88\textwidth,trim={0cm, 3cm, 1cm, 0cm}, clip]{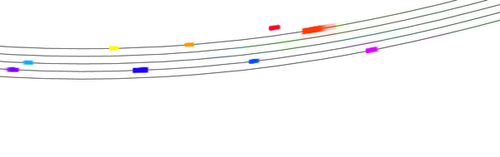}
        };
        \node[inner sep=0pt, outer sep=0, anchor=north west, draw=black, line width=2pt] (five)        at (zero.south west)    {
            \includegraphics[width=0.88\textwidth, trim={0cm, 3cm, 1cm, 0cm}, clip]{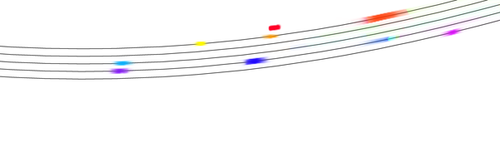}
        };
        \node[inner sep=0pt, outer sep=0, anchor=north west, draw=black, line width=2pt] (ten)        at (five.south west)    {
            \includegraphics[width=0.88\textwidth, trim={0cm, 3cm, 1cm, 0cm}, clip]{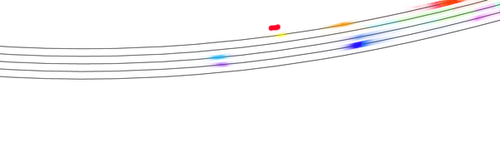}
        };
        \node[anchor = north west] at (zero.north west) {$\Delta t = 0.0\; s$};
        \node[anchor = north west] at (five.north west) {$\Delta t = 2.5\; s$};
        \node[anchor = north west] at (ten.north west) {$\Delta t = 5.0\; s$};
    \end{tikzpicture}
    \caption{Qualitative examples of \ourmodel{} inferring agent ID.} \label{fig:ID}
\end{figure*}

\Cref{fig:ID} visualizes the predicted occupancy colored by inferred instance ID (i.e., its associated detection).
This simple approach illustrates that agent ID can be preserved in our implicit occupancy-flow method,
and opens the door to future work on this area to improve this heuristic.

%% file: figures/multihead_qualitative_v2.tex
\begin{figure*}
    \centering
    \begin{tikzpicture}
        \node[inner sep = 0pt, outer sep = 0pt, anchor = north, draw = black, line width = \lw] (104-45-gt) at (0,0) {\includegraphics[width=\w,trim={0cm, 2cm, 0cm, 0cm}, clip]{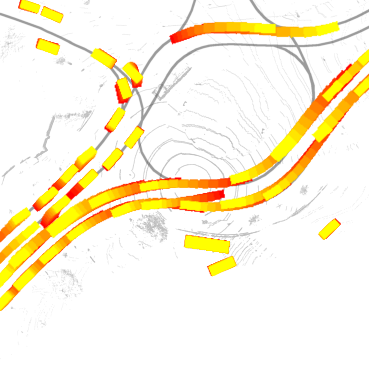}};
        \node[inner sep = 0pt, outer sep = 0pt, anchor = west, draw = black, line width = \lw] (104-45-preds) at (104-45-gt.east) {\includegraphics[width=\w,trim={0cm, 2cm, 0cm, 0cm}, clip]{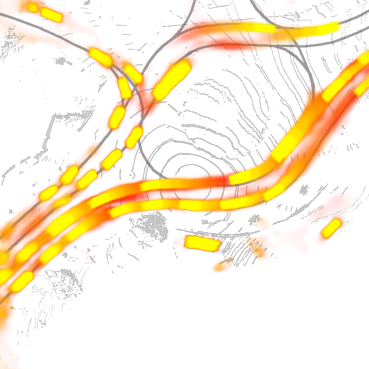}};

        \node[inner sep = 0pt, outer sep = 0pt, anchor = south west, draw = black, line width = \lw, xshift=5pt] (104-45-h1-off) at (104-45-preds.south east) {\includegraphics[width=\w,trim={0cm, 2cm, 0cm, 0cm}, clip]{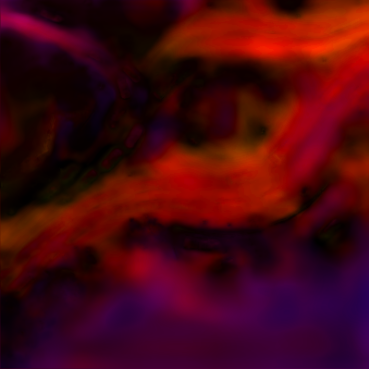}};
        \node[inner sep = 0pt, outer sep = 0pt, anchor = north west, draw = black, line width = \lw, xshift=5pt] (104-45-h1-cross) at (104-45-preds.south east) {\includegraphics[width=\w,trim={0cm, 2cm, 0cm, 0cm}, clip]{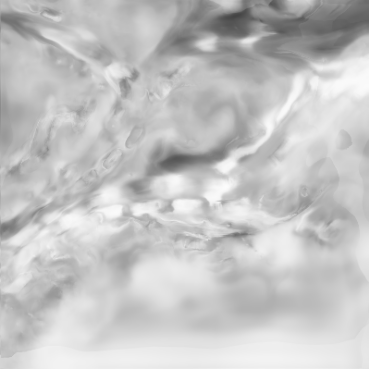}};

        \node[inner sep=0pt, outer sep=0, anchor=north west, line width=1.5pt] at (104-45-h1-off.north west)    {
            \includegraphics[width = 0.03\linewidth]{figures/raw-images/argo_color_wheel.png}
        };
        \node[anchor=south west] at (104-45-h1-off.south west) {
            \textcolor{white}{\scriptsize{Offsets}}
        };
        \node[anchor=south west] at (104-45-h1-cross.south west) {
            \scriptsize{Cross Attention Weights}
        };

        \node[inner sep = 0pt, outer sep = 0pt, anchor = west, draw = black, line width = \lw] (104-45-h2-off) at (104-45-h1-off.east) {\includegraphics[width=\w,trim={0cm, 2cm, 0cm, 0cm}, clip]{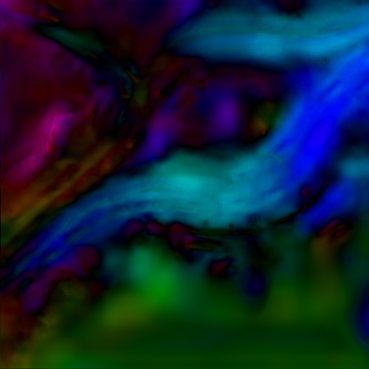}};
        \node[inner sep = 0pt, outer sep = 0pt, anchor = west, draw = black, line width = \lw] (104-45-h2-cross) at (104-45-h1-cross.east) {\includegraphics[width=\w,trim={0cm, 2cm, 0cm, 0cm}, clip]{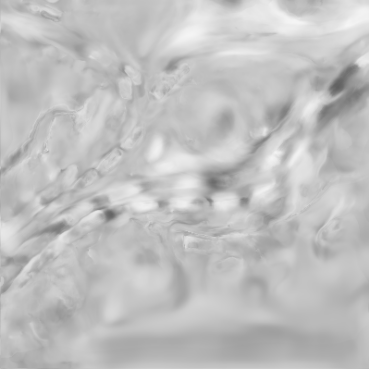}};

        \node[inner sep = 0pt, outer sep = 0pt, anchor = west, draw = black, line width = \lw] (104-45-h3-off) at (104-45-h2-off.east) {\includegraphics[width=\w,trim={0cm, 2cm, 0cm, 0cm}, clip]{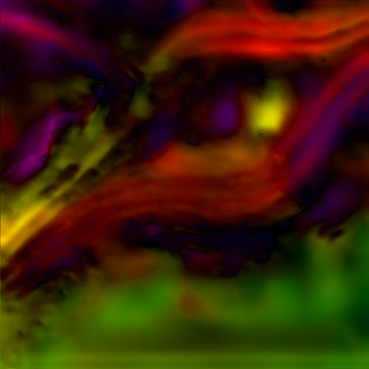}};
        \node[inner sep = 0pt, outer sep = 0pt, anchor = west, draw = black, line width = \lw] (104-45-h3-cross) at (104-45-h2-cross.east) {\includegraphics[width=\w,trim={0cm, 2cm, 0cm, 0cm}, clip]{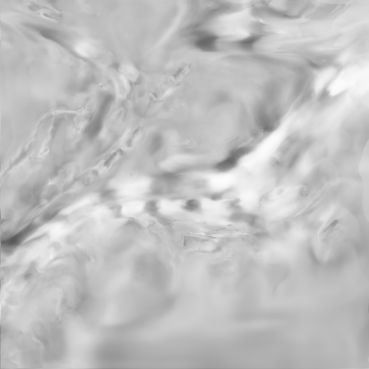}};

        \node[inner sep = 0pt, outer sep = 0pt, anchor = west, draw = black, line width = \lw] (104-45-h4-off) at (104-45-h3-off.east) {\includegraphics[width=\w,trim={0cm, 2cm, 0cm, 0cm}, clip]{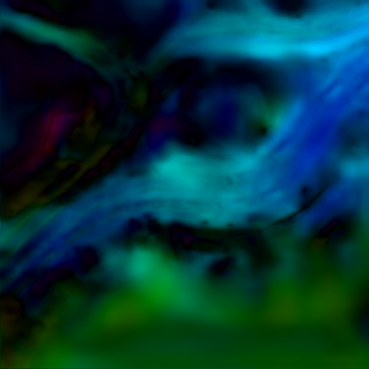}};
        \node[inner sep = 0pt, outer sep = 0pt, anchor = west, draw = black, line width = \lw] (104-45-h4-cross) at (104-45-h3-cross.east) {\includegraphics[width=\w,trim={0cm, 2cm, 0cm, 0cm}, clip]{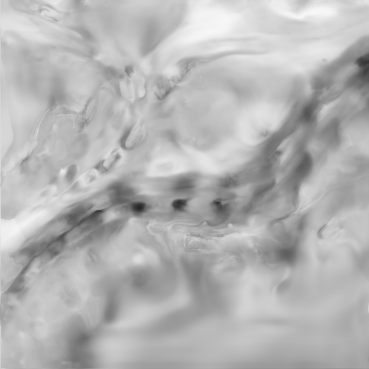}};

        \node[anchor = south] at (104-45-h1-off.north -| 104-45-gt.north) {Ground Truth};
        \node[anchor = south] at (104-45-h1-off.north -| 104-45-preds.north) {Predictions};
        \node[anchor = south] at (104-45-h1-off.north) {Offset 1};
        \node[anchor = south] at (104-45-h2-off.north) {Offset 2};
        \node[anchor = south] at (104-45-h3-off.north) {Offset 3};
        \node[anchor = south] at (104-45-h4-off.north) {Offset 4};

        \node[anchor = south, rotate=90] at (104-45-gt.south west) {Scene 1};

        \node[inner sep = 0pt, outer sep = 0pt, anchor = north, draw = black, line width = \lw] (104-45-motion-preds) at (104-45-preds.south) {\includegraphics[width=\w,trim={0cm, 2cm, 0cm, 0cm}, clip]{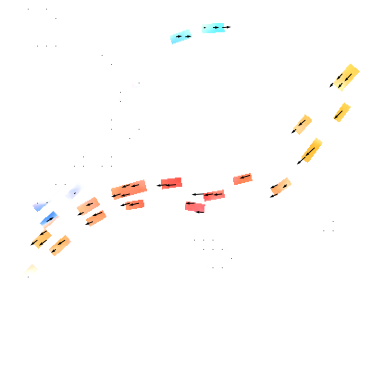}};
        \node[inner sep = 0pt, outer sep = 0pt, anchor = north, draw = black, line width = \lw] (104-45-motion-gt) at (104-45-gt.south) {\includegraphics[width=\w,trim={0cm, 2cm, 0cm, 0cm}, clip]{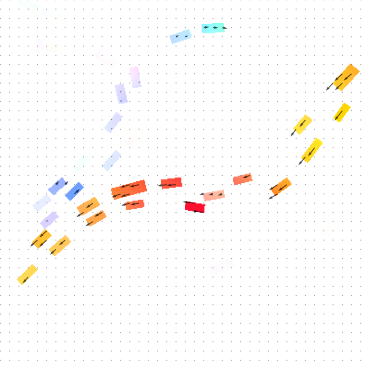}};

        \node[inner sep=0pt, outer sep=0, anchor=north west, line width=1.5pt] at (104-45-motion-gt.north west)    {
                \includegraphics[width = 0.03\linewidth]{figures/raw-images/argo_color_wheel_inverted.png}
        };

        \node[inner sep = 0pt, outer sep = 0pt, anchor = north, draw = black, line width = \lw, yshift=-5pt] (31-45-h1-off) at (104-45-h1-cross.south) {\includegraphics[width=\w,trim={0cm, 2cm, 0cm, 0cm}, clip]{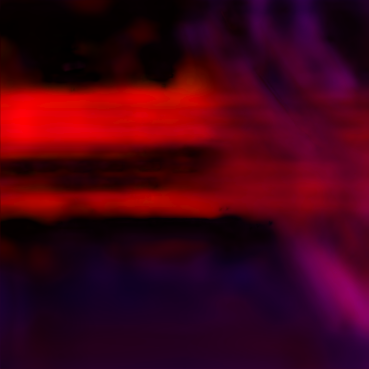}};
        \node[inner sep = 0pt, outer sep = 0pt, anchor = north, draw = black, line width = \lw] (31-45-h1-cross) at (31-45-h1-off.south) {\includegraphics[width=\w,trim={0cm, 2cm, 0cm, 0cm}, clip]{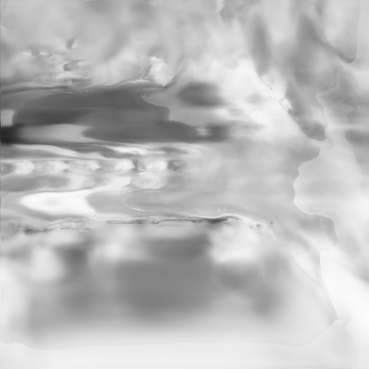}};

        \node[inner sep = 0pt, outer sep = 0pt, anchor = west, draw = black, line width = \lw] (31-45-h2-off) at (31-45-h1-off.east) {\includegraphics[width=\w,trim={0cm, 2cm, 0cm, 0cm}, clip]{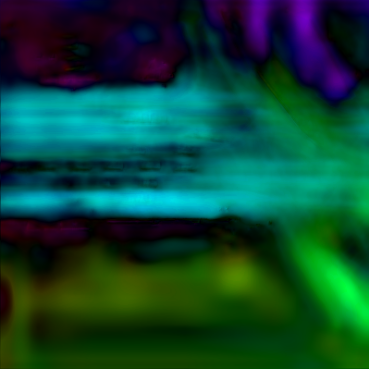}};
        \node[inner sep = 0pt, outer sep = 0pt, anchor = west, draw = black, line width = \lw] (31-45-h2-cross) at (31-45-h1-cross.east) {\includegraphics[width=\w,trim={0cm, 2cm, 0cm, 0cm}, clip]{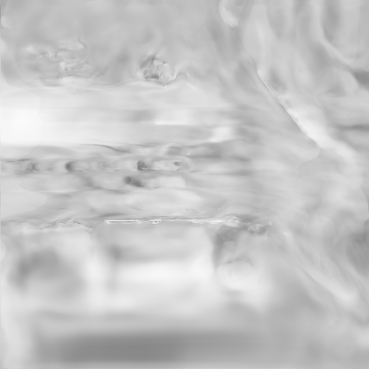}};

        \node[inner sep = 0pt, outer sep = 0pt, anchor = west, draw = black, line width = \lw] (31-45-h3-off) at (31-45-h2-off.east) {\includegraphics[width=\w,trim={0cm, 2cm, 0cm, 0cm}, clip]{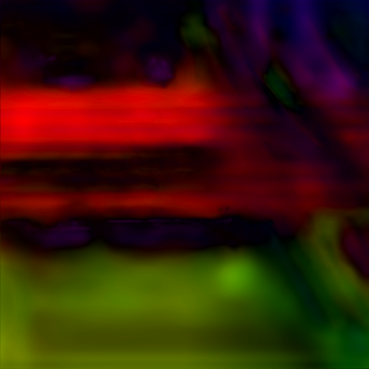}};
        \node[inner sep = 0pt, outer sep = 0pt, anchor = west, draw = black, line width = \lw] (31-45-h3-cross) at (31-45-h2-cross.east) {\includegraphics[width=\w,trim={0cm, 2cm, 0cm, 0cm}, clip]{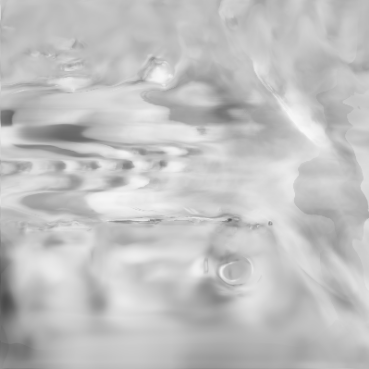}};

        \node[inner sep = 0pt, outer sep = 0pt, anchor = west, draw = black, line width = \lw] (31-45-h4-off) at (31-45-h3-off.east) {\includegraphics[width=\w,trim={0cm, 2cm, 0cm, 0cm}, clip]{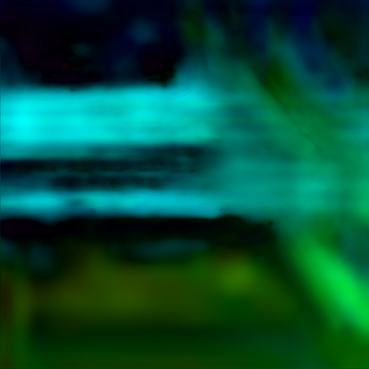}};
        \node[inner sep = 0pt, outer sep = 0pt, anchor = west, draw = black, line width = \lw] (31-45-h4-cross) at (31-45-h3-cross.east) {\includegraphics[width=\w,trim={0cm, 2cm, 0cm, 0cm}, clip]{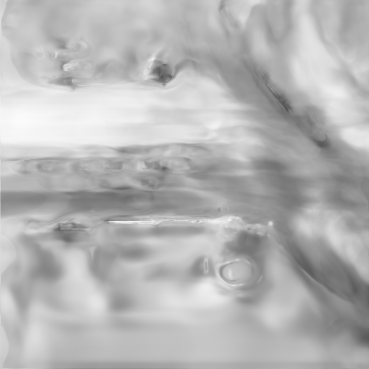}};

        \node[inner sep = 0pt, outer sep = 0pt, anchor = east, draw = black, line width = \lw, xshift=-5pt] (31-45-preds) at (31-45-h1-off.west) {\includegraphics[width=\w,trim={0cm, 2cm, 0cm, 0cm}, clip]{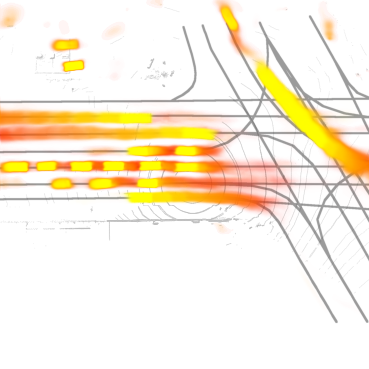}};
        \node[inner sep = 0pt, outer sep = 0pt, anchor = east, draw = black, line width = \lw] (31-45-gt) at (31-45-preds.west) {\includegraphics[width=\w,trim={0cm, 2cm, 0cm, 0cm}, clip]{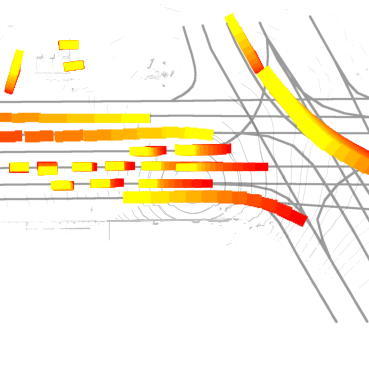}};

        \node[anchor = south, rotate=90] at (31-45-gt.south west) {Scene 2};

        \node[inner sep = 0pt, outer sep = 0pt, anchor = north, draw = black, line width = \lw] (31-45-motion-preds) at (31-45-preds.south) {\includegraphics[width=\w,trim={0cm, 2cm, 0cm, 0cm}, clip]{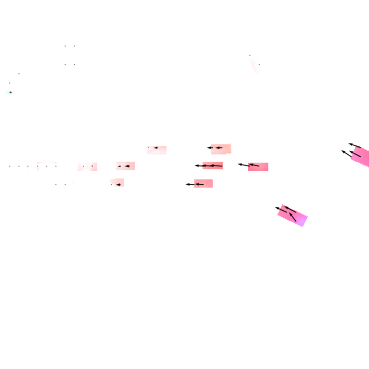}};
        \node[inner sep = 0pt, outer sep = 0pt, anchor = north, draw = black, line width = \lw] (31-45-motion-gt) at (31-45-gt.south) {\includegraphics[width=\w,trim={0cm, 2cm, 0cm, 0cm}, clip]{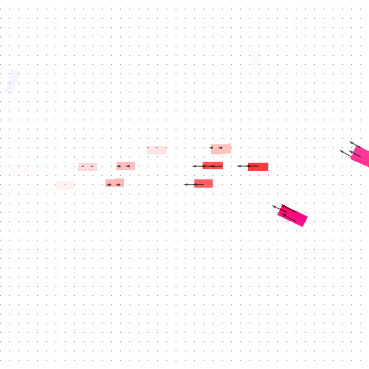}};

        \node[inner sep = 0pt, outer sep = 0pt, anchor = north, draw = black, line width = \lw, yshift=-5pt] (24-75-h1-off) at (31-45-h1-cross.south) {\includegraphics[width=\w,trim={0cm, 2cm, 0cm, 0cm}, clip]{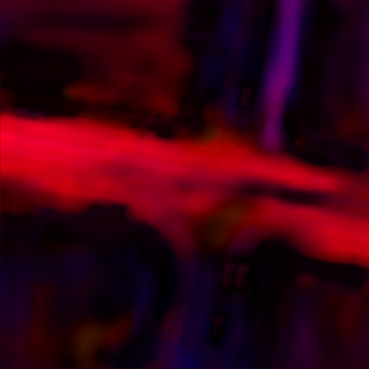}};
        \node[inner sep = 0pt, outer sep = 0pt, anchor = north, draw = black, line width = \lw] (24-75-h1-cross) at (24-75-h1-off.south) {\includegraphics[width=\w,trim={0cm, 2cm, 0cm, 0cm}, clip]{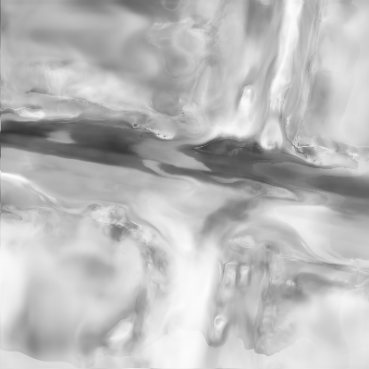}};

        \node[inner sep = 0pt, outer sep = 0pt, anchor = west, draw = black, line width = \lw] (24-75-h2-off) at (24-75-h1-off.east) {\includegraphics[width=\w,trim={0cm, 2cm, 0cm, 0cm}, clip]{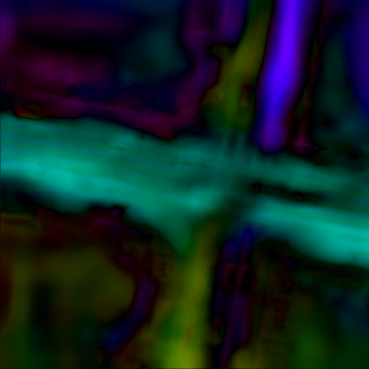}};
        \node[inner sep = 0pt, outer sep = 0pt, anchor = west, draw = black, line width = \lw] (24-75-h2-cross) at (24-75-h1-cross.east) {\includegraphics[width=\w,trim={0cm, 2cm, 0cm, 0cm}, clip]{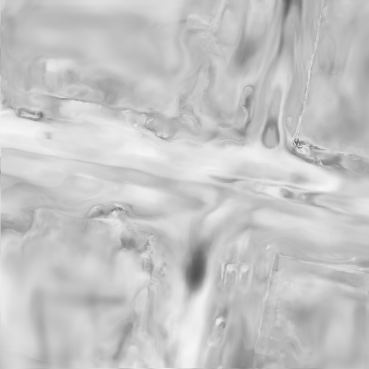}};

        \node[inner sep = 0pt, outer sep = 0pt, anchor = west, draw = black, line width = \lw] (24-75-h3-off) at (24-75-h2-off.east) {\includegraphics[width=\w,trim={0cm, 2cm, 0cm, 0cm}, clip]{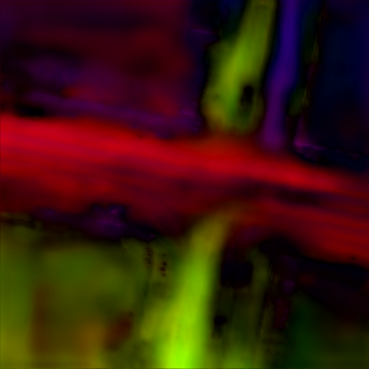}};
        \node[inner sep = 0pt, outer sep = 0pt, anchor = west, draw = black, line width = \lw] (24-75-h3-cross) at (24-75-h2-cross.east) {\includegraphics[width=\w,trim={0cm, 2cm, 0cm, 0cm}, clip]{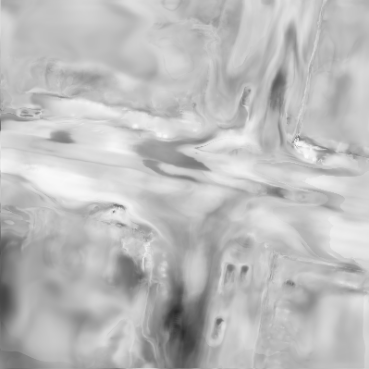}};

        \node[inner sep = 0pt, outer sep = 0pt, anchor = west, draw = black, line width = \lw] (24-75-h4-off) at (24-75-h3-off.east) {\includegraphics[width=\w,trim={0cm, 2cm, 0cm, 0cm}, clip]{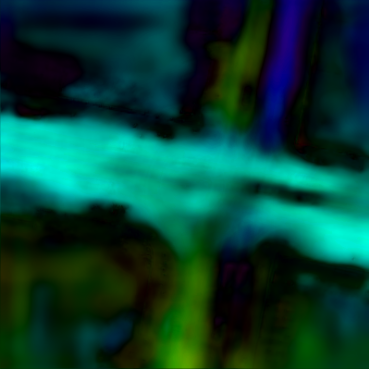}};
        \node[inner sep = 0pt, outer sep = 0pt, anchor = west, draw = black, line width = \lw] (24-75-h4-cross) at (24-75-h3-cross.east) {\includegraphics[width=\w,trim={0cm, 2cm, 0cm, 0cm}, clip]{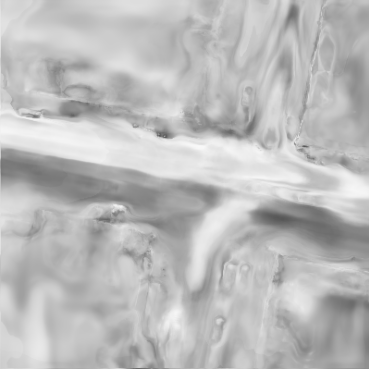}};

        \node[inner sep = 0pt, outer sep = 0pt, anchor = east, draw = black, line width = \lw, xshift=-5pt] (24-75-preds) at (24-75-h1-off.west) {\includegraphics[width=\w,trim={0cm, 2cm, 0cm, 0cm}, clip]{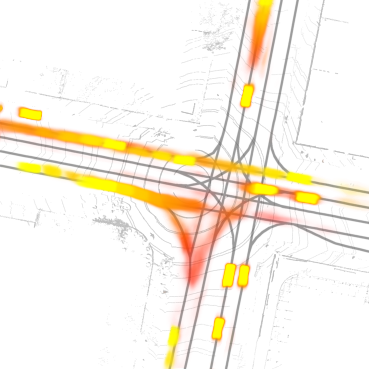}};
        \node[inner sep = 0pt, outer sep = 0pt, anchor = east, draw = black, line width = \lw] (24-75-gt) at (24-75-preds.west) {\includegraphics[width=\w,trim={0cm, 2cm, 0cm, 0cm}, clip]{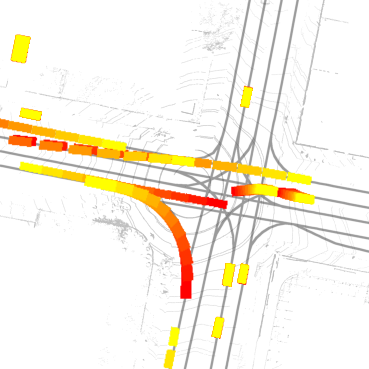}};

        \node[inner sep = 0pt, outer sep = 0pt, anchor = north, draw = black, line width = \lw] (24-75-motion-preds) at (24-75-preds.south) {\includegraphics[width=\w,trim={0cm, 2cm, 0cm, 0cm}, clip]{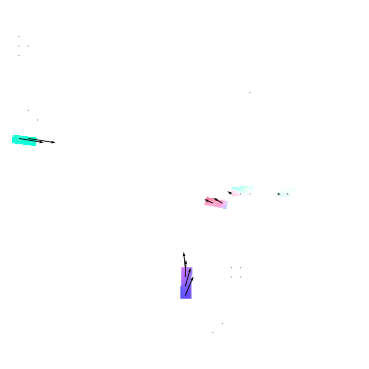}};
        \node[inner sep = 0pt, outer sep = 0pt, anchor = north, draw = black, line width = \lw] (24-75-motion-gt) at (24-75-gt.south) {\includegraphics[width=\w,trim={0cm, 2cm, 0cm, 0cm}, clip]{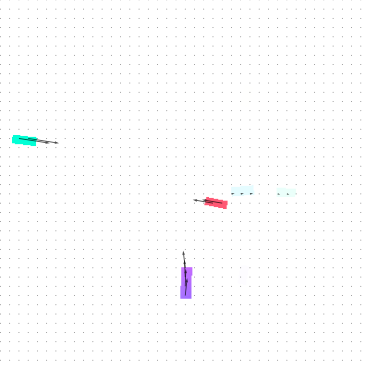}};

        \node[anchor = south, rotate=90] at (24-75-gt.south west) {Scene 3};

        \node[inner sep = 0pt, outer sep = 0pt, anchor = north, draw = black, line width = \lw, yshift=-5pt] (14-105-h1-off) at (24-75-h1-cross.south) {\includegraphics[width=\w,trim={0cm, 2cm, 0cm, 0cm}, clip]{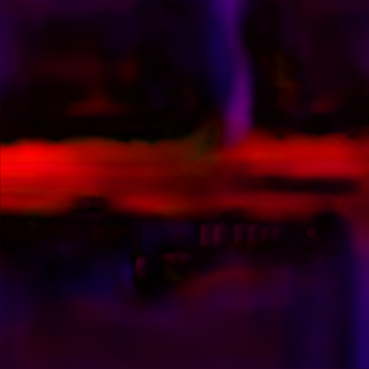}};
        \node[inner sep = 0pt, outer sep = 0pt, anchor = north, draw = black, line width = \lw] (14-105-h1-cross) at (14-105-h1-off.south) {\includegraphics[width=\w,trim={0cm, 2cm, 0cm, 0cm}, clip]{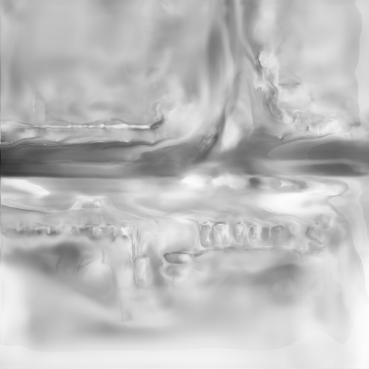}};

        \node[inner sep = 0pt, outer sep = 0pt, anchor = west, draw = black, line width = \lw] (14-105-h2-off) at (14-105-h1-off.east) {\includegraphics[width=\w,trim={0cm, 2cm, 0cm, 0cm}, clip]{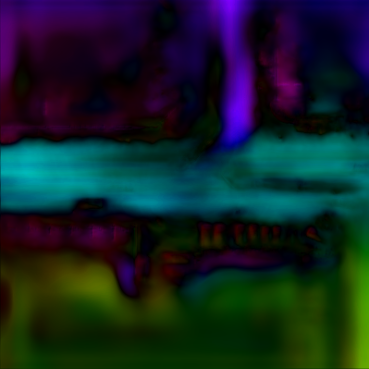}};
        \node[inner sep = 0pt, outer sep = 0pt, anchor = west, draw = black, line width = \lw] (14-105-h2-cross) at (14-105-h1-cross.east) {\includegraphics[width=\w,trim={0cm, 2cm, 0cm, 0cm}, clip]{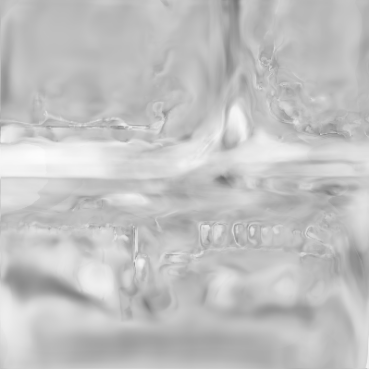}};

        \node[inner sep = 0pt, outer sep = 0pt, anchor = west, draw = black, line width = \lw] (14-105-h3-off) at (14-105-h2-off.east) {\includegraphics[width=\w,trim={0cm, 2cm, 0cm, 0cm}, clip]{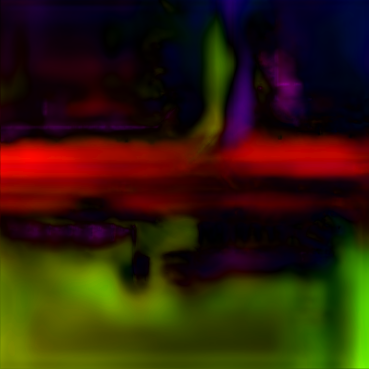}};
        \node[inner sep = 0pt, outer sep = 0pt, anchor = west, draw = black, line width = \lw] (14-105-h3-cross) at (14-105-h2-cross.east) {\includegraphics[width=\w,trim={0cm, 2cm, 0cm, 0cm}, clip]{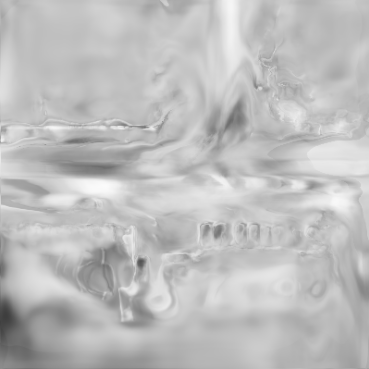}};

        \node[inner sep = 0pt, outer sep = 0pt, anchor = west, draw = black, line width = \lw] (14-105-h4-off) at (14-105-h3-off.east) {\includegraphics[width=\w,trim={0cm, 2cm, 0cm, 0cm}, clip]{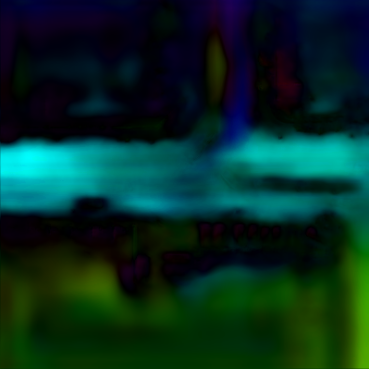}};
        \node[inner sep = 0pt, outer sep = 0pt, anchor = west, draw = black, line width = \lw] (14-105-h4-cross) at (14-105-h3-cross.east) {\includegraphics[width=\w,trim={0cm, 2cm, 0cm, 0cm}, clip]{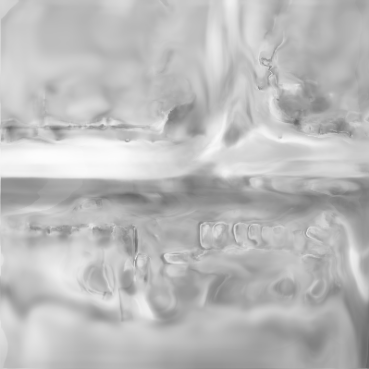}};

        \node[inner sep = 0pt, outer sep = 0pt, anchor = east, draw = black, line width = \lw, xshift=-5pt] (14-105-preds) at (14-105-h1-off.west) {\includegraphics[width=\w,trim={0cm, 2cm, 0cm, 0cm}, clip]{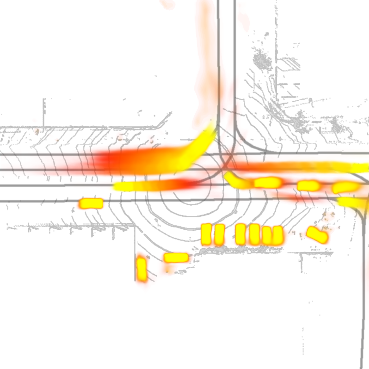}};
        \node[inner sep = 0pt, outer sep = 0pt, anchor = east, draw = black, line width = \lw] (14-105-gt) at (14-105-preds.west) {\includegraphics[width=\w,trim={0cm, 2cm, 0cm, 0cm}, clip]{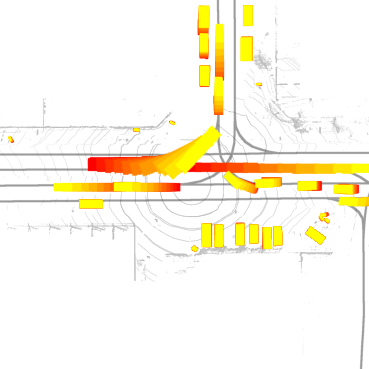}};

        \node[inner sep = 0pt, outer sep = 0pt, anchor = north, draw = black, line width = \lw] (14-105-motion-preds) at (14-105-preds.south) {\includegraphics[width=\w,trim={0cm, 2cm, 0cm, 0cm}, clip]{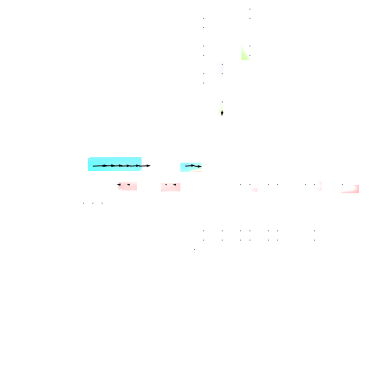}};
        \node[inner sep = 0pt, outer sep = 0pt, anchor = north, draw = black, line width = \lw] (14-105-motion-gt) at (14-105-gt.south) {\includegraphics[width=\w,trim={0cm, 2cm, 0cm, 0cm}, clip]{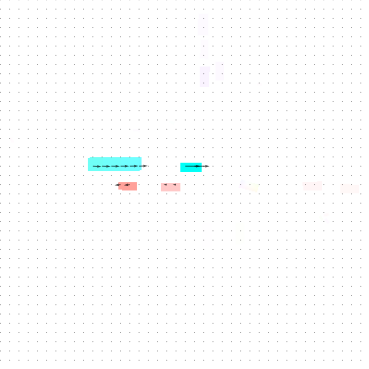}};

        \node[anchor = south, rotate=90] at (14-105-gt.south west) {Scene 4};

        \node[anchor= west, inner sep=0pt, outer sep=0pt,xshift=5pt,yshift=-2.5pt] (off-legend) at (31-45-h4-cross.south east) {\includegraphics[width=0.025\textwidth, trim={20pt, 0pt, 20pt, 0pt}, clip]{figures/Occupancy-Legend.pdf}};
        \node[anchor=west, inner sep=0pt, outer sep=0pt] at (off-legend.east) {\includegraphics[width=0.025\textwidth, trim={20pt, 0pt, 20pt, 0pt}, clip]{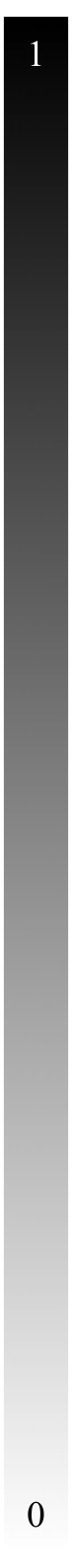}};

    \end{tikzpicture}
    \caption{Qualitative results of $\ourmodel{}$ with $K = 4$ on AV2. 
    The grey scale images display the cross attention weights, which sum to 1 across the columns.
    The HSV color images are the offset predictions. 
    The predicted reverse flow field is masked by the ground truth occupancy for visual comparison (see \cref{fig:flow-types}).
    The reverse flow, attention offsets, and cross attention weights are all visualized at the final timestep.
    }
    \vspace{-10pt}
    \label{fig:multihead-visualization}
\end{figure*}

%% file: figures/side-by-side.tex
\begin{figure*}
    \centering
    \begin{tikzpicture}
        \node[inner sep = 0pt, outer sep = 0pt, anchor = east, draw = black, line width = \lw] (gt-scene) at (0, 0) {\includegraphics[width=0.3\textwidth, trim={1cm, 1cm, 1cm, 1cm}, clip]{figures/raw-images/argo_implicit_deformable_multihead_no_query_vis_0.2/24/75/occ_gt_24_75_False.png}};
        \node[inner sep=0pt, outer sep=0, anchor=west, draw=black, line width=\lw] (singlehead-scene) at (gt-scene.east) {\includegraphics[width=0.3\textwidth, trim={1cm, 1cm, 1cm, 1cm}, clip]{figures/raw-images/argo_implicit_deformable_continuous_vis_0.2/24/75/occ_preds_24_75_False.png}};
        \node[inner sep = 0pt, outer sep = 0pt, anchor = west, draw = black, line width = \lw] (multihead-scene) at (singlehead-scene.east) {\includegraphics[width=0.3\textwidth, trim={1cm, 1cm, 1cm, 1cm}, clip]{figures/raw-images/argo_implicit_deformable_multihead_no_query_vis_0.2/24/75/occ_preds_24_75_False_ones.png}};

        \node[anchor = south] at (gt-scene.north) {Ground Truth};
        \node[anchor = south] at (singlehead-scene.north) {$K = 1$};
        \node[anchor = south] at (multihead-scene.north) {$K = 4$};

        \node[xshift=0.08\textwidth, yshift=-0.005\textwidth, draw, line width = \lw] at (gt-scene.center) {\textbf{A}};
        \node[xshift=-0.005\textwidth, yshift=-0.02\textwidth, draw, line width = \lw] at (gt-scene.center) {\textbf{B}};
        \node[draw = blue, line width=2pt, minimum width=0.09\textwidth, minimum height=0.11\textwidth, yshift=-0.05\textwidth, xshift=0.05\textwidth] at (multihead-scene) {};

    \end{tikzpicture}
    \caption{A side by side comparison of \ourmodel{} with $K = 1$ and $K = 4$. The blue box highlights that
    $K = 4$ is qualitatively better able to predict occupancy in the intersection that could have come from
    two past locations, namely \textbf{A} and \textbf{B}.}
    \vspace{-10pt}
    \label{fig:side-by-side}
\end{figure*}

%% file: figures/flow_types.tex
\begin{figure*}
    \centering
    \begin{tikzpicture}
        \node[inner sep = 0pt, outer sep = 0pt, anchor = north, draw = black, line width = \lw] (gt) at (0,0) {\includegraphics[width=0.24\textwidth,trim={0cm, 2cm, 0cm, 0cm}, clip]{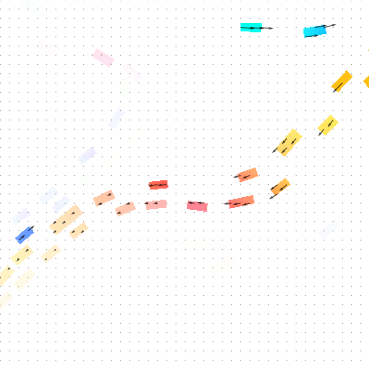}};
        \node[anchor = north] at (gt.south) {(a)};
        \node[inner sep = 0pt, outer sep = 0pt, anchor = west, draw = black, line width = \lw] (field) at (gt.east) {\includegraphics[width=0.24\textwidth,trim={0cm, 2cm, 0cm, 0cm}, clip]{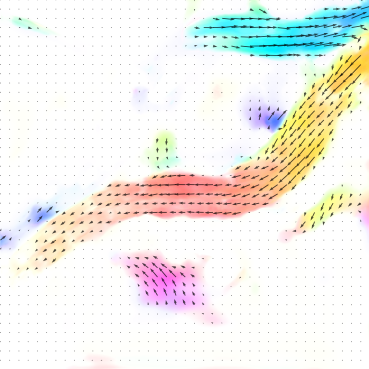}};
        \node[anchor = north] at (field.south) {(b)};
        \node[inner sep = 0pt, outer sep = 0pt, anchor = west, draw = black, line width = \lw] (pred-mask) at (field.east) {\includegraphics[width=0.24\textwidth,trim={0cm, 2cm, 0cm, 0cm}, clip]{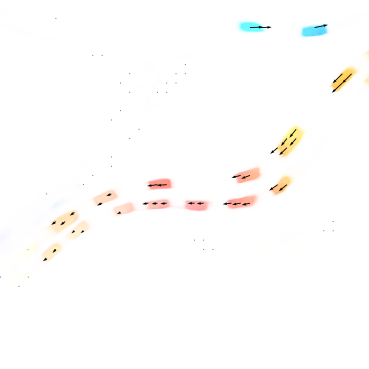}};
        \node[anchor = north] at (pred-mask.south) {(c)};
        \node[inner sep = 0pt, outer sep = 0pt, anchor = west, draw = black, line width = \lw] (gt-mask) at (pred-mask.east) {\includegraphics[width=0.24\textwidth,trim={0cm, 2cm, 0cm, 0cm}, clip]{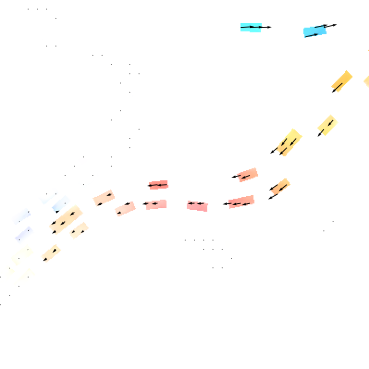}};
        \node[anchor = north] at (gt-mask.south) {(d)};
        \node[inner sep=0pt, outer sep=0, anchor=north west, line width=1.5pt] at (gt.north west)    {
                \includegraphics[width = 0.04\linewidth]{figures/raw-images/argo_color_wheel_inverted.png}
        };
    \end{tikzpicture}
    \caption{Visualizations of the reverse flow predictions of \ourmodel{} with $K = 4$ at the initial timestep.
    (a) is the ground truth flow field, (b) is the predicted flow field, (c) is the predicted flow field
    masked by the occupancy predictions, and (d) is the predicted flow field masked by the ground truth occupancy.}
    \vspace{-10pt}
    \label{fig:flow-types}
\end{figure*}

%% file: figures/mAP-vs-time-hwysim.tex
\begin{tikzpicture}

\definecolor{darkgray176}{RGB}{176,176,176}
\definecolor{lightgray}{RGB}{211,211,211}
\definecolor{lightgray204}{RGB}{204,204,204}

\begin{axis}[
tick align=outside,
tick pos=left,
x grid style={darkgray176},
xlabel={Prediction Time $\Delta t$ (s)},
xmin=-0.25, xmax=5.25,
xtick style={color=black},
y grid style={darkgray176},
ylabel={mAP},
ymin=0.254588831050063, ymax=0.946260562695295,
ytick style={color=black},
height = 6cm,
width=\textwidth,
]
\addplot [semithick, lightgray, dashed, forget plot]
table {%
0 0.254588831050063
0 0.946260562695295
};
\addplot [semithick, lightgray, dashed, forget plot]
table {%
0.5 0.254588831050063
0.5 0.946260562695295
};
\addplot [semithick, lightgray, dashed, forget plot]
table {%
1 0.254588831050063
1 0.946260562695295
};
\addplot [semithick, lightgray, dashed, forget plot]
table {%
1.5 0.254588831050063
1.5 0.946260562695295
};
\addplot [semithick, lightgray, dashed, forget plot]
table {%
2 0.254588831050063
2 0.946260562695295
};
\addplot [semithick, lightgray, dashed, forget plot]
table {%
2.5 0.254588831050063
2.5 0.946260562695295
};
\addplot [semithick, lightgray, dashed, forget plot]
table {%
3 0.254588831050063
3 0.946260562695295
};
\addplot [semithick, lightgray, dashed, forget plot]
table {%
3.5 0.254588831050063
3.5 0.946260562695295
};
\addplot [semithick, lightgray, dashed, forget plot]
table {%
4 0.254588831050063
4 0.946260562695295
};
\addplot [semithick, lightgray, dashed, forget plot]
table {%
4.5 0.254588831050063
4.5 0.946260562695295
};
\addplot [semithick, lightgray, dashed, forget plot]
table {%
5 0.254588831050063
5 0.946260562695295
};
\addplot [thick, MP3Color]
table {%
0 0.897633725909115
0.1 0.774939840564571
0.2 0.633018915829957
0.3 0.632463544254934
0.4 0.765818147001237
0.5 0.855260832473598
0.6 0.755333173166961
0.7 0.662894097220119
0.8 0.662278674912525
0.9 0.755476941556021
1 0.810675026826965
1.1 0.730525293772313
1.2 0.655383637336611
1.3 0.658385691609601
1.4 0.729686927933935
1.5 0.746727271102892
1.6 0.690090093096764
1.7 0.638253623469481
1.8 0.641757205836877
1.9 0.687559282130924
2 0.687567965559309
2.1 0.6534384607839
2.2 0.620062139978257
2.3 0.619336845281827
2.4 0.645681798258587
2.5 0.641158733015445
2.6 0.616662741870165
2.7 0.591813852809916
2.8 0.588302497837043
2.9 0.596574797494663
3 0.587872614289226
3.1 0.572793527530048
3.2 0.555303457065328
3.3 0.547014622450627
3.4 0.546463954604021
3.5 0.534209636276885
3.6 0.520423799607672
3.7 0.506879850577875
3.8 0.498928277715088
3.9 0.493934659516349
4 0.481038066080557
4.1 0.474147404902317
4.2 0.464297786919481
4.3 0.456683514083424
4.4 0.448693587439331
4.5 0.433181883980882
4.6 0.432294180535016
4.7 0.426361404347828
4.8 0.419289276225174
4.9 0.408279757576634
5 0.391866668192543
};
\addplot [thick, OccFlowColor]
table {%
0 0.85976412504213
0.1 0.6754243353383
0.2 0.540651903534471
0.3 0.558509212006409
0.4 0.716505771013441
0.5 0.843964687493307
0.6 0.71707386270984
0.7 0.592313571559868
0.8 0.607121112055799
0.9 0.730075491109736
1 0.80899598752586
1.1 0.719177994718868
1.2 0.619290422690163
1.3 0.617835234204623
1.4 0.707651353151364
1.5 0.755049481139674
1.6 0.684220392979183
1.7 0.61058159538225
1.8 0.61019343363547
1.9 0.668299945969452
2 0.68479653789977
2.1 0.629583065340485
2.2 0.579895398558061
2.3 0.578037132163973
2.4 0.605063660032419
2.5 0.603657173390156
2.6 0.565965891546323
2.7 0.534757280905695
2.8 0.53095295615005
2.9 0.536973645372086
3 0.528591675743991
3.1 0.504031402612382
3.2 0.485495873375945
3.3 0.480125315187127
3.4 0.478976935695415
3.5 0.467133812913876
3.6 0.450911576614325
3.7 0.435484547693066
3.8 0.425250748887645
3.9 0.415855070256088
4 0.401986168292021
4.1 0.393276716044902
4.2 0.382867398013666
4.3 0.372002512825527
4.4 0.358984004414594
4.5 0.342995818800224
4.6 0.335820220259961
4.7 0.325276406610099
4.8 0.313803374591064
4.9 0.30165510335194
5 0.286028455215755
};
\addplot [thick, ourmodelcolor1]
table {%
0 0.900873463331218
0.1 0.901974100219513
0.2 0.901618066876254
0.3 0.900293748541198
0.4 0.899124628573954
0.5 0.896596285397336
0.6 0.893291431984605
0.7 0.888684949704829
0.8 0.88337145686341
0.9 0.877302862877701
1 0.872005217506158
1.1 0.866667566010056
1.2 0.858701026429512
1.3 0.850341273580757
1.4 0.842300724258904
1.5 0.832826845577157
1.6 0.824489564653542
1.7 0.815888710071347
1.8 0.806194290949714
1.9 0.796561058939124
2 0.785716539070301
2.1 0.774684118227083
2.2 0.763420206952974
2.3 0.751965787186307
2.4 0.741489231682127
2.5 0.730453381280746
2.6 0.720168518682088
2.7 0.709665217263791
2.8 0.699524799769474
2.9 0.687749400149899
3 0.677411942930251
3.1 0.665637796792868
3.2 0.655637818487147
3.3 0.644802109254497
3.4 0.634652829502346
3.5 0.623455293947478
3.6 0.61238328659733
3.7 0.601839709241791
3.8 0.59035354399449
3.9 0.58136546831924
4 0.573411772886602
4.1 0.563515028734132
4.2 0.554301614208539
4.3 0.544180775771844
4.4 0.535709452981061
4.5 0.526491373748538
4.6 0.517847361896384
4.7 0.509121361310027
4.8 0.500600362696702
4.9 0.49131607854198
5 0.481665408502285
};
\end{axis}

\end{tikzpicture}

%% file: figures/iou-vs-time-hwysim.tex
\begin{tikzpicture}

\definecolor{darkgray176}{RGB}{176,176,176}
\definecolor{lightgray}{RGB}{211,211,211}
\definecolor{lightgray204}{RGB}{204,204,204}

\begin{axis}[
legend cell align={left},
legend style={
  fill opacity=0.8,
  draw opacity=1,
  text opacity=1,
  at={(0.97,0.97)},
  anchor=north east,
  draw=lightgray204
},
tick align=outside,
tick pos=left,
x grid style={darkgray176},
xlabel={Prediction Time $\Delta t$ (s)},
xmin=-0.25, xmax=5.25,
xtick style={color=black},
y grid style={darkgray176},
ylabel={SoftIoU},
ymin=0.0623408801853657, ymax=0.670607855170965,
ytick style={color=black},
height = 6cm,
width=\textwidth,
]
\addplot [semithick, lightgray, dashed, forget plot]
table {%
0 0.0623408801853657
0 0.670607855170965
};
\addplot [semithick, lightgray, dashed, forget plot]
table {%
0.5 0.0623408801853657
0.5 0.670607855170965
};
\addplot [semithick, lightgray, dashed, forget plot]
table {%
1 0.0623408801853657
1 0.670607855170965
};
\addplot [semithick, lightgray, dashed, forget plot]
table {%
1.5 0.0623408801853657
1.5 0.670607855170965
};
\addplot [semithick, lightgray, dashed, forget plot]
table {%
2 0.0623408801853657
2 0.670607855170965
};
\addplot [semithick, lightgray, dashed, forget plot]
table {%
2.5 0.0623408801853657
2.5 0.670607855170965
};
\addplot [semithick, lightgray, dashed, forget plot]
table {%
3 0.0623408801853657
3 0.670607855170965
};
\addplot [semithick, lightgray, dashed, forget plot]
table {%
3.5 0.0623408801853657
3.5 0.670607855170965
};
\addplot [semithick, lightgray, dashed, forget plot]
table {%
4 0.0623408801853657
4 0.670607855170965
};
\addplot [semithick, lightgray, dashed, forget plot]
table {%
4.5 0.0623408801853657
4.5 0.670607855170965
};
\addplot [semithick, lightgray, dashed, forget plot]
table {%
5 0.0623408801853657
5 0.670607855170965
};
\addplot [thick, MP3Color]
table {%
0 0.577877044677734
0.1 0.468126952648163
0.2 0.338367730379105
0.3 0.293609112501144
0.4 0.340863764286041
0.5 0.443093836307526
0.6 0.323060721158981
0.7 0.24230369925499
0.8 0.226160869002342
0.9 0.269008666276932
1 0.356964468955994
1.1 0.261286467313766
1.2 0.205218568444252
1.3 0.195461079478264
1.4 0.22621288895607
1.5 0.286530017852783
1.6 0.215684100985527
1.7 0.17796379327774
1.8 0.172136068344116
1.9 0.193614438176155
2 0.236501783132553
2.1 0.186213761568069
2.2 0.158989995718002
2.3 0.15320847928524
2.4 0.1666000187397
2.5 0.197270110249519
2.6 0.161942824721336
2.7 0.142710998654366
2.8 0.13828432559967
2.9 0.146340325474739
3 0.166982725262642
3.1 0.140714943408966
3.2 0.126611903309822
3.3 0.122581698000431
3.4 0.127712652087212
3.5 0.141794353723526
3.6 0.122894912958145
3.7 0.112864509224892
3.8 0.109963409602642
3.9 0.113249957561493
4 0.122552946209908
4.1 0.108631551265717
4.2 0.101159773766994
4.3 0.0988550558686256
4.4 0.100960075855255
4.5 0.107309095561504
4.6 0.0971479937434196
4.7 0.0916902869939804
4.8 0.0899893790483475
4.9 0.0912547409534454
5 0.095550075173378
};
\addlegendentry{MP3}
\addplot [thick, OccFlowColor]
table {%
0 0.623418867588043
0.1 0.447589844465256
0.2 0.287091672420502
0.3 0.296424865722656
0.4 0.442231863737106
0.5 0.582831978797913
0.6 0.443301886320114
0.7 0.319662690162659
0.8 0.311833709478378
0.9 0.417645335197449
1 0.526436686515808
1.1 0.420984297990799
1.2 0.32353088259697
1.3 0.309539914131165
1.4 0.381665885448456
1.5 0.464520663022995
1.6 0.379804164171219
1.7 0.307683080434799
1.8 0.298620492219925
1.9 0.352223992347717
2 0.407736957073212
2.1 0.340164870023727
2.2 0.290061563253403
2.3 0.286293298006058
2.4 0.321412056684494
2.5 0.354595243930817
2.6 0.305118650197983
2.7 0.269321143627167
2.8 0.265697121620178
2.9 0.28679096698761
3 0.307135134935379
3.1 0.271199554204941
3.2 0.244800165295601
3.3 0.240614786744118
3.4 0.254510939121246
3.5 0.267239630222321
3.6 0.240915417671204
3.7 0.219586104154587
3.8 0.212613493204117
3.9 0.218054056167603
4 0.225501209497452
4.1 0.206757202744484
4.2 0.191149950027466
4.3 0.182656526565552
4.4 0.181398332118988
4.5 0.18345433473587
4.6 0.16913078725338
4.7 0.155765652656555
4.8 0.1465023458004
4.9 0.142115131020546
5 0.14061051607132
};
\addlegendentry{\textsc{OccFlow}}
\addplot [thick, ourmodelcolor1]
table {%
0 0.6319779753685
0.1 0.625171422958374
0.2 0.619730353355408
0.3 0.615367293357849
0.4 0.611974954605103
0.5 0.606936931610107
0.6 0.600633442401886
0.7 0.592729926109314
0.8 0.583236813545227
0.9 0.573406219482422
1 0.564374566078186
1.1 0.55495673418045
1.2 0.543358743190765
1.3 0.531654357910156
1.4 0.520941138267517
1.5 0.509771585464478
1.6 0.500081837177277
1.7 0.490450382232666
1.8 0.480491280555725
1.9 0.470621407032013
2 0.4603191614151
2.1 0.450050532817841
2.2 0.439705699682236
2.3 0.429124772548676
2.4 0.418867468833923
2.5 0.408780574798584
2.6 0.39954873919487
2.7 0.390466302633286
2.8 0.381611049175262
2.9 0.372336566448212
3 0.3638576567173
3.1 0.355039685964584
3.2 0.347451061010361
3.3 0.339365005493164
3.4 0.331694006919861
3.5 0.323392421007156
3.6 0.315021157264709
3.7 0.306887626647949
3.8 0.298688948154449
3.9 0.291661381721497
4 0.285150825977325
4.1 0.278012871742249
4.2 0.271609514951706
4.3 0.264960825443268
4.4 0.258657664060593
4.5 0.251934379339218
4.6 0.245330259203911
4.7 0.238926887512207
4.8 0.232772037386894
4.9 0.226381585001945
5 0.219980508089066
};
\addlegendentry{\ourmodel}
\end{axis}

\end{tikzpicture}

%% file: figures/mAP-vs-time-argo.tex
\begin{tikzpicture}

\definecolor{darkgray176}{RGB}{176,176,176}
\definecolor{darkorange25512714}{RGB}{255,127,14}
\definecolor{forestgreen4416044}{RGB}{44,160,44}
\definecolor{lightgray}{RGB}{211,211,211}
\definecolor{lightgray204}{RGB}{204,204,204}
\definecolor{steelblue31119180}{RGB}{31,119,180}

\begin{axis}[
tick align=outside,
tick pos=left,
x grid style={darkgray176},
xlabel={Prediction Time $\Delta t$ (s)},
xmin=-0.25, xmax=5.25,
xtick style={color=black},
y grid style={darkgray176},
ylabel={mAP},
ymin=0.608142470803419, ymax=0.910851442708268,
ytick style={color=black},
height=6cm,
width=\textwidth,
]
\addplot [semithick, lightgray, dashed, forget plot]
table {%
0 0.608142470803419
0 0.910851442708268
};
\addplot [semithick, lightgray, dashed, forget plot]
table {%
0.5 0.608142470803419
0.5 0.910851442708268
};
\addplot [semithick, lightgray, dashed, forget plot]
table {%
1 0.608142470803419
1 0.910851442708268
};
\addplot [semithick, lightgray, dashed, forget plot]
table {%
1.5 0.608142470803419
1.5 0.910851442708268
};
\addplot [semithick, lightgray, dashed, forget plot]
table {%
2 0.608142470803419
2 0.910851442708268
};
\addplot [semithick, lightgray, dashed, forget plot]
table {%
2.5 0.608142470803419
2.5 0.910851442708268
};
\addplot [semithick, lightgray, dashed, forget plot]
table {%
3 0.608142470803419
3 0.910851442708268
};
\addplot [semithick, lightgray, dashed, forget plot]
table {%
3.5 0.608142470803419
3.5 0.910851442708268
};
\addplot [semithick, lightgray, dashed, forget plot]
table {%
4 0.608142470803419
4 0.910851442708268
};
\addplot [semithick, lightgray, dashed, forget plot]
table {%
4.5 0.608142470803419
4.5 0.910851442708268
};
\addplot [semithick, lightgray, dashed, forget plot]
table {%
5 0.608142470803419
5 0.910851442708268
};
\addplot [thick, MP3Color]
table {%
0 0.89709194398532
0.1 0.886797646318206
0.2 0.8714041368196
0.3 0.864891986528169
0.4 0.868967184916662
0.5 0.873314005625535
0.6 0.859787143895557
0.7 0.846170642097696
0.8 0.841683907249191
0.9 0.84460249714065
1 0.848445688500141
1.1 0.839637474514992
1.2 0.827729059539963
1.3 0.814794079305763
1.4 0.797376079253536
1.5 0.768955498857062
1.6 0.776104945095104
1.7 0.779714103703729
1.8 0.781975314931854
1.9 0.783148211826908
2 0.781278509894123
2.1 0.784569978179724
2.2 0.784198163514656
2.3 0.782148065845705
2.4 0.778604426357071
2.5 0.772333967320471
2.6 0.766200032969735
2.7 0.759286254536815
2.8 0.752684785228252
2.9 0.745920916220318
3 0.73804196406924
3.1 0.741318706323701
3.2 0.743034092714612
3.3 0.743560686933125
3.4 0.743141423519161
3.5 0.740956130435648
3.6 0.740729695247854
3.7 0.739193100609576
3.8 0.73715725352373
3.9 0.734618888953167
4 0.731089367325808
4.1 0.729442197673483
4.2 0.727930990373356
4.3 0.726403854648791
4.4 0.724727058289587
4.5 0.722726134776355
4.6 0.720965666206782
4.7 0.719158883169231
4.8 0.71770887031676
4.9 0.716020448197822
5 0.714063953831184
};
\addplot [thick, OccFlowColor]
table {%
0 0.809223207674215
0.1 0.788289948133855
0.2 0.766328535776591
0.3 0.759535737418128
0.4 0.769461119611558
0.5 0.780189138484958
0.6 0.768938678075198
0.7 0.753599805414862
0.8 0.746094882968662
0.9 0.746489184014412
1 0.746028617013852
1.1 0.739073356430362
1.2 0.730074995336922
1.3 0.72303973202777
1.4 0.719143918732337
1.5 0.715617608766221
1.6 0.710945148898785
1.7 0.705296440645419
1.8 0.699966437353875
1.9 0.695240058418283
2 0.690924060481543
2.1 0.686632755363086
2.2 0.681940333816359
2.3 0.677736969891637
2.4 0.674177938736405
2.5 0.670809860050769
2.6 0.667004073740221
2.7 0.66354271925276
2.8 0.660686171390624
2.9 0.657592215686608
3 0.654691139322946
3.1 0.651776069763704
3.2 0.649033339771595
3.3 0.646431837708401
3.4 0.644453743860632
3.5 0.642527906844451
3.6 0.640964533718767
3.7 0.639321432578456
3.8 0.637783340280576
3.9 0.636183663497738
4 0.634765958153131
4.1 0.633073240446643
4.2 0.631862663033501
4.3 0.630359449150489
4.4 0.628906012235719
4.5 0.627370191294997
4.6 0.625907112612943
4.7 0.624574544617664
4.8 0.623670954223629
4.9 0.62282029071656
5 0.621901969526366
};
\addplot [thick, ourmodelcolor1]
table {%
0 0.890567565340808
0.1 0.892704877471571
0.2 0.891340995721867
0.3 0.888633604552931
0.4 0.885997268966368
0.5 0.883263750293559
0.6 0.880101044734443
0.7 0.876336039626596
0.8 0.872436634157041
0.9 0.868411799052319
1 0.863982154991284
1.1 0.859120609516313
1.2 0.854121941264755
1.3 0.848946437111271
1.4 0.843527033992179
1.5 0.838045938746064
1.6 0.832834137166938
1.7 0.827699714257943
1.8 0.822689088203073
1.9 0.817781478032613
2 0.813015683330469
2.1 0.808173304022595
2.2 0.803017300112417
2.3 0.798212845853443
2.4 0.793728419947866
2.5 0.789528625066271
2.6 0.78531799797525
2.7 0.781200292579528
2.8 0.777467000765085
2.9 0.773649886030809
3 0.770061751658002
3.1 0.766442315894482
3.2 0.76282039366133
3.3 0.759227441120166
3.4 0.756099272072001
3.5 0.753066425694252
3.6 0.749891890960221
3.7 0.746754262180196
3.8 0.743815470818993
3.9 0.741145169560769
4 0.738834516377645
4.1 0.736485942025845
4.2 0.734403104186863
4.3 0.732280298423934
4.4 0.729937482447336
4.5 0.727538689331685
4.6 0.724764307391949
4.7 0.722188522367888
4.8 0.72015292050545
4.9 0.718034654592052
5 0.715926435334935
};
\end{axis}

\end{tikzpicture}

%% file: figures/iou-vs-time-argo.tex
\begin{tikzpicture}

\definecolor{darkgray176}{RGB}{176,176,176}
\definecolor{darkorange25512714}{RGB}{255,127,14}
\definecolor{forestgreen4416044}{RGB}{44,160,44}
\definecolor{lightgray}{RGB}{211,211,211}
\definecolor{lightgray204}{RGB}{204,204,204}
\definecolor{steelblue31119180}{RGB}{31,119,180}

\begin{axis}[
legend cell align={left},
legend style={fill opacity=0.8, draw opacity=1, text opacity=1, draw=lightgray204},
tick align=outside,
tick pos=left,
x grid style={darkgray176},
xlabel={Prediction Time $\Delta t$ (s)},
xmin=-0.25, xmax=5.25,
xtick style={color=black},
y grid style={darkgray176},
ylabel={SoftIoU},
ymin=0.288826937973499, ymax=0.621733222901821,
ytick style={color=black},
height =6cm,
width=\textwidth,
]
\addplot [semithick, lightgray, dashed, forget plot]
table {%
0 0.288826937973499
0 0.621733222901821
};
\addplot [semithick, lightgray, dashed, forget plot]
table {%
0.5 0.288826937973499
0.5 0.621733222901821
};
\addplot [semithick, lightgray, dashed, forget plot]
table {%
1 0.288826937973499
1 0.621733222901821
};
\addplot [semithick, lightgray, dashed, forget plot]
table {%
1.5 0.288826937973499
1.5 0.621733222901821
};
\addplot [semithick, lightgray, dashed, forget plot]
table {%
2 0.288826937973499
2 0.621733222901821
};
\addplot [semithick, lightgray, dashed, forget plot]
table {%
2.5 0.288826937973499
2.5 0.621733222901821
};
\addplot [semithick, lightgray, dashed, forget plot]
table {%
3 0.288826937973499
3 0.621733222901821
};
\addplot [semithick, lightgray, dashed, forget plot]
table {%
3.5 0.288826937973499
3.5 0.621733222901821
};
\addplot [semithick, lightgray, dashed, forget plot]
table {%
4 0.288826937973499
4 0.621733222901821
};
\addplot [semithick, lightgray, dashed, forget plot]
table {%
4.5 0.288826937973499
4.5 0.621733222901821
};
\addplot [semithick, lightgray, dashed, forget plot]
table {%
5 0.288826937973499
5 0.621733222901821
};
\addplot [thick, MP3Color]
table {%
0 0.595241367816925
0.1 0.577364802360535
0.2 0.550729811191559
0.3 0.53683865070343
0.4 0.537189245223999
0.5 0.541719079017639
0.6 0.520493924617767
0.7 0.504448533058167
0.8 0.498372256755829
0.9 0.498851031064987
1 0.503180027008057
1.1 0.485730886459351
1.2 0.468827933073044
1.3 0.451655089855194
1.4 0.43261444568634
1.5 0.411881864070892
1.6 0.410098373889923
1.7 0.409662902355194
1.8 0.409781932830811
1.9 0.409743517637253
2 0.409017562866211
2.1 0.409394830465317
2.2 0.408620268106461
2.3 0.406653046607971
2.4 0.403094947338104
2.5 0.398069262504578
2.6 0.394303381443024
2.7 0.390514194965363
2.8 0.386827707290649
2.9 0.382872104644775
3 0.378711014986038
3.1 0.379959136247635
3.2 0.380791991949081
3.3 0.381269246339798
3.4 0.381217032670975
3.5 0.380368858575821
3.6 0.380081593990326
3.7 0.378940105438232
3.8 0.377231806516647
3.9 0.374805837869644
4 0.371653437614441
4.1 0.371270179748535
4.2 0.370839357376099
4.3 0.370131254196167
4.4 0.369145482778549
4.5 0.367866843938828
4.6 0.367087125778198
4.7 0.366242408752441
4.8 0.365488171577454
4.9 0.364545911550522
5 0.363373100757599
};
\addlegendentry{MP3}
\addplot [semithick, OccFlowColor]
table {%
0 0.500388622283936
0.1 0.473141014575958
0.2 0.443321973085403
0.3 0.431949645280838
0.4 0.441626340150833
0.5 0.45632266998291
0.6 0.439038336277008
0.7 0.418749362230301
0.8 0.407855331897736
0.9 0.4080471098423
1 0.411756902933121
1.1 0.400683790445328
1.2 0.389346808195114
1.3 0.381665617227554
1.4 0.378337562084198
1.5 0.377336293458939
1.6 0.371231436729431
1.7 0.365212559700012
1.8 0.36033308506012
1.9 0.35657125711441
2 0.353662401437759
2.1 0.349575102329254
2.2 0.345706850290298
2.3 0.342403531074524
2.4 0.339604258537292
2.5 0.337086915969849
2.6 0.334189683198929
2.7 0.331573039293289
2.8 0.329245537519455
2.9 0.326810926198959
3 0.324422985315323
3.1 0.322607040405273
3.2 0.32090699672699
3.3 0.319271564483643
3.4 0.317910999059677
3.5 0.316542685031891
3.6 0.315563410520554
3.7 0.314494788646698
3.8 0.313481509685516
3.9 0.312401980161667
4 0.311369329690933
4.1 0.310405611991882
4.2 0.309633135795593
4.3 0.308760732412338
4.4 0.307870924472809
4.5 0.306967675685883
4.6 0.306235790252686
4.7 0.30554923415184
4.8 0.305002212524414
4.9 0.304505944252014
5 0.303959041833878
};
\addlegendentry{\textsc{OccFlow}}
\addplot [thick, ourmodelcolor1]
table {%
0 0.606560885906219
0.1 0.606601119041443
0.2 0.602710008621216
0.3 0.597226202487946
0.4 0.59190708398819
0.5 0.58650267124176
0.6 0.580664217472076
0.7 0.574295282363892
0.8 0.567817032337189
0.9 0.561437249183655
1 0.554613471031189
1.1 0.547288477420807
1.2 0.539730727672577
1.3 0.532030284404755
1.4 0.524241983890533
1.5 0.516650021076202
1.6 0.509472489356995
1.7 0.502541601657867
1.8 0.495943129062653
1.9 0.489650040864944
2 0.483543366193771
2.1 0.477374464273453
2.2 0.471210181713104
2.3 0.465445935726166
2.4 0.460061460733414
2.5 0.454976469278336
2.6 0.450067400932312
2.7 0.445414841175079
2.8 0.441105097532272
2.9 0.436818420886993
3 0.432756334543228
3.1 0.428798943758011
3.2 0.425037682056427
3.3 0.421391069889069
3.4 0.418079882860184
3.5 0.414945602416992
3.6 0.411845296621323
3.7 0.408840388059616
3.8 0.406007468700409
3.9 0.403309524059296
4 0.40091261267662
4.1 0.398577958345413
4.2 0.396386981010437
4.3 0.394217044115067
4.4 0.391973793506622
4.5 0.389735072851181
4.6 0.387380301952362
4.7 0.385240346193314
4.8 0.383415222167969
4.9 0.381590485572815
5 0.379806876182556
};
\addlegendentry{\ourmodel}
\end{axis}

\end{tikzpicture}

%% file: figures/unrolled_occupancy.tex
\newcommand*{\width}{0.125\textwidth}

\begin{figure*}
    \definecolor{inconsistent-color}{RGB}{219, 94, 86}
    \definecolor{smearing-color}{RGB}{184, 219, 86}
    \definecolor{disjoint-color}{RGB}{86, 131, 219}
    \definecolor{mis-detection-color}{RGB}{131, 86, 200}
    \definecolor{multi-modal-color}{RGB}{86, 219, 147}
    \centering
    \begin{tikzpicture}
        \node[inner sep = 0pt, outer sep = 0pt, anchor = north, draw = black, line width = \lw] (gt-0)  at (0,0) {\includegraphics[width=\width,trim={0cm, 2cm, 0cm, 0cm}, clip]{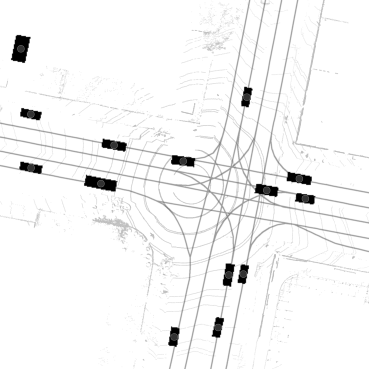}};
        \node[inner sep = 0pt, outer sep = 0pt, anchor = north, draw = black, line width = \lw] (gt-1)  at (gt-0.south) {\includegraphics[width=\width,trim={0cm, 2cm, 0cm, 0cm}, clip]{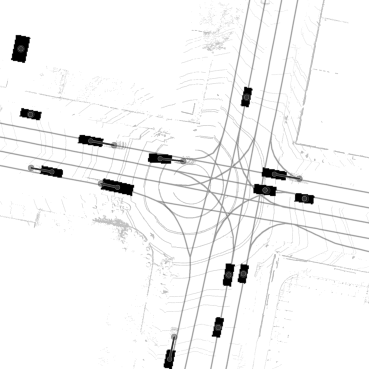}};
        \node[inner sep = 0pt, outer sep = 0pt, anchor = north, draw = black, line width = \lw] (gt-2)  at (gt-1.south) {\includegraphics[width=\width,trim={0cm, 2cm, 0cm, 0cm}, clip]{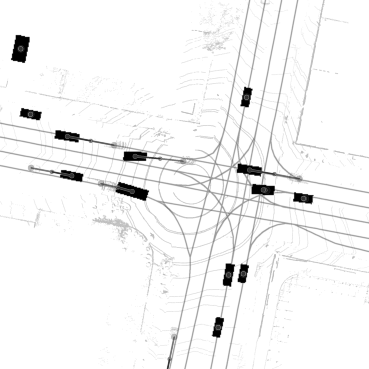}};
        \node[inner sep = 0pt, outer sep = 0pt, anchor = north, draw = black, line width = \lw] (gt-3)  at (gt-2.south) {\includegraphics[width=\width,trim={0cm, 2cm, 0cm, 0cm}, clip]{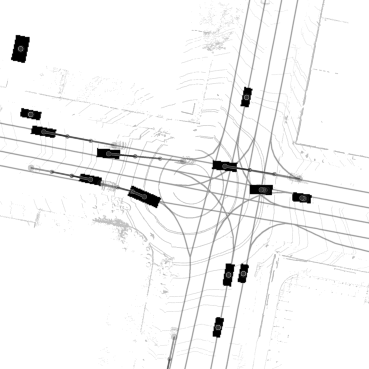}};
        \node[inner sep = 0pt, outer sep = 0pt, anchor = north, draw = black, line width = \lw] (gt-4)  at (gt-3.south) {\includegraphics[width=\width,trim={0cm, 2cm, 0cm, 0cm}, clip]{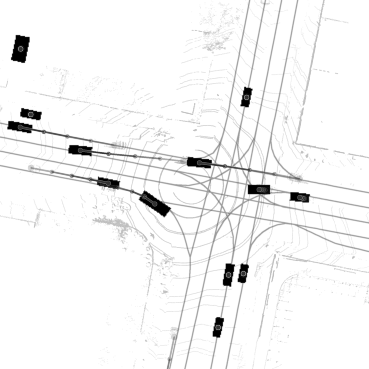}};
        \node[inner sep = 0pt, outer sep = 0pt, anchor = north, draw = black, line width = \lw] (gt-5)  at (gt-4.south) {\includegraphics[width=\width,trim={0cm, 2cm, 0cm, 0cm}, clip]{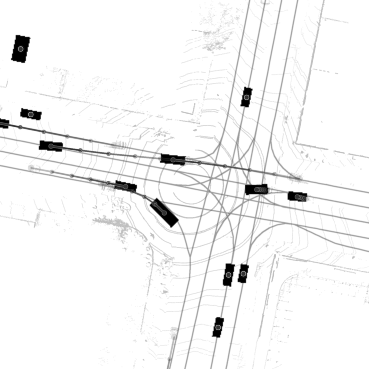}};
        \node[inner sep = 0pt, outer sep = 0pt, anchor = north, draw = black, line width = \lw] (gt-6)  at (gt-5.south) {\includegraphics[width=\width,trim={0cm, 2cm, 0cm, 0cm}, clip]{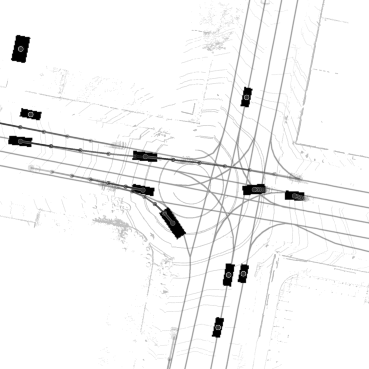}};
        \node[inner sep = 0pt, outer sep = 0pt, anchor = north, draw = black, line width = \lw] (gt-7)  at (gt-6.south) {\includegraphics[width=\width,trim={0cm, 2cm, 0cm, 0cm}, clip]{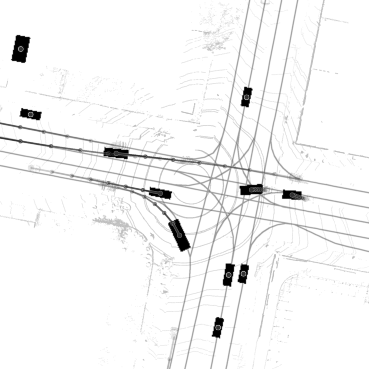}};
        \node[inner sep = 0pt, outer sep = 0pt, anchor = north, draw = black, line width = \lw] (gt-8)  at (gt-7.south) {\includegraphics[width=\width,trim={0cm, 2cm, 0cm, 0cm}, clip]{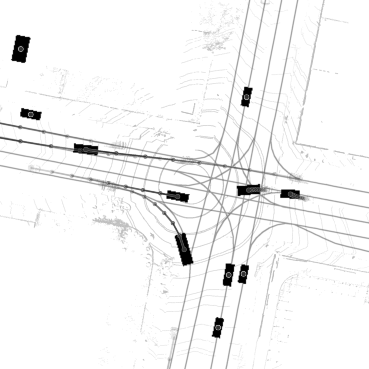}};
        \node[inner sep = 0pt, outer sep = 0pt, anchor = north, draw = black, line width = \lw] (gt-9)  at (gt-8.south) {\includegraphics[width=\width,trim={0cm, 2cm, 0cm, 0cm}, clip]{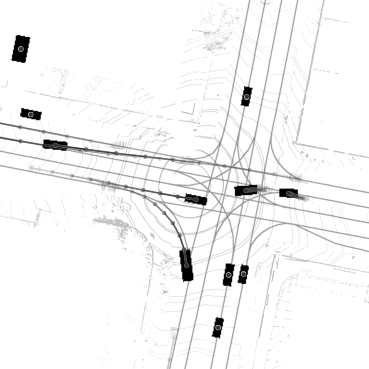}};
        \node[inner sep = 0pt, outer sep = 0pt, anchor = north, draw = black, line width = \lw] (gt-10) at (gt-9.south) {\includegraphics[width=\width,trim={0cm, 2cm, 0cm, 0cm}, clip]{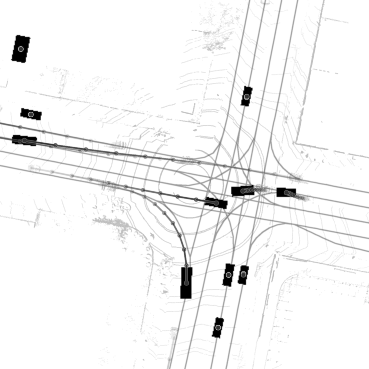}};
        \node[anchor=south] at (gt-0.north) {Ground Truth};

        \node[anchor=south, rotate=90] at (gt-0.west) {$\Delta t = \SI{0.0}{s}$};
        \node[anchor=south, rotate=90] at (gt-1.west) {$\Delta t = \SI{0.5}{s}$};
        \node[anchor=south, rotate=90] at (gt-2.west) {$\Delta t = \SI{1.0}{s}$};
        \node[anchor=south, rotate=90] at (gt-3.west) {$\Delta t = \SI{1.5}{s}$};
        \node[anchor=south, rotate=90] at (gt-4.west) {$\Delta t = \SI{2.0}{s}$};
        \node[anchor=south, rotate=90] at (gt-5.west) {$\Delta t = \SI{2.5}{s}$};
        \node[anchor=south, rotate=90] at (gt-6.west) {$\Delta t = \SI{3.0}{s}$};
        \node[anchor=south, rotate=90] at (gt-7.west) {$\Delta t = \SI{3.5}{s}$};
        \node[anchor=south, rotate=90] at (gt-8.west) {$\Delta t = \SI{4.0}{s}$};
        \node[anchor=south, rotate=90] at (gt-9.west) {$\Delta t = \SI{4.5}{s}$};
        \node[anchor=south, rotate=90] at (gt-10.west) {$\Delta t = \SI{5.0}{s}$};

        \node[inner sep = 0pt, outer sep = 0pt, anchor = west, draw = black, line width = \lw] (gorela-0)  at (gt-0.east) {\includegraphics[width=\width,trim={0cm, 2cm, 0cm, 0cm}, clip]{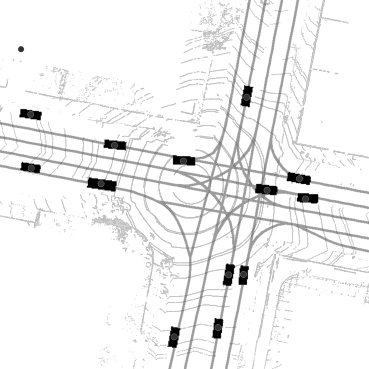}};
        \node[inner sep = 0pt, outer sep = 0pt, anchor = north, draw = black, line width = \lw] (gorela-1)  at (gorela-0.south) {\includegraphics[width=\width,trim={0cm, 2cm, 0cm, 0cm}, clip]{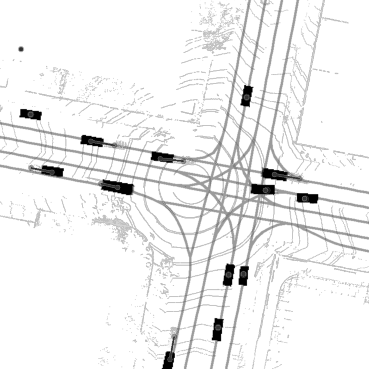}};
        \node[inner sep = 0pt, outer sep = 0pt, anchor = north, draw = black, line width = \lw] (gorela-2)  at (gorela-1.south) {\includegraphics[width=\width,trim={0cm, 2cm, 0cm, 0cm}, clip]{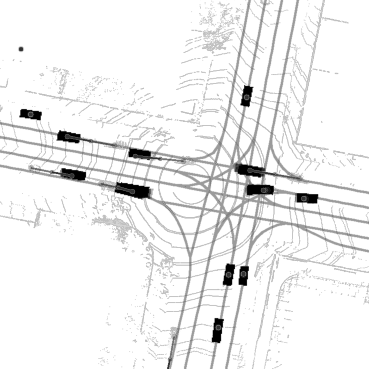}};
        \node[inner sep = 0pt, outer sep = 0pt, anchor = north, draw = black, line width = \lw] (gorela-3)  at (gorela-2.south) {\includegraphics[width=\width,trim={0cm, 2cm, 0cm, 0cm}, clip]{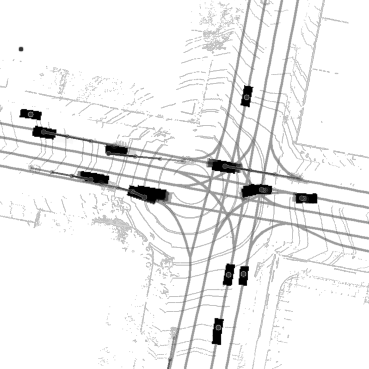}};
        \node[inner sep = 0pt, outer sep = 0pt, anchor = north, draw = black, line width = \lw] (gorela-4)  at (gorela-3.south) {\includegraphics[width=\width,trim={0cm, 2cm, 0cm, 0cm}, clip]{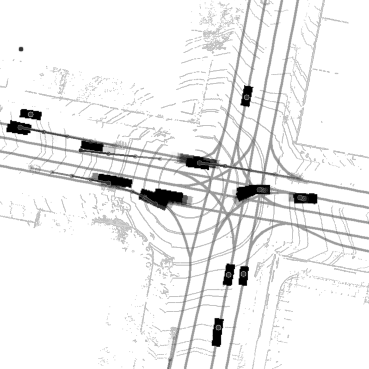}};
        \node[inner sep = 0pt, outer sep = 0pt, anchor = north, draw = black, line width = \lw] (gorela-5)  at (gorela-4.south) {\includegraphics[width=\width,trim={0cm, 2cm, 0cm, 0cm}, clip]{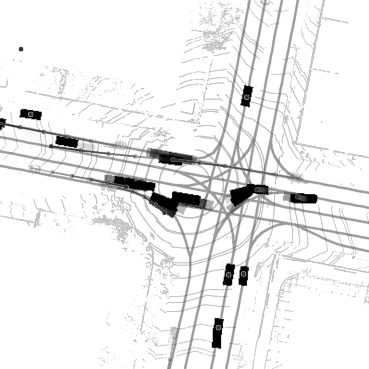}};
        \node[inner sep = 0pt, outer sep = 0pt, anchor = north, draw = black, line width = \lw] (gorela-6)  at (gorela-5.south) {\includegraphics[width=\width,trim={0cm, 2cm, 0cm, 0cm}, clip]{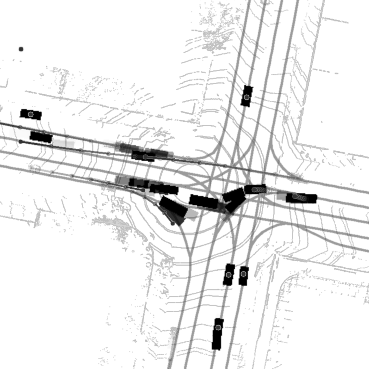}};
        \node[inner sep = 0pt, outer sep = 0pt, anchor = north, draw = black, line width = \lw] (gorela-7)  at (gorela-6.south) {\includegraphics[width=\width,trim={0cm, 2cm, 0cm, 0cm}, clip]{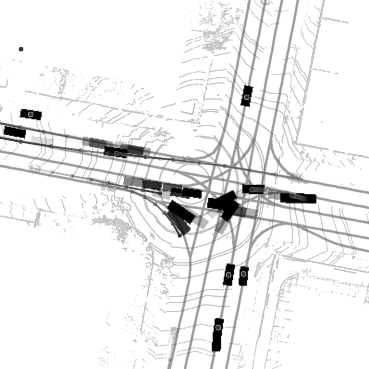}};
        \node[inner sep = 0pt, outer sep = 0pt, anchor = north, draw = black, line width = \lw] (gorela-8)  at (gorela-7.south) {\includegraphics[width=\width,trim={0cm, 2cm, 0cm, 0cm}, clip]{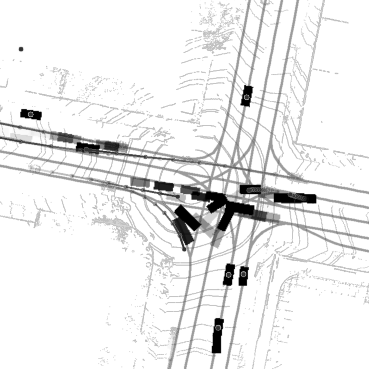}};
        \node[inner sep = 0pt, outer sep = 0pt, anchor = north, draw = black, line width = \lw] (gorela-9)  at (gorela-8.south) {\includegraphics[width=\width,trim={0cm, 2cm, 0cm, 0cm}, clip]{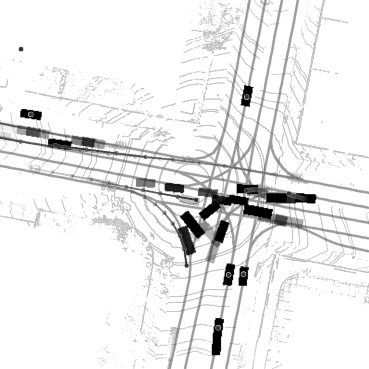}};
        \node[inner sep = 0pt, outer sep = 0pt, anchor = north, draw = black, line width = \lw] (gorela-10) at (gorela-9.south) {\includegraphics[width=\width,trim={0cm, 2cm, 0cm, 0cm}, clip]{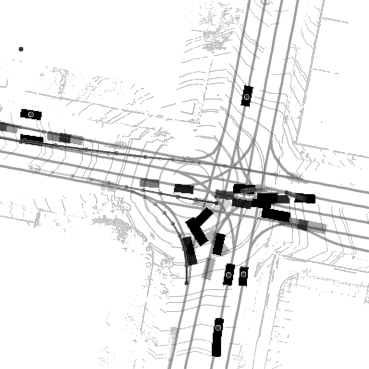}};
        \node[anchor=south] at (gorela-0.north) {\textsc{GoRela}};
        \node[draw = inconsistent-color, line width=2pt, minimum width=0.08\textwidth, minimum height=0.06\textwidth, yshift=-0.02\textwidth] at (gorela-7.center) {};
        \node[draw = inconsistent-color, line width=2pt, minimum width=0.08\textwidth, minimum height=0.06\textwidth, yshift=-0.02\textwidth] at (gorela-8.center) {};
        \node[draw = inconsistent-color, line width=2pt, minimum width=0.08\textwidth, minimum height=0.06\textwidth, yshift=-0.02\textwidth] at (gorela-9.center) {};
        \node[draw = inconsistent-color, line width=2pt, minimum width=0.08\textwidth, minimum height=0.06\textwidth, yshift=-0.02\textwidth] at (gorela-10.center) {};

        \node[inner sep = 0pt, outer sep = 0pt, anchor = west, draw = black, line width = \lw] (occflow-0)  at (gorela-0.east) {\includegraphics[width=\width,trim={0cm, 2cm, 0cm, 0cm}, clip]{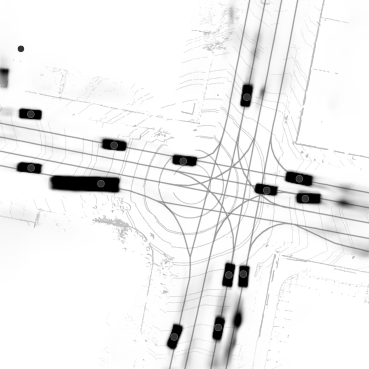}};
        \node[inner sep = 0pt, outer sep = 0pt, anchor = north, draw = black, line width = \lw] (occflow-1)  at (occflow-0.south) {\includegraphics[width=\width,trim={0cm, 2cm, 0cm, 0cm}, clip]{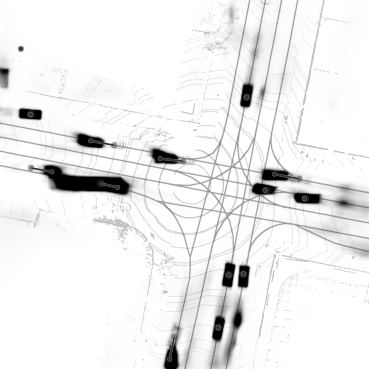}};
        \node[inner sep = 0pt, outer sep = 0pt, anchor = north, draw = black, line width = \lw] (occflow-2)  at (occflow-1.south) {\includegraphics[width=\width,trim={0cm, 2cm, 0cm, 0cm}, clip]{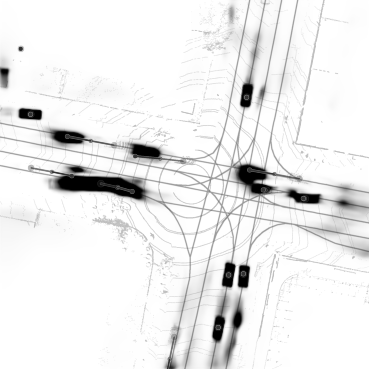}};
        \node[inner sep = 0pt, outer sep = 0pt, anchor = north, draw = black, line width = \lw] (occflow-3)  at (occflow-2.south) {\includegraphics[width=\width,trim={0cm, 2cm, 0cm, 0cm}, clip]{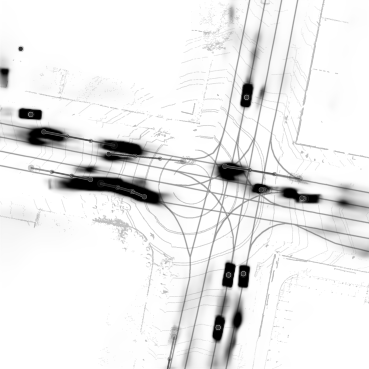}};
        \node[inner sep = 0pt, outer sep = 0pt, anchor = north, draw = black, line width = \lw] (occflow-4)  at (occflow-3.south) {\includegraphics[width=\width,trim={0cm, 2cm, 0cm, 0cm}, clip]{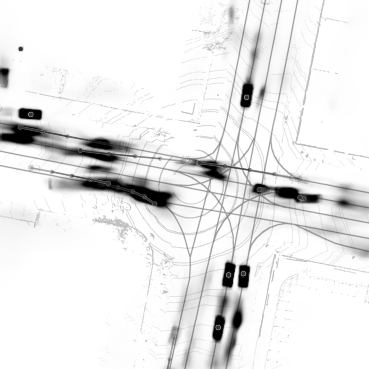}};
        \node[inner sep = 0pt, outer sep = 0pt, anchor = north, draw = black, line width = \lw] (occflow-5)  at (occflow-4.south) {\includegraphics[width=\width,trim={0cm, 2cm, 0cm, 0cm}, clip]{figures/raw-images/unrolled/argo_waymo_occ_flow_vis_0.2/24/75/occ_preds_24_75_04.png}};
        \node[inner sep = 0pt, outer sep = 0pt, anchor = north, draw = black, line width = \lw] (occflow-6)  at (occflow-5.south) {\includegraphics[width=\width,trim={0cm, 2cm, 0cm, 0cm}, clip]{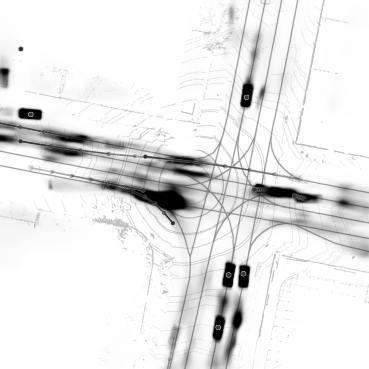}};
        \node[inner sep = 0pt, outer sep = 0pt, anchor = north, draw = black, line width = \lw] (occflow-7)  at (occflow-6.south) {\includegraphics[width=\width,trim={0cm, 2cm, 0cm, 0cm}, clip]{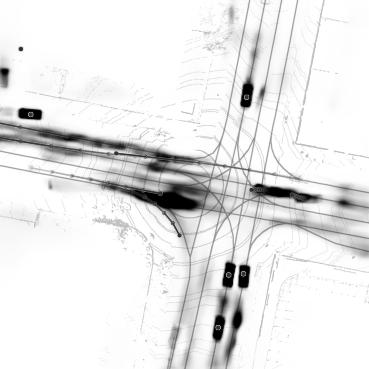}};
        \node[inner sep = 0pt, outer sep = 0pt, anchor = north, draw = black, line width = \lw] (occflow-8)  at (occflow-7.south) {\includegraphics[width=\width,trim={0cm, 2cm, 0cm, 0cm}, clip]{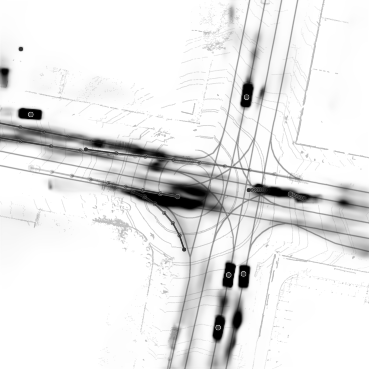}};
        \node[inner sep = 0pt, outer sep = 0pt, anchor = north, draw = black, line width = \lw] (occflow-9)  at (occflow-8.south) {\includegraphics[width=\width,trim={0cm, 2cm, 0cm, 0cm}, clip]{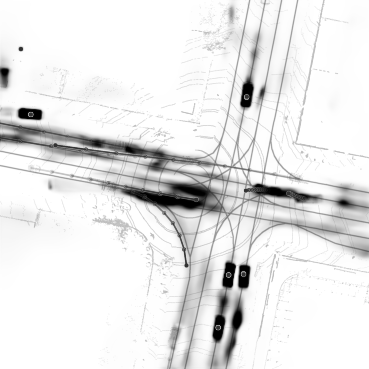}};
        \node[inner sep = 0pt, outer sep = 0pt, anchor = north, draw = black, line width = \lw] (occflow-10) at (occflow-9.south) {\includegraphics[width=\width,trim={0cm, 2cm, 0cm, 0cm}, clip]{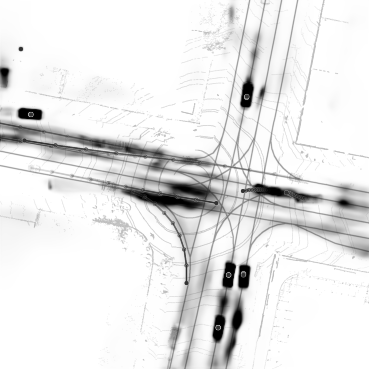}};
        \node[anchor=south] at (occflow-0.north) {\textsc{OccFlow}};
        \node[draw = mis-detection-color, line width=2pt, minimum width=0.035\textwidth, minimum height=0.02\textwidth, yshift=-0.015\textwidth, xshift=-0.033\textwidth] at (occflow-0.center) {};
        \node[draw = smearing-color, line width=2pt, minimum width=0.08\textwidth, minimum height=0.03\textwidth, yshift=-0.01\textwidth, xshift=-0.02\textwidth] at (occflow-6.center) {};
        \node[draw = smearing-color, line width=2pt, minimum width=0.08\textwidth, minimum height=0.03\textwidth, yshift=-0.01\textwidth, xshift=-0.02\textwidth] at (occflow-7.center) {};
        \node[draw = smearing-color, line width=2pt, minimum width=0.08\textwidth, minimum height=0.03\textwidth, yshift=-0.01\textwidth, xshift=-0.02\textwidth] at (occflow-8.center) {};
        \node[draw = smearing-color, line width=2pt, minimum width=0.08\textwidth, minimum height=0.03\textwidth, yshift=-0.01\textwidth, xshift=-0.02\textwidth] at (occflow-9.center) {};
        \node[draw = smearing-color, line width=2pt, minimum width=0.08\textwidth, minimum height=0.03\textwidth, yshift=-0.01\textwidth, xshift=-0.02\textwidth] at (occflow-10.center) {};

        \node[inner sep = 0pt, outer sep = 0pt, anchor = west, draw = black, line width = \lw] (mp3-0)  at (occflow-0.east) {\includegraphics[width=\width,trim={0cm, 2cm, 0cm, 0cm}, clip]{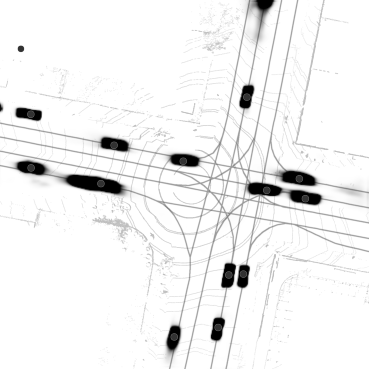}};
        \node[inner sep = 0pt, outer sep = 0pt, anchor = north, draw = black, line width = \lw] (mp3-1)  at (mp3-0.south) {\includegraphics[width=\width,trim={0cm, 2cm, 0cm, 0cm}, clip]{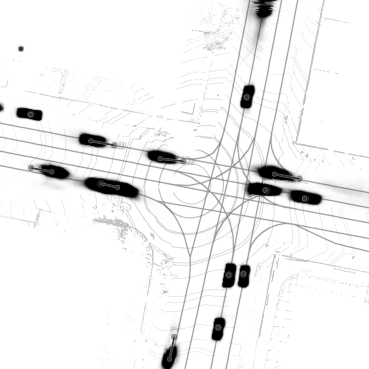}};
        \node[inner sep = 0pt, outer sep = 0pt, anchor = north, draw = black, line width = \lw] (mp3-2)  at (mp3-1.south) {\includegraphics[width=\width,trim={0cm, 2cm, 0cm, 0cm}, clip]{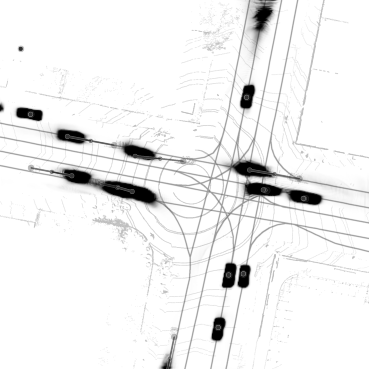}};
        \node[inner sep = 0pt, outer sep = 0pt, anchor = north, draw = black, line width = \lw] (mp3-3)  at (mp3-2.south) {\includegraphics[width=\width,trim={0cm, 2cm, 0cm, 0cm}, clip]{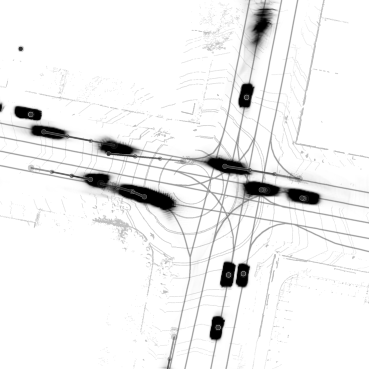}};
        \node[inner sep = 0pt, outer sep = 0pt, anchor = north, draw = black, line width = \lw] (mp3-4)  at (mp3-3.south) {\includegraphics[width=\width,trim={0cm, 2cm, 0cm, 0cm}, clip]{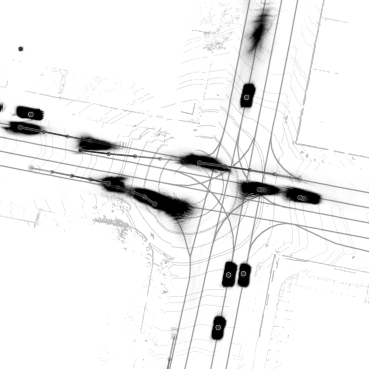}};
        \node[inner sep = 0pt, outer sep = 0pt, anchor = north, draw = black, line width = \lw] (mp3-5)  at (mp3-4.south) {\includegraphics[width=\width,trim={0cm, 2cm, 0cm, 0cm}, clip]{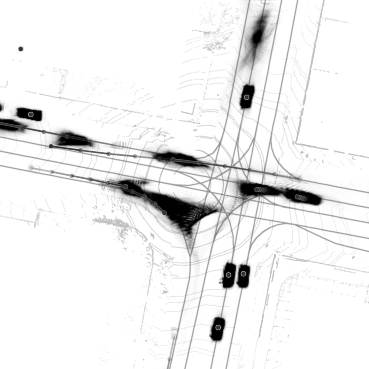}};
        \node[inner sep = 0pt, outer sep = 0pt, anchor = north, draw = black, line width = \lw] (mp3-6)  at (mp3-5.south) {\includegraphics[width=\width,trim={0cm, 2cm, 0cm, 0cm}, clip]{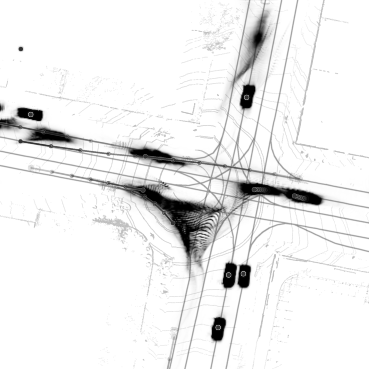}};
        \node[inner sep = 0pt, outer sep = 0pt, anchor = north, draw = black, line width = \lw] (mp3-7)  at (mp3-6.south) {\includegraphics[width=\width,trim={0cm, 2cm, 0cm, 0cm}, clip]{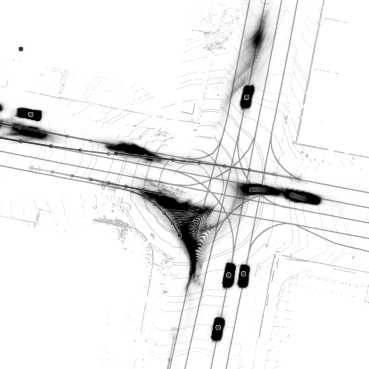}};
        \node[inner sep = 0pt, outer sep = 0pt, anchor = north, draw = black, line width = \lw] (mp3-8)  at (mp3-7.south) {\includegraphics[width=\width,trim={0cm, 2cm, 0cm, 0cm}, clip]{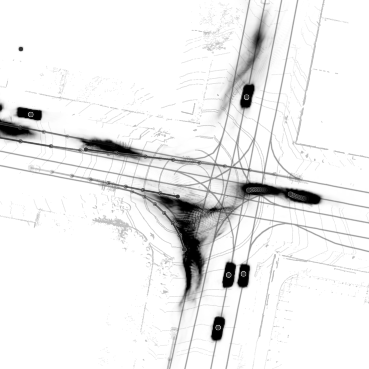}};
        \node[inner sep = 0pt, outer sep = 0pt, anchor = north, draw = black, line width = \lw] (mp3-9)  at (mp3-8.south) {\includegraphics[width=\width,trim={0cm, 2cm, 0cm, 0cm}, clip]{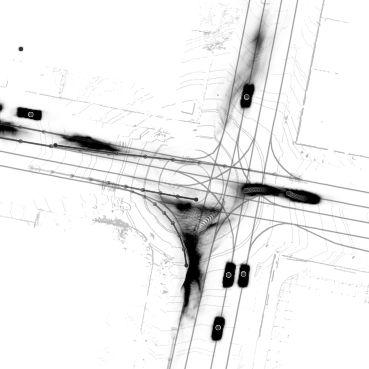}};
        \node[inner sep = 0pt, outer sep = 0pt, anchor = north, draw = black, line width = \lw] (mp3-10) at (mp3-9.south) {\includegraphics[width=\width,trim={0cm, 2cm, 0cm, 0cm}, clip]{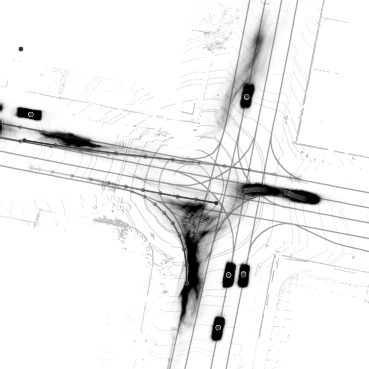}};
        \node[anchor=south] at (mp3-0.north) {\textsc{MP3}};
        \node[draw = disjoint-color, line width=2pt, minimum width=0.05\textwidth, minimum height=0.035\textwidth, yshift=-0.03\textwidth, xshift=0.0\textwidth] at (mp3-6.center) {};
        \node[draw = disjoint-color, line width=2pt, minimum width=0.05\textwidth, minimum height=0.035\textwidth, yshift=-0.03\textwidth, xshift=0.0\textwidth] at (mp3-7.center) {};
        \node[draw = disjoint-color, line width=2pt, minimum width=0.05\textwidth, minimum height=0.035\textwidth, yshift=-0.03\textwidth, xshift=0.0\textwidth] at (mp3-8.center) {};
        \node[draw = disjoint-color, line width=2pt, minimum width=0.05\textwidth, minimum height=0.035\textwidth, yshift=-0.03\textwidth, xshift=0.0\textwidth] at (mp3-9.center) {};

        \node[inner sep = 0pt, outer sep = 0pt, anchor = west, draw = black, line width = \lw] (implicitO-0)  at (mp3-0.east) {\includegraphics[width=\width,trim={0cm, 2cm, 0cm, 0cm}, clip]{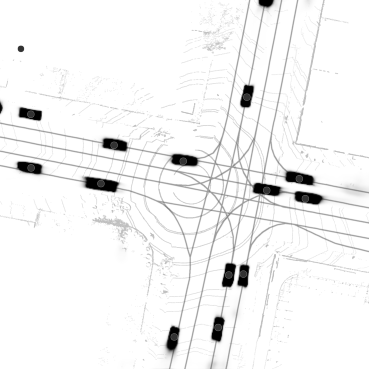}};
        \node[inner sep = 0pt, outer sep = 0pt, anchor = north, draw = black, line width = \lw] (implicitO-1)  at (implicitO-0.south) {\includegraphics[width=\width,trim={0cm, 2cm, 0cm, 0cm}, clip]{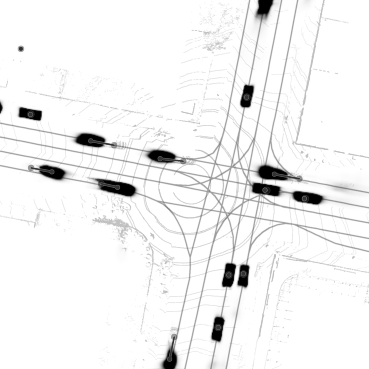}};
        \node[inner sep = 0pt, outer sep = 0pt, anchor = north, draw = black, line width = \lw] (implicitO-2)  at (implicitO-1.south) {\includegraphics[width=\width,trim={0cm, 2cm, 0cm, 0cm}, clip]{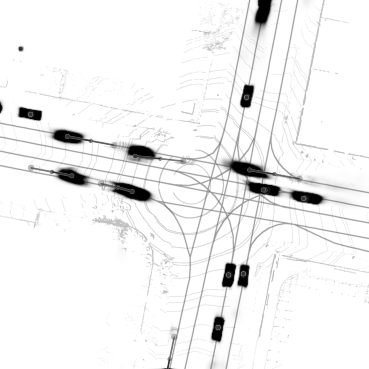}};
        \node[inner sep = 0pt, outer sep = 0pt, anchor = north, draw = black, line width = \lw] (implicitO-3)  at (implicitO-2.south) {\includegraphics[width=\width,trim={0cm, 2cm, 0cm, 0cm}, clip]{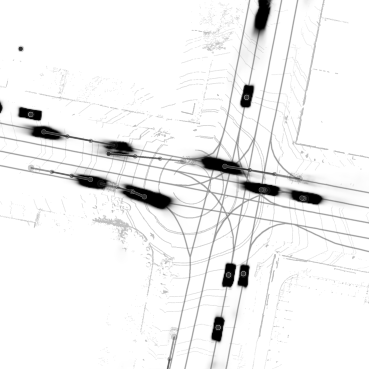}};
        \node[inner sep = 0pt, outer sep = 0pt, anchor = north, draw = black, line width = \lw] (implicitO-4)  at (implicitO-3.south) {\includegraphics[width=\width,trim={0cm, 2cm, 0cm, 0cm}, clip]{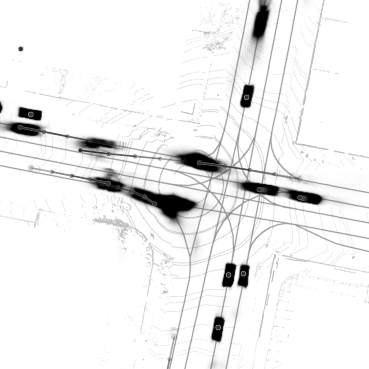}};
        \node[inner sep = 0pt, outer sep = 0pt, anchor = north, draw = black, line width = \lw] (implicitO-5)  at (implicitO-4.south) {\includegraphics[width=\width,trim={0cm, 2cm, 0cm, 0cm}, clip]{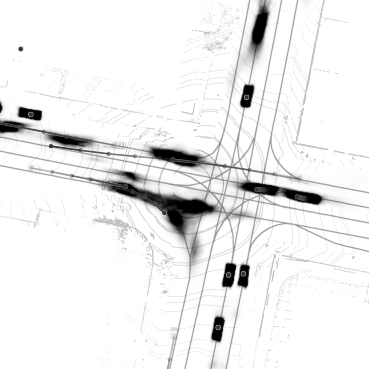}};
        \node[inner sep = 0pt, outer sep = 0pt, anchor = north, draw = black, line width = \lw] (implicitO-6)  at (implicitO-5.south) {\includegraphics[width=\width,trim={0cm, 2cm, 0cm, 0cm}, clip]{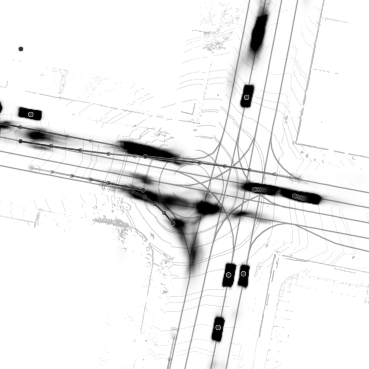}};
        \node[inner sep = 0pt, outer sep = 0pt, anchor = north, draw = black, line width = \lw] (implicitO-7)  at (implicitO-6.south) {\includegraphics[width=\width,trim={0cm, 2cm, 0cm, 0cm}, clip]{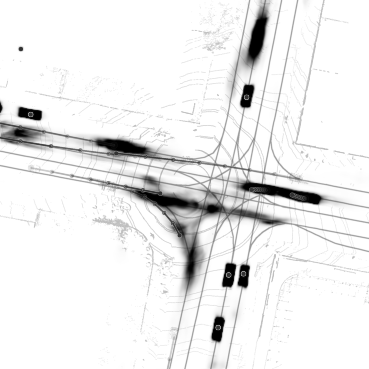}};
        \node[inner sep = 0pt, outer sep = 0pt, anchor = north, draw = black, line width = \lw] (implicitO-8)  at (implicitO-7.south) {\includegraphics[width=\width,trim={0cm, 2cm, 0cm, 0cm}, clip]{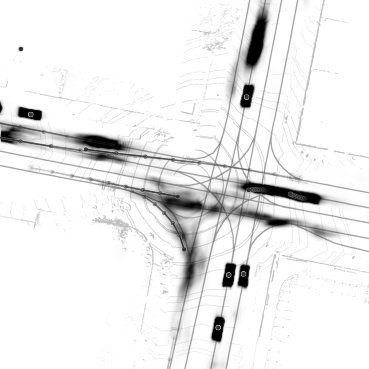}};
        \node[inner sep = 0pt, outer sep = 0pt, anchor = north, draw = black, line width = \lw] (implicitO-9)  at (implicitO-8.south) {\includegraphics[width=\width,trim={0cm, 2cm, 0cm, 0cm}, clip]{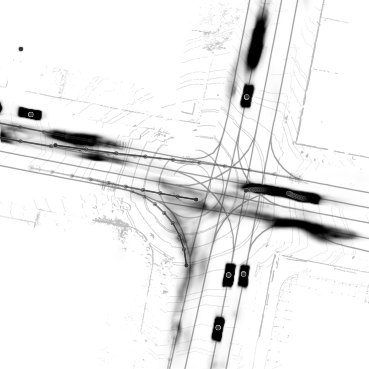}};
        \node[inner sep = 0pt, outer sep = 0pt, anchor = north, draw = black, line width = \lw] (implicitO-10) at (implicitO-9.south) {\includegraphics[width=\width,trim={0cm, 2cm, 0cm, 0cm}, clip]{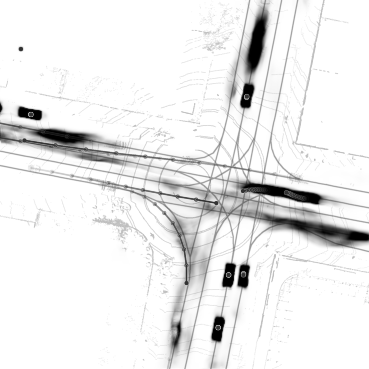}};
        \node[anchor=south] at (implicitO-0.north) {\ourmodel{}};
        \node[draw = multi-modal-color, line width=2pt, minimum width=0.08\textwidth, minimum height=0.04\textwidth, yshift=-0.03\textwidth, xshift=0.02\textwidth] at (implicitO-6.center) {};
        \node[draw = multi-modal-color, line width=2pt, minimum width=0.08\textwidth, minimum height=0.04\textwidth, yshift=-0.03\textwidth, xshift=0.02\textwidth] at (implicitO-7.center) {};
        \node[draw = multi-modal-color, line width=2pt, minimum width=0.08\textwidth, minimum height=0.04\textwidth, yshift=-0.03\textwidth, xshift=0.02\textwidth] at (implicitO-8.center) {};
        \node[draw = multi-modal-color, line width=2pt, minimum width=0.08\textwidth, minimum height=0.04\textwidth, yshift=-0.03\textwidth, xshift=0.02\textwidth] at (implicitO-9.center) {};
        \node[draw = multi-modal-color, line width=2pt, minimum width=0.08\textwidth, minimum height=0.04\textwidth, yshift=-0.03\textwidth, xshift=0.02\textwidth] at (implicitO-10.center) {};

        \node[draw = multi-modal-color, line width=2pt, minimum width=0.035\textwidth, minimum height=0.03\textwidth, yshift=0.0\textwidth, xshift=-0.03\textwidth] at (implicitO-8.center) {};
        \node[draw = multi-modal-color, line width=2pt, minimum width=0.035\textwidth, minimum height=0.03\textwidth, yshift=0.0\textwidth, xshift=-0.04\textwidth] at (implicitO-9.center) {};
        \node[draw = multi-modal-color, line width=2pt, minimum width=0.035\textwidth, minimum height=0.03\textwidth, yshift=0.0\textwidth, xshift=-0.045\textwidth] at (implicitO-10.center) {};

    \end{tikzpicture}
    \caption{Occupancy predictions over time on a single scene from AV2 for various baseline models and \ourmodel{}.
    The alpha channel denotes occupancy probability, colored in black.
    Observations are denoted with colored boxes:
        \textbf{\textcolor{inconsistent-color}{inconsistent with actors}},
        \textbf{\textcolor{smearing-color}{spreading occupancy predictions}},
        \textbf{\textcolor{mis-detection-color}{incorrect detection size}},
        \textbf{\textcolor{disjoint-color}{disjoint-pixel occupancy predictions}},
        \textbf{\textcolor{multi-modal-color}{realistic multi-modal predictions}},
    }
    \vspace{-10pt}
    \label{fig:unrolled-occupancy}
\end{figure*}

%% file: figures/reliability_diagram.tex
\begin{figure*}
    \centering
    {
    \setlength\tabcolsep{1.5pt}
    \begin{tabular} {c c c}
        \input{figures/reliability_plots/argo_multipath.tex} &
        \input{figures/reliability_plots/argo_lanegcn.tex} &
        \input{figures/reliability_plots/argo_gorela.tex} \\
        \input{figures/reliability_plots/argo_occ_flow.tex} &
        \input{figures/reliability_plots/argo_mp3.tex} &
        \input{figures/reliability_plots/argo_implicitO.tex} 
    \end{tabular}
    }
    \caption{Reliability diagrams on AV2.}
    \vspace{-10pt}
    \label{fig:argo-reliability-plots}    
\end{figure*}

%% file: figures/reliability_plots/argo_multipath.tex
\begin{tikzpicture}

\definecolor{darkgray176}{RGB}{176,176,176}
\definecolor{gray}{RGB}{128,128,128}
\definecolor{lightgray204}{RGB}{204,204,204}
\definecolor{tomato2406060}{RGB}{240,60,60}

\begin{axis}[
legend cell align={left},
legend style={
  fill opacity=0.8,
  draw opacity=1,
  text opacity=1,
  at={(0.03,0.97)},
  anchor=north west,
  draw=lightgray204
},
tick align=outside,
tick pos=left,
x grid style={darkgray176},
xlabel={\footnotesize{Confidence}},
xmin=0, xmax=1,
xtick style={color=black},
y grid style={darkgray176},
ylabel={\footnotesize{Expected Accuracy}},
ymin=0, ymax=1,
ytick style={color=black},
scale=0.6,
title={\textsc{MultiPath}}
]
\draw[draw=tomato2406060,fill=tomato2406060] (axis cs:-3.46944695195361e-18,0.000134489628) rectangle (axis cs:0.03333333,0.00550377);
\addlegendimage{ybar,ybar legend,draw=tomato2406060,fill=tomato2406060}
\addlegendentry{Gap}

\draw[draw=tomato2406060,fill=tomato2406060,opacity=0.10329303993935] (axis cs:0.03333333,0.0503385356) rectangle (axis cs:0.06666667,0.08015523);
\draw[draw=tomato2406060,fill=tomato2406060,opacity=0.102884206664503] (axis cs:0.06666667,0.0815596788) rectangle (axis cs:0.1,0.09818788);
\draw[draw=tomato2406060,fill=tomato2406060,opacity=0.102335895176175] (axis cs:0.1,0.11719139) rectangle (axis cs:0.13333333,0.120555179);
\draw[draw=tomato2406060,fill=tomato2406060,opacity=0.101615786664498] (axis cs:0.13333333,0.13221168) rectangle (axis cs:0.16666667,0.147014118);
\draw[draw=tomato2406060,fill=tomato2406060,opacity=0.10148767301155] (axis cs:0.16666667,0.15039349) rectangle (axis cs:0.2,0.184500458);
\draw[draw=tomato2406060,fill=tomato2406060,opacity=0.101654852626902] (axis cs:0.2,0.1486731) rectangle (axis cs:0.23333333,0.216891108);
\draw[draw=tomato2406060,fill=tomato2406060,opacity=0.100809650181109] (axis cs:0.23333333,0.16895061) rectangle (axis cs:0.26666667,0.247258489);
\draw[draw=tomato2406060,fill=tomato2406060,opacity=0.100560340610385] (axis cs:0.26666667,0.23436719) rectangle (axis cs:0.3,0.283626183);
\draw[draw=tomato2406060,fill=tomato2406060,opacity=0.10081133366928] (axis cs:0.3,0.25504931) rectangle (axis cs:0.33333333,0.316495426);
\draw[draw=tomato2406060,fill=tomato2406060,opacity=0.10052711050339] (axis cs:0.33333333,0.27275551) rectangle (axis cs:0.36666667,0.349883472);
\draw[draw=tomato2406060,fill=tomato2406060,opacity=0.100508627671549] (axis cs:0.36666667,0.28413879) rectangle (axis cs:0.4,0.383687119);
\draw[draw=tomato2406060,fill=tomato2406060,opacity=0.100427419674718] (axis cs:0.4,0.30192612) rectangle (axis cs:0.43333333,0.414185674);
\draw[draw=tomato2406060,fill=tomato2406060,opacity=0.10024286449161] (axis cs:0.43333333,0.37603477) rectangle (axis cs:0.46666667,0.450879741);
\draw[draw=tomato2406060,fill=tomato2406060,opacity=0.105763007991681] (axis cs:0.46666667,0.02993464) rectangle (axis cs:0.5,0.498860891);
\draw[draw=tomato2406060,fill=tomato2406060,opacity=0.100353013152867] (axis cs:0.5,0.45626259) rectangle (axis cs:0.53333333,0.516964072);
\draw[draw=tomato2406060,fill=tomato2406060,opacity=0.100226381436582] (axis cs:0.53333333,0.49656132) rectangle (axis cs:0.56666667,0.550183829);
\draw[draw=tomato2406060,fill=tomato2406060,opacity=0.100291726771192] (axis cs:0.56666667,0.5680837) rectangle (axis cs:0.6,0.584355234);
\draw[draw=tomato2406060,fill=tomato2406060,opacity=0.100394332227607] (axis cs:0.6,0.617335497) rectangle (axis cs:0.63333333,0.62580716);
\draw[draw=tomato2406060,fill=tomato2406060,opacity=0.100411848424501] (axis cs:0.63333333,0.650973121) rectangle (axis cs:0.66666667,0.68366016);
\draw[draw=tomato2406060,fill=tomato2406060,opacity=0.101271817437533] (axis cs:0.66666667,0.675370234) rectangle (axis cs:0.7,0.8398935);
\draw[draw=tomato2406060,fill=tomato2406060,opacity=0.100542949812036] (axis cs:0.7,0.717119358) rectangle (axis cs:0.73333333,0.76146878);
\draw[draw=tomato2406060,fill=tomato2406060,opacity=0.100717622371274] (axis cs:0.73333333,0.750628083) rectangle (axis cs:0.76666667,0.79365377);
\draw[draw=tomato2406060,fill=tomato2406060,opacity=0.100896368606353] (axis cs:0.76666667,0.784248307) rectangle (axis cs:0.8,0.80361074);
\draw[draw=tomato2406060,fill=tomato2406060,opacity=0.101381165984991] (axis cs:0.8,0.8117898) rectangle (axis cs:0.83333333,0.817649078);
\draw[draw=tomato2406060,fill=tomato2406060,opacity=0.102248756386574] (axis cs:0.83333333,0.84448468) rectangle (axis cs:0.86666667,0.851106529);
\draw[draw=tomato2406060,fill=tomato2406060,opacity=0.103371268832264] (axis cs:0.86666667,0.88222011) rectangle (axis cs:0.9,0.884351546);
\draw[draw=tomato2406060,fill=tomato2406060,opacity=0.102576895443247] (axis cs:0.9,0.90644381) rectangle (axis cs:0.93333333,0.913365652);
\draw[draw=tomato2406060,fill=tomato2406060,opacity=0.10027669823776] (axis cs:0.93333333,0.89957432) rectangle (axis cs:0.96666667,0.940154358);
\draw[draw=tomato2406060,fill=tomato2406060,opacity=0.1] (axis cs:0.96666667,0.80636002) rectangle (axis cs:1,0.980562942);
\draw[draw=black,fill=black,very thick] (axis cs:-3.46944695195361e-18,0.00550377) rectangle (axis cs:0.03333333,0.00550377);
\addlegendimage{ybar,ybar legend,draw=black,fill=black,very thick}
\addlegendentry{Accuracy}

\draw[draw=black,fill=black,very thick] (axis cs:0.03333333,0.08015523) rectangle (axis cs:0.06666667,0.08015523);
\draw[draw=black,fill=black,very thick] (axis cs:0.06666667,0.09818788) rectangle (axis cs:0.1,0.09818788);
\draw[draw=black,fill=black,very thick] (axis cs:0.1,0.11719139) rectangle (axis cs:0.13333333,0.11719139);
\draw[draw=black,fill=black,very thick] (axis cs:0.13333333,0.13221168) rectangle (axis cs:0.16666667,0.13221168);
\draw[draw=black,fill=black,very thick] (axis cs:0.16666667,0.15039349) rectangle (axis cs:0.2,0.15039349);
\draw[draw=black,fill=black,very thick] (axis cs:0.2,0.1486731) rectangle (axis cs:0.23333333,0.1486731);
\draw[draw=black,fill=black,very thick] (axis cs:0.23333333,0.16895061) rectangle (axis cs:0.26666667,0.16895061);
\draw[draw=black,fill=black,very thick] (axis cs:0.26666667,0.23436719) rectangle (axis cs:0.3,0.23436719);
\draw[draw=black,fill=black,very thick] (axis cs:0.3,0.25504931) rectangle (axis cs:0.33333333,0.25504931);
\draw[draw=black,fill=black,very thick] (axis cs:0.33333333,0.27275551) rectangle (axis cs:0.36666667,0.27275551);
\draw[draw=black,fill=black,very thick] (axis cs:0.36666667,0.28413879) rectangle (axis cs:0.4,0.28413879);
\draw[draw=black,fill=black,very thick] (axis cs:0.4,0.30192612) rectangle (axis cs:0.43333333,0.30192612);
\draw[draw=black,fill=black,very thick] (axis cs:0.43333333,0.37603477) rectangle (axis cs:0.46666667,0.37603477);
\draw[draw=black,fill=black,very thick] (axis cs:0.46666667,0.02993464) rectangle (axis cs:0.5,0.02993464);
\draw[draw=black,fill=black,very thick] (axis cs:0.5,0.45626259) rectangle (axis cs:0.53333333,0.45626259);
\draw[draw=black,fill=black,very thick] (axis cs:0.53333333,0.49656132) rectangle (axis cs:0.56666667,0.49656132);
\draw[draw=black,fill=black,very thick] (axis cs:0.56666667,0.5680837) rectangle (axis cs:0.6,0.5680837);
\draw[draw=black,fill=black,very thick] (axis cs:0.6,0.62580716) rectangle (axis cs:0.63333333,0.62580716);
\draw[draw=black,fill=black,very thick] (axis cs:0.63333333,0.68366016) rectangle (axis cs:0.66666667,0.68366016);
\draw[draw=black,fill=black,very thick] (axis cs:0.66666667,0.8398935) rectangle (axis cs:0.7,0.8398935);
\draw[draw=black,fill=black,very thick] (axis cs:0.7,0.76146878) rectangle (axis cs:0.73333333,0.76146878);
\draw[draw=black,fill=black,very thick] (axis cs:0.73333333,0.79365377) rectangle (axis cs:0.76666667,0.79365377);
\draw[draw=black,fill=black,very thick] (axis cs:0.76666667,0.80361074) rectangle (axis cs:0.8,0.80361074);
\draw[draw=black,fill=black,very thick] (axis cs:0.8,0.8117898) rectangle (axis cs:0.83333333,0.8117898);
\draw[draw=black,fill=black,very thick] (axis cs:0.83333333,0.84448468) rectangle (axis cs:0.86666667,0.84448468);
\draw[draw=black,fill=black,very thick] (axis cs:0.86666667,0.88222011) rectangle (axis cs:0.9,0.88222011);
\draw[draw=black,fill=black,very thick] (axis cs:0.9,0.90644381) rectangle (axis cs:0.93333333,0.90644381);
\draw[draw=black,fill=black,very thick] (axis cs:0.93333333,0.89957432) rectangle (axis cs:0.96666667,0.89957432);
\draw[draw=black,fill=black,very thick] (axis cs:0.96666667,0.80636002) rectangle (axis cs:1,0.80636002);
\addplot [semithick, gray, dashed, forget plot]
table {%
0 0
1 1
};
\draw (axis cs:0.98,0.02) node[
  scale=1.0,
  anchor=south east,
  text=black,
  rotate=0.0
]{ECE=0.916};
\end{axis}

\end{tikzpicture}

%% file: figures/reliability_plots/argo_lanegcn.tex
\begin{tikzpicture}

\definecolor{darkgray176}{RGB}{176,176,176}
\definecolor{gray}{RGB}{128,128,128}
\definecolor{lightgray204}{RGB}{204,204,204}
\definecolor{tomato2406060}{RGB}{240,60,60}

\begin{axis}[
legend cell align={left},
legend style={
  fill opacity=0.8,
  draw opacity=1,
  text opacity=1,
  at={(0.03,0.97)},
  anchor=north west,
  draw=lightgray204
},
tick align=outside,
tick pos=left,
x grid style={darkgray176},
xlabel={\footnotesize{Confidence}},
xmin=0, xmax=1,
xtick style={color=black},
y grid style={darkgray176},
ylabel={\footnotesize{Expected Accuracy}},
ymin=0, ymax=1,
ytick style={color=black},
scale=0.6,
title={\textsc{LaneGCN}}
]
\draw[draw=tomato2406060,fill=tomato2406060] (axis cs:-3.46944695195361e-18,0.000158251348) rectangle (axis cs:0.03333333,0.0061387);
\addlegendimage{ybar,ybar legend,draw=tomato2406060,fill=tomato2406060}
\addlegendentry{Gap}

\draw[draw=tomato2406060,fill=tomato2406060,opacity=0.102579787536033] (axis cs:0.03333333,0.0479637138) rectangle (axis cs:0.06666667,0.12957472);
\draw[draw=tomato2406060,fill=tomato2406060,opacity=0.100885877176658] (axis cs:0.06666667,0.0816398597) rectangle (axis cs:0.1,0.13839997);
\draw[draw=tomato2406060,fill=tomato2406060,opacity=0.100415889230206] (axis cs:0.1,0.115607839) rectangle (axis cs:0.13333333,0.14440735);
\draw[draw=tomato2406060,fill=tomato2406060,opacity=0.100301393756242] (axis cs:0.13333333,0.149303395) rectangle (axis cs:0.16666667,0.17795983);
\draw[draw=tomato2406060,fill=tomato2406060,opacity=0.100229319694617] (axis cs:0.16666667,0.18295235) rectangle (axis cs:0.2,0.183038652);
\draw[draw=tomato2406060,fill=tomato2406060,opacity=0.100156455231305] (axis cs:0.2,0.216674171) rectangle (axis cs:0.23333333,0.22116693);
\draw[draw=tomato2406060,fill=tomato2406060,opacity=0.100126524682827] (axis cs:0.23333333,0.22670303) rectangle (axis cs:0.26666667,0.249136856);
\draw[draw=tomato2406060,fill=tomato2406060,opacity=0.1000900468898] (axis cs:0.26666667,0.21958389) rectangle (axis cs:0.3,0.282772701);
\draw[draw=tomato2406060,fill=tomato2406060,opacity=0.100060017163531] (axis cs:0.3,0.23500841) rectangle (axis cs:0.33333333,0.316095022);
\draw[draw=tomato2406060,fill=tomato2406060,opacity=0.100023461337216] (axis cs:0.33333333,0.24696872) rectangle (axis cs:0.36666667,0.349797153);
\draw[draw=tomato2406060,fill=tomato2406060,opacity=0.10003018226876] (axis cs:0.36666667,0.27513001) rectangle (axis cs:0.4,0.383901115);
\draw[draw=tomato2406060,fill=tomato2406060,opacity=0.100035506656177] (axis cs:0.4,0.28957702) rectangle (axis cs:0.43333333,0.416301995);
\draw[draw=tomato2406060,fill=tomato2406060,opacity=0.1] (axis cs:0.43333333,0.29212384) rectangle (axis cs:0.46666667,0.449541565);
\draw[draw=tomato2406060,fill=tomato2406060,opacity=0.105321914516695] (axis cs:0.46666667,0.00941006) rectangle (axis cs:0.5,0.49937015);
\draw[draw=tomato2406060,fill=tomato2406060,opacity=0.100037206271525] (axis cs:0.5,0.30851686) rectangle (axis cs:0.53333333,0.517475057);
\draw[draw=tomato2406060,fill=tomato2406060,opacity=0.100030126890298] (axis cs:0.53333333,0.35731146) rectangle (axis cs:0.56666667,0.549825271);
\draw[draw=tomato2406060,fill=tomato2406060,opacity=0.100024938431834] (axis cs:0.56666667,0.36410652) rectangle (axis cs:0.6,0.583632629);
\draw[draw=tomato2406060,fill=tomato2406060,opacity=0.100049796816618] (axis cs:0.6,0.39366002) rectangle (axis cs:0.63333333,0.617023485);
\draw[draw=tomato2406060,fill=tomato2406060,opacity=0.100086477496193] (axis cs:0.63333333,0.39508572) rectangle (axis cs:0.66666667,0.651045509);
\draw[draw=tomato2406060,fill=tomato2406060,opacity=0.100201370688203] (axis cs:0.66666667,0.49510888) rectangle (axis cs:0.7,0.684887004);
\draw[draw=tomato2406060,fill=tomato2406060,opacity=0.100447338658868] (axis cs:0.7,0.62151212) rectangle (axis cs:0.73333333,0.717590568);
\draw[draw=tomato2406060,fill=tomato2406060,opacity=0.100650823294005] (axis cs:0.73333333,0.68446063) rectangle (axis cs:0.76666667,0.749022541);
\draw[draw=tomato2406060,fill=tomato2406060,opacity=0.100483020008921] (axis cs:0.76666667,0.65685793) rectangle (axis cs:0.8,0.783177332);
\draw[draw=tomato2406060,fill=tomato2406060,opacity=0.100443128888863] (axis cs:0.8,0.69433169) rectangle (axis cs:0.83333333,0.816949801);
\draw[draw=tomato2406060,fill=tomato2406060,opacity=0.100573527043398] (axis cs:0.83333333,0.69673472) rectangle (axis cs:0.86666667,0.850871319);
\draw[draw=tomato2406060,fill=tomato2406060,opacity=0.100960518785165] (axis cs:0.86666667,0.75298431) rectangle (axis cs:0.9,0.884375138);
\draw[draw=tomato2406060,fill=tomato2406060,opacity=0.102195448422603] (axis cs:0.9,0.80157333) rectangle (axis cs:0.93333333,0.919226686);
\draw[draw=tomato2406060,fill=tomato2406060,opacity=0.103882799958295] (axis cs:0.93333333,0.85114904) rectangle (axis cs:0.96666667,0.950393304);
\draw[draw=tomato2406060,fill=tomato2406060,opacity=0.103996594642992] (axis cs:0.96666667,0.88025302) rectangle (axis cs:1,0.981633927);
\draw[draw=black,fill=black,very thick] (axis cs:-3.46944695195361e-18,0.0061387) rectangle (axis cs:0.03333333,0.0061387);
\addlegendimage{ybar,ybar legend,draw=black,fill=black,very thick}
\addlegendentry{Accuracy}

\draw[draw=black,fill=black,very thick] (axis cs:0.03333333,0.12957472) rectangle (axis cs:0.06666667,0.12957472);
\draw[draw=black,fill=black,very thick] (axis cs:0.06666667,0.13839997) rectangle (axis cs:0.1,0.13839997);
\draw[draw=black,fill=black,very thick] (axis cs:0.1,0.14440735) rectangle (axis cs:0.13333333,0.14440735);
\draw[draw=black,fill=black,very thick] (axis cs:0.13333333,0.17795983) rectangle (axis cs:0.16666667,0.17795983);
\draw[draw=black,fill=black,very thick] (axis cs:0.16666667,0.18295235) rectangle (axis cs:0.2,0.18295235);
\draw[draw=black,fill=black,very thick] (axis cs:0.2,0.22116693) rectangle (axis cs:0.23333333,0.22116693);
\draw[draw=black,fill=black,very thick] (axis cs:0.23333333,0.22670303) rectangle (axis cs:0.26666667,0.22670303);
\draw[draw=black,fill=black,very thick] (axis cs:0.26666667,0.21958389) rectangle (axis cs:0.3,0.21958389);
\draw[draw=black,fill=black,very thick] (axis cs:0.3,0.23500841) rectangle (axis cs:0.33333333,0.23500841);
\draw[draw=black,fill=black,very thick] (axis cs:0.33333333,0.24696872) rectangle (axis cs:0.36666667,0.24696872);
\draw[draw=black,fill=black,very thick] (axis cs:0.36666667,0.27513001) rectangle (axis cs:0.4,0.27513001);
\draw[draw=black,fill=black,very thick] (axis cs:0.4,0.28957702) rectangle (axis cs:0.43333333,0.28957702);
\draw[draw=black,fill=black,very thick] (axis cs:0.43333333,0.29212384) rectangle (axis cs:0.46666667,0.29212384);
\draw[draw=black,fill=black,very thick] (axis cs:0.46666667,0.00941006) rectangle (axis cs:0.5,0.00941006);
\draw[draw=black,fill=black,very thick] (axis cs:0.5,0.30851686) rectangle (axis cs:0.53333333,0.30851686);
\draw[draw=black,fill=black,very thick] (axis cs:0.53333333,0.35731146) rectangle (axis cs:0.56666667,0.35731146);
\draw[draw=black,fill=black,very thick] (axis cs:0.56666667,0.36410652) rectangle (axis cs:0.6,0.36410652);
\draw[draw=black,fill=black,very thick] (axis cs:0.6,0.39366002) rectangle (axis cs:0.63333333,0.39366002);
\draw[draw=black,fill=black,very thick] (axis cs:0.63333333,0.39508572) rectangle (axis cs:0.66666667,0.39508572);
\draw[draw=black,fill=black,very thick] (axis cs:0.66666667,0.49510888) rectangle (axis cs:0.7,0.49510888);
\draw[draw=black,fill=black,very thick] (axis cs:0.7,0.62151212) rectangle (axis cs:0.73333333,0.62151212);
\draw[draw=black,fill=black,very thick] (axis cs:0.73333333,0.68446063) rectangle (axis cs:0.76666667,0.68446063);
\draw[draw=black,fill=black,very thick] (axis cs:0.76666667,0.65685793) rectangle (axis cs:0.8,0.65685793);
\draw[draw=black,fill=black,very thick] (axis cs:0.8,0.69433169) rectangle (axis cs:0.83333333,0.69433169);
\draw[draw=black,fill=black,very thick] (axis cs:0.83333333,0.69673472) rectangle (axis cs:0.86666667,0.69673472);
\draw[draw=black,fill=black,very thick] (axis cs:0.86666667,0.75298431) rectangle (axis cs:0.9,0.75298431);
\draw[draw=black,fill=black,very thick] (axis cs:0.9,0.80157333) rectangle (axis cs:0.93333333,0.80157333);
\draw[draw=black,fill=black,very thick] (axis cs:0.93333333,0.85114904) rectangle (axis cs:0.96666667,0.85114904);
\draw[draw=black,fill=black,very thick] (axis cs:0.96666667,0.88025302) rectangle (axis cs:1,0.88025302);
\addplot [semithick, gray, dashed, forget plot]
table {%
0 0
1 1
};
\draw (axis cs:0.98,0.02) node[
  scale=1.0,
  anchor=south east,
  text=black,
  rotate=0.0
]{ECE=1.138};
\end{axis}

\end{tikzpicture}

%% file: figures/reliability_plots/argo_gorela.tex
\begin{tikzpicture}

\definecolor{darkgray176}{RGB}{176,176,176}
\definecolor{gray}{RGB}{128,128,128}
\definecolor{lightgray204}{RGB}{204,204,204}
\definecolor{tomato2406060}{RGB}{240,60,60}

\begin{axis}[
legend cell align={left},
legend style={
  fill opacity=0.8,
  draw opacity=1,
  text opacity=1,
  at={(0.03,0.97)},
  anchor=north west,
  draw=lightgray204,
},
tick align=outside,
tick pos=left,
x grid style={darkgray176},
xlabel={\footnotesize{Confidence}},
xmin=0, xmax=1,
xtick style={color=black},
y grid style={darkgray176},
ylabel={\footnotesize{Expected Accuracy}},
ymin=0, ymax=1,
ytick style={color=black},
scale=0.6,
title={\textsc{GoRela}}
]
\draw[draw=tomato2406060,fill=tomato2406060] (axis cs:-3.46944695195361e-18,7.95308889e-05) rectangle (axis cs:0.03333333,0.00634069);
\addlegendimage{ybar,ybar legend,draw=tomato2406060,fill=tomato2406060}
\addlegendentry{Gap}

\draw[draw=tomato2406060,fill=tomato2406060,opacity=0.101441107013025] (axis cs:0.03333333,0.0481502056) rectangle (axis cs:0.06666667,0.14563391);
\draw[draw=tomato2406060,fill=tomato2406060,opacity=0.100696000700119] (axis cs:0.06666667,0.0823408813) rectangle (axis cs:0.1,0.1658524);
\draw[draw=tomato2406060,fill=tomato2406060,opacity=0.1004711297953] (axis cs:0.1,0.115922317) rectangle (axis cs:0.13333333,0.19571821);
\draw[draw=tomato2406060,fill=tomato2406060,opacity=0.100304370658729] (axis cs:0.13333333,0.149466162) rectangle (axis cs:0.16666667,0.18871064);
\draw[draw=tomato2406060,fill=tomato2406060,opacity=0.100202761642613] (axis cs:0.16666667,0.183287304) rectangle (axis cs:0.2,0.20543708);
\draw[draw=tomato2406060,fill=tomato2406060,opacity=0.100121969622508] (axis cs:0.2,0.21421764) rectangle (axis cs:0.23333333,0.216488347);
\draw[draw=tomato2406060,fill=tomato2406060,opacity=0.100089915791391] (axis cs:0.23333333,0.2386891) rectangle (axis cs:0.26666667,0.249761165);
\draw[draw=tomato2406060,fill=tomato2406060,opacity=0.100065964886502] (axis cs:0.26666667,0.24661162) rectangle (axis cs:0.3,0.283317544);
\draw[draw=tomato2406060,fill=tomato2406060,opacity=0.100045688722241] (axis cs:0.3,0.26627665) rectangle (axis cs:0.33333333,0.316756788);
\draw[draw=tomato2406060,fill=tomato2406060,opacity=0.100023775989044] (axis cs:0.33333333,0.27406126) rectangle (axis cs:0.36666667,0.350262603);
\draw[draw=tomato2406060,fill=tomato2406060,opacity=0.100023732762715] (axis cs:0.36666667,0.30677109) rectangle (axis cs:0.4,0.383009264);
\draw[draw=tomato2406060,fill=tomato2406060,opacity=0.100014109878115] (axis cs:0.4,0.28718026) rectangle (axis cs:0.43333333,0.416847046);
\draw[draw=tomato2406060,fill=tomato2406060,opacity=0.100008713623789] (axis cs:0.43333333,0.31088999) rectangle (axis cs:0.46666667,0.449999903);
\draw[draw=tomato2406060,fill=tomato2406060,opacity=0.105323343571499] (axis cs:0.46666667,0.01566851) rectangle (axis cs:0.5,0.499212453);
\draw[draw=tomato2406060,fill=tomato2406060,opacity=0.1] (axis cs:0.5,0.32121729) rectangle (axis cs:0.53333333,0.516399351);
\draw[draw=tomato2406060,fill=tomato2406060,opacity=0.10000181751636] (axis cs:0.53333333,0.3663695) rectangle (axis cs:0.56666667,0.549789872);
\draw[draw=tomato2406060,fill=tomato2406060,opacity=0.100014931681004] (axis cs:0.56666667,0.36775519) rectangle (axis cs:0.6,0.583364265);
\draw[draw=tomato2406060,fill=tomato2406060,opacity=0.100042662376554] (axis cs:0.6,0.41044036) rectangle (axis cs:0.63333333,0.617141246);
\draw[draw=tomato2406060,fill=tomato2406060,opacity=0.100099343654878] (axis cs:0.63333333,0.46054299) rectangle (axis cs:0.66666667,0.651184737);
\draw[draw=tomato2406060,fill=tomato2406060,opacity=0.100592007199045] (axis cs:0.66666667,0.68202698) rectangle (axis cs:0.7,0.686539817);
\draw[draw=tomato2406060,fill=tomato2406060,opacity=0.100725548409397] (axis cs:0.7,0.715923872) rectangle (axis cs:0.73333333,0.72997999);
\draw[draw=tomato2406060,fill=tomato2406060,opacity=0.100451058704788] (axis cs:0.73333333,0.73000007) rectangle (axis cs:0.76666667,0.749005524);
\draw[draw=tomato2406060,fill=tomato2406060,opacity=0.100234829547419] (axis cs:0.76666667,0.72822931) rectangle (axis cs:0.8,0.782606394);
\draw[draw=tomato2406060,fill=tomato2406060,opacity=0.100136649987644] (axis cs:0.8,0.71930809) rectangle (axis cs:0.83333333,0.816453695);
\draw[draw=tomato2406060,fill=tomato2406060,opacity=0.100139784901786] (axis cs:0.83333333,0.72826712) rectangle (axis cs:0.86666667,0.85033752);
\draw[draw=tomato2406060,fill=tomato2406060,opacity=0.100193122673797] (axis cs:0.86666667,0.74631739) rectangle (axis cs:0.9,0.884226688);
\draw[draw=tomato2406060,fill=tomato2406060,opacity=0.100304308835025] (axis cs:0.9,0.78341248) rectangle (axis cs:0.93333333,0.917898585);
\draw[draw=tomato2406060,fill=tomato2406060,opacity=0.100735013970264] (axis cs:0.93333333,0.80972645) rectangle (axis cs:0.96666667,0.95195921);
\draw[draw=tomato2406060,fill=tomato2406060,opacity=0.109070075640056] (axis cs:0.96666667,0.85528151) rectangle (axis cs:1,0.994529691);
\draw[draw=black,fill=black,very thick] (axis cs:-3.46944695195361e-18,0.00634069) rectangle (axis cs:0.03333333,0.00634069);
\addlegendimage{ybar,ybar legend,draw=black,fill=black,very thick}
\addlegendentry{Accuracy}

\draw[draw=black,fill=black,very thick] (axis cs:0.03333333,0.14563391) rectangle (axis cs:0.06666667,0.14563391);
\draw[draw=black,fill=black,very thick] (axis cs:0.06666667,0.1658524) rectangle (axis cs:0.1,0.1658524);
\draw[draw=black,fill=black,very thick] (axis cs:0.1,0.19571821) rectangle (axis cs:0.13333333,0.19571821);
\draw[draw=black,fill=black,very thick] (axis cs:0.13333333,0.18871064) rectangle (axis cs:0.16666667,0.18871064);
\draw[draw=black,fill=black,very thick] (axis cs:0.16666667,0.20543708) rectangle (axis cs:0.2,0.20543708);
\draw[draw=black,fill=black,very thick] (axis cs:0.2,0.21421764) rectangle (axis cs:0.23333333,0.21421764);
\draw[draw=black,fill=black,very thick] (axis cs:0.23333333,0.2386891) rectangle (axis cs:0.26666667,0.2386891);
\draw[draw=black,fill=black,very thick] (axis cs:0.26666667,0.24661162) rectangle (axis cs:0.3,0.24661162);
\draw[draw=black,fill=black,very thick] (axis cs:0.3,0.26627665) rectangle (axis cs:0.33333333,0.26627665);
\draw[draw=black,fill=black,very thick] (axis cs:0.33333333,0.27406126) rectangle (axis cs:0.36666667,0.27406126);
\draw[draw=black,fill=black,very thick] (axis cs:0.36666667,0.30677109) rectangle (axis cs:0.4,0.30677109);
\draw[draw=black,fill=black,very thick] (axis cs:0.4,0.28718026) rectangle (axis cs:0.43333333,0.28718026);
\draw[draw=black,fill=black,very thick] (axis cs:0.43333333,0.31088999) rectangle (axis cs:0.46666667,0.31088999);
\draw[draw=black,fill=black,very thick] (axis cs:0.46666667,0.01566851) rectangle (axis cs:0.5,0.01566851);
\draw[draw=black,fill=black,very thick] (axis cs:0.5,0.32121729) rectangle (axis cs:0.53333333,0.32121729);
\draw[draw=black,fill=black,very thick] (axis cs:0.53333333,0.3663695) rectangle (axis cs:0.56666667,0.3663695);
\draw[draw=black,fill=black,very thick] (axis cs:0.56666667,0.36775519) rectangle (axis cs:0.6,0.36775519);
\draw[draw=black,fill=black,very thick] (axis cs:0.6,0.41044036) rectangle (axis cs:0.63333333,0.41044036);
\draw[draw=black,fill=black,very thick] (axis cs:0.63333333,0.46054299) rectangle (axis cs:0.66666667,0.46054299);
\draw[draw=black,fill=black,very thick] (axis cs:0.66666667,0.68202698) rectangle (axis cs:0.7,0.68202698);
\draw[draw=black,fill=black,very thick] (axis cs:0.7,0.72997999) rectangle (axis cs:0.73333333,0.72997999);
\draw[draw=black,fill=black,very thick] (axis cs:0.73333333,0.73000007) rectangle (axis cs:0.76666667,0.73000007);
\draw[draw=black,fill=black,very thick] (axis cs:0.76666667,0.72822931) rectangle (axis cs:0.8,0.72822931);
\draw[draw=black,fill=black,very thick] (axis cs:0.8,0.71930809) rectangle (axis cs:0.83333333,0.71930809);
\draw[draw=black,fill=black,very thick] (axis cs:0.83333333,0.72826712) rectangle (axis cs:0.86666667,0.72826712);
\draw[draw=black,fill=black,very thick] (axis cs:0.86666667,0.74631739) rectangle (axis cs:0.9,0.74631739);
\draw[draw=black,fill=black,very thick] (axis cs:0.9,0.78341248) rectangle (axis cs:0.93333333,0.78341248);
\draw[draw=black,fill=black,very thick] (axis cs:0.93333333,0.80972645) rectangle (axis cs:0.96666667,0.80972645);
\draw[draw=black,fill=black,very thick] (axis cs:0.96666667,0.85528151) rectangle (axis cs:1,0.85528151);
\addplot [semithick, gray, dashed, forget plot]
table {%
0 0
1 1
};
\draw (axis cs:0.98,0.02) node[
  scale=1.0,
  anchor=south east,
  text=black,
  rotate=0.0
]{ECE=1.161};
\end{axis}

\end{tikzpicture}

%% file: figures/reliability_plots/argo_occ_flow.tex
\begin{tikzpicture}

\definecolor{darkgray176}{RGB}{176,176,176}
\definecolor{gray}{RGB}{128,128,128}
\definecolor{lightgray204}{RGB}{204,204,204}
\definecolor{tomato2406060}{RGB}{240,60,60}

\begin{axis}[
legend cell align={left},
legend style={
  fill opacity=0.8,
  draw opacity=1,
  text opacity=1,
  at={(0.03,0.97)},
  anchor=north west,
  draw=lightgray204
},
tick align=outside,
tick pos=left,
x grid style={darkgray176},
xlabel={\footnotesize{Confidence}},
xmin=0, xmax=1,
xtick style={color=black},
y grid style={darkgray176},
ylabel={\footnotesize{Expected Accuracy}},
ymin=0, ymax=1,
ytick style={color=black},
scale=0.6,
title={\textsc{OccFlow}},
]
\draw[draw=tomato2406060,fill=tomato2406060] (axis cs:-3.46944695195361e-18,0.00262597) rectangle (axis cs:0.03333333,0.00377693);
\addlegendimage{ybar,ybar legend,draw=tomato2406060,fill=tomato2406060}
\addlegendentry{Gap}

\draw[draw=tomato2406060,fill=tomato2406060,opacity=0.126262567498645] (axis cs:0.03333333,0.03903893) rectangle (axis cs:0.06666667,0.04663443);
\draw[draw=tomato2406060,fill=tomato2406060,opacity=0.109879380546959] (axis cs:0.06666667,0.06482199) rectangle (axis cs:0.1,0.08157047);
\draw[draw=tomato2406060,fill=tomato2406060,opacity=0.105504608312366] (axis cs:0.1,0.08995013) rectangle (axis cs:0.13333333,0.11539677);
\draw[draw=tomato2406060,fill=tomato2406060,opacity=0.103497305136456] (axis cs:0.13333333,0.1156981) rectangle (axis cs:0.16666667,0.14908037);
\draw[draw=tomato2406060,fill=tomato2406060,opacity=0.102445003444216] (axis cs:0.16666667,0.14152879) rectangle (axis cs:0.2,0.18254514);
\draw[draw=tomato2406060,fill=tomato2406060,opacity=0.101690068271672] (axis cs:0.2,0.17092388) rectangle (axis cs:0.23333333,0.21589413);
\draw[draw=tomato2406060,fill=tomato2406060,opacity=0.101190856474484] (axis cs:0.23333333,0.2003071) rectangle (axis cs:0.26666667,0.24938546);
\draw[draw=tomato2406060,fill=tomato2406060,opacity=0.100870049892156] (axis cs:0.26666667,0.2286464) rectangle (axis cs:0.3,0.28283528);
\draw[draw=tomato2406060,fill=tomato2406060,opacity=0.100641570871252] (axis cs:0.3,0.25599938) rectangle (axis cs:0.33333333,0.31619773);
\draw[draw=tomato2406060,fill=tomato2406060,opacity=0.100467623381327] (axis cs:0.33333333,0.28572773) rectangle (axis cs:0.36666667,0.34961983);
\draw[draw=tomato2406060,fill=tomato2406060,opacity=0.100336319803446] (axis cs:0.36666667,0.31472362) rectangle (axis cs:0.4,0.38296196);
\draw[draw=tomato2406060,fill=tomato2406060,opacity=0.100239943177118] (axis cs:0.4,0.34527111) rectangle (axis cs:0.43333333,0.4164139);
\draw[draw=tomato2406060,fill=tomato2406060,opacity=0.100167614052459] (axis cs:0.43333333,0.37328498) rectangle (axis cs:0.46666667,0.44975143);
\draw[draw=tomato2406060,fill=tomato2406060,opacity=0.100113513385547] (axis cs:0.46666667,0.40122276) rectangle (axis cs:0.5,0.48315924);
\draw[draw=tomato2406060,fill=tomato2406060,opacity=0.100076810110966] (axis cs:0.5,0.43107582) rectangle (axis cs:0.53333333,0.51652998);
\draw[draw=tomato2406060,fill=tomato2406060,opacity=0.100050077506719] (axis cs:0.53333333,0.46132845) rectangle (axis cs:0.56666667,0.5498908);
\draw[draw=tomato2406060,fill=tomato2406060,opacity=0.100029077244912] (axis cs:0.56666667,0.48891713) rectangle (axis cs:0.6,0.58324498);
\draw[draw=tomato2406060,fill=tomato2406060,opacity=0.100011949473041] (axis cs:0.6,0.51559975) rectangle (axis cs:0.63333333,0.61659023);
\draw[draw=tomato2406060,fill=tomato2406060,opacity=0.1] (axis cs:0.63333333,0.54735462) rectangle (axis cs:0.66666667,0.64998098);
\draw[draw=tomato2406060,fill=tomato2406060,opacity=0.1000054933847] (axis cs:0.66666667,0.58056165) rectangle (axis cs:0.7,0.68337622);
\draw[draw=tomato2406060,fill=tomato2406060,opacity=0.100023947054831] (axis cs:0.7,0.61561336) rectangle (axis cs:0.73333333,0.71679845);
\draw[draw=tomato2406060,fill=tomato2406060,opacity=0.100055928270695] (axis cs:0.73333333,0.65252044) rectangle (axis cs:0.76666667,0.75017294);
\draw[draw=tomato2406060,fill=tomato2406060,opacity=0.100104129007546] (axis cs:0.76666667,0.6910145) rectangle (axis cs:0.8,0.78356624);
\draw[draw=tomato2406060,fill=tomato2406060,opacity=0.100176004006764] (axis cs:0.8,0.72803345) rectangle (axis cs:0.83333333,0.81699297);
\draw[draw=tomato2406060,fill=tomato2406060,opacity=0.100267588261588] (axis cs:0.83333333,0.76777938) rectangle (axis cs:0.86666667,0.85031021);
\draw[draw=tomato2406060,fill=tomato2406060,opacity=0.100402979497105] (axis cs:0.86666667,0.81139448) rectangle (axis cs:0.9,0.88383573);
\draw[draw=tomato2406060,fill=tomato2406060,opacity=0.100636416362463] (axis cs:0.9,0.85772449) rectangle (axis cs:0.93333333,0.91737761);
\draw[draw=tomato2406060,fill=tomato2406060,opacity=0.10119820283901] (axis cs:0.93333333,0.90882174) rectangle (axis cs:0.96666667,0.95144863);
\draw[draw=tomato2406060,fill=tomato2406060,opacity=0.103318596995081] (axis cs:0.96666667,0.97634688) rectangle (axis cs:1,0.98613164);
\draw[draw=black,fill=black,very thick] (axis cs:-3.46944695195361e-18,0.00377693) rectangle (axis cs:0.03333333,0.00377693);
\addlegendimage{ybar,ybar legend,draw=black,fill=black,very thick}
\addlegendentry{Accuracy}

\draw[draw=black,fill=black,very thick] (axis cs:0.03333333,0.03903893) rectangle (axis cs:0.06666667,0.03903893);
\draw[draw=black,fill=black,very thick] (axis cs:0.06666667,0.06482199) rectangle (axis cs:0.1,0.06482199);
\draw[draw=black,fill=black,very thick] (axis cs:0.1,0.08995013) rectangle (axis cs:0.13333333,0.08995013);
\draw[draw=black,fill=black,very thick] (axis cs:0.13333333,0.1156981) rectangle (axis cs:0.16666667,0.1156981);
\draw[draw=black,fill=black,very thick] (axis cs:0.16666667,0.14152879) rectangle (axis cs:0.2,0.14152879);
\draw[draw=black,fill=black,very thick] (axis cs:0.2,0.17092388) rectangle (axis cs:0.23333333,0.17092388);
\draw[draw=black,fill=black,very thick] (axis cs:0.23333333,0.2003071) rectangle (axis cs:0.26666667,0.2003071);
\draw[draw=black,fill=black,very thick] (axis cs:0.26666667,0.2286464) rectangle (axis cs:0.3,0.2286464);
\draw[draw=black,fill=black,very thick] (axis cs:0.3,0.25599938) rectangle (axis cs:0.33333333,0.25599938);
\draw[draw=black,fill=black,very thick] (axis cs:0.33333333,0.28572773) rectangle (axis cs:0.36666667,0.28572773);
\draw[draw=black,fill=black,very thick] (axis cs:0.36666667,0.31472362) rectangle (axis cs:0.4,0.31472362);
\draw[draw=black,fill=black,very thick] (axis cs:0.4,0.34527111) rectangle (axis cs:0.43333333,0.34527111);
\draw[draw=black,fill=black,very thick] (axis cs:0.43333333,0.37328498) rectangle (axis cs:0.46666667,0.37328498);
\draw[draw=black,fill=black,very thick] (axis cs:0.46666667,0.40122276) rectangle (axis cs:0.5,0.40122276);
\draw[draw=black,fill=black,very thick] (axis cs:0.5,0.43107582) rectangle (axis cs:0.53333333,0.43107582);
\draw[draw=black,fill=black,very thick] (axis cs:0.53333333,0.46132845) rectangle (axis cs:0.56666667,0.46132845);
\draw[draw=black,fill=black,very thick] (axis cs:0.56666667,0.48891713) rectangle (axis cs:0.6,0.48891713);
\draw[draw=black,fill=black,very thick] (axis cs:0.6,0.51559975) rectangle (axis cs:0.63333333,0.51559975);
\draw[draw=black,fill=black,very thick] (axis cs:0.63333333,0.54735462) rectangle (axis cs:0.66666667,0.54735462);
\draw[draw=black,fill=black,very thick] (axis cs:0.66666667,0.58056165) rectangle (axis cs:0.7,0.58056165);
\draw[draw=black,fill=black,very thick] (axis cs:0.7,0.61561336) rectangle (axis cs:0.73333333,0.61561336);
\draw[draw=black,fill=black,very thick] (axis cs:0.73333333,0.65252044) rectangle (axis cs:0.76666667,0.65252044);
\draw[draw=black,fill=black,very thick] (axis cs:0.76666667,0.6910145) rectangle (axis cs:0.8,0.6910145);
\draw[draw=black,fill=black,very thick] (axis cs:0.8,0.72803345) rectangle (axis cs:0.83333333,0.72803345);
\draw[draw=black,fill=black,very thick] (axis cs:0.83333333,0.76777938) rectangle (axis cs:0.86666667,0.76777938);
\draw[draw=black,fill=black,very thick] (axis cs:0.86666667,0.81139448) rectangle (axis cs:0.9,0.81139448);
\draw[draw=black,fill=black,very thick] (axis cs:0.9,0.85772449) rectangle (axis cs:0.93333333,0.85772449);
\draw[draw=black,fill=black,very thick] (axis cs:0.93333333,0.90882174) rectangle (axis cs:0.96666667,0.90882174);
\draw[draw=black,fill=black,very thick] (axis cs:0.96666667,0.97634688) rectangle (axis cs:1,0.97634688);
\addplot [semithick, gray, dashed, forget plot]
table {%
0 0
1 1
};
\draw (axis cs:0.98,0.02) node[
  scale=1.0,
  anchor=south east,
  text=black,
  rotate=0.0
]{ECE=0.348};
\end{axis}

\end{tikzpicture}

%% file: figures/reliability_plots/argo_mp3.tex
\begin{tikzpicture}

\definecolor{darkgray176}{RGB}{176,176,176}
\definecolor{gray}{RGB}{128,128,128}
\definecolor{lightgray204}{RGB}{204,204,204}
\definecolor{tomato2406060}{RGB}{240,60,60}

\begin{axis}[
legend cell align={left},
legend style={
  fill opacity=0.8,
  draw opacity=1,
  text opacity=1,
  at={(0.03,0.97)},
  anchor=north west,
  draw=lightgray204
},
tick align=outside,
tick pos=left,
x grid style={darkgray176},
xlabel={\footnotesize{Confidence}},
xmin=0, xmax=1,
xtick style={color=black},
y grid style={darkgray176},
ylabel={\footnotesize{Expected Accuracy}},
ymin=0, ymax=1,
ytick style={color=black},
scale=0.6,
title={\textsc{MP3}},
]
\draw[draw=tomato2406060,fill=tomato2406060] (axis cs:-3.46944695195361e-18,0.00114878) rectangle (axis cs:0.03333333,0.00202443);
\addlegendimage{ybar,ybar legend,draw=tomato2406060,fill=tomato2406060}
\addlegendentry{Gap}

\draw[draw=tomato2406060,fill=tomato2406060,opacity=0.112337408616746] (axis cs:0.03333333,0.04759564) rectangle (axis cs:0.06666667,0.05823061);
\draw[draw=tomato2406060,fill=tomato2406060,opacity=0.105998321191503] (axis cs:0.06666667,0.08193905) rectangle (axis cs:0.1,0.09815204);
\draw[draw=tomato2406060,fill=tomato2406060,opacity=0.103705158078378] (axis cs:0.1,0.11566328) rectangle (axis cs:0.13333333,0.13719403);
\draw[draw=tomato2406060,fill=tomato2406060,opacity=0.102500435734927] (axis cs:0.13333333,0.14917011) rectangle (axis cs:0.16666667,0.17577803);
\draw[draw=tomato2406060,fill=tomato2406060,opacity=0.101747150205726] (axis cs:0.16666667,0.18263529) rectangle (axis cs:0.2,0.21171193);
\draw[draw=tomato2406060,fill=tomato2406060,opacity=0.10126246522235] (axis cs:0.2,0.21607467) rectangle (axis cs:0.23333333,0.24718025);
\draw[draw=tomato2406060,fill=tomato2406060,opacity=0.100931721312249] (axis cs:0.23333333,0.24950091) rectangle (axis cs:0.26666667,0.28208594);
\draw[draw=tomato2406060,fill=tomato2406060,opacity=0.100702662130577] (axis cs:0.26666667,0.28292831) rectangle (axis cs:0.3,0.31892765);
\draw[draw=tomato2406060,fill=tomato2406060,opacity=0.100535088899981] (axis cs:0.3,0.31630052) rectangle (axis cs:0.33333333,0.35578338);
\draw[draw=tomato2406060,fill=tomato2406060,opacity=0.10040305884579] (axis cs:0.33333333,0.34968495) rectangle (axis cs:0.36666667,0.39197403);
\draw[draw=tomato2406060,fill=tomato2406060,opacity=0.100302346551813] (axis cs:0.36666667,0.3830658) rectangle (axis cs:0.4,0.42737753);
\draw[draw=tomato2406060,fill=tomato2406060,opacity=0.10022197171995] (axis cs:0.4,0.41642072) rectangle (axis cs:0.43333333,0.46256584);
\draw[draw=tomato2406060,fill=tomato2406060,opacity=0.10015585908791] (axis cs:0.43333333,0.44980753) rectangle (axis cs:0.46666667,0.49912101);
\draw[draw=tomato2406060,fill=tomato2406060,opacity=0.100110122650104] (axis cs:0.46666667,0.48319339) rectangle (axis cs:0.5,0.5332863);
\draw[draw=tomato2406060,fill=tomato2406060,opacity=0.100075912375091] (axis cs:0.5,0.51652719) rectangle (axis cs:0.53333333,0.56833525);
\draw[draw=tomato2406060,fill=tomato2406060,opacity=0.100049013175245] (axis cs:0.53333333,0.54990273) rectangle (axis cs:0.56666667,0.60216177);
\draw[draw=tomato2406060,fill=tomato2406060,opacity=0.100030034952365] (axis cs:0.56666667,0.58326058) rectangle (axis cs:0.6,0.63365485);
\draw[draw=tomato2406060,fill=tomato2406060,opacity=0.100014950082292] (axis cs:0.6,0.61660147) rectangle (axis cs:0.63333333,0.66404522);
\draw[draw=tomato2406060,fill=tomato2406060,opacity=0.100004670060941] (axis cs:0.63333333,0.64996553) rectangle (axis cs:0.66666667,0.69178331);
\draw[draw=tomato2406060,fill=tomato2406060,opacity=0.1] (axis cs:0.66666667,0.68333719) rectangle (axis cs:0.7,0.71679316);
\draw[draw=tomato2406060,fill=tomato2406060,opacity=0.100007234643781] (axis cs:0.7,0.71672638) rectangle (axis cs:0.73333333,0.74203804);
\draw[draw=tomato2406060,fill=tomato2406060,opacity=0.100026797463475] (axis cs:0.73333333,0.75010426) rectangle (axis cs:0.76666667,0.76615841);
\draw[draw=tomato2406060,fill=tomato2406060,opacity=0.100062731576088] (axis cs:0.76666667,0.78353512) rectangle (axis cs:0.8,0.79166786);
\draw[draw=tomato2406060,fill=tomato2406060,opacity=0.10012454287698] (axis cs:0.8,0.81695942) rectangle (axis cs:0.83333333,0.81817617);
\draw[draw=tomato2406060,fill=tomato2406060,opacity=0.100229061763813] (axis cs:0.83333333,0.84613682) rectangle (axis cs:0.86666667,0.85047645);
\draw[draw=tomato2406060,fill=tomato2406060,opacity=0.100441695427595] (axis cs:0.86666667,0.88029274) rectangle (axis cs:0.9,0.88413208);
\draw[draw=tomato2406060,fill=tomato2406060,opacity=0.100961051856439] (axis cs:0.9,0.91810989) rectangle (axis cs:0.93333333,0.92348337);
\draw[draw=tomato2406060,fill=tomato2406060,opacity=0.101393570869919] (axis cs:0.93333333,0.94976212) rectangle (axis cs:0.96666667,0.95799198);
\draw[draw=tomato2406060,fill=tomato2406060,opacity=0.101830724827215] (axis cs:0.96666667,0.9839845) rectangle (axis cs:1,0.98633127);
\draw[draw=black,fill=black,very thick] (axis cs:-3.46944695195361e-18,0.00202443) rectangle (axis cs:0.03333333,0.00202443);
\addlegendimage{ybar,ybar legend,draw=black,fill=black,very thick}
\addlegendentry{Accuracy}

\draw[draw=black,fill=black,very thick] (axis cs:0.03333333,0.05823061) rectangle (axis cs:0.06666667,0.05823061);
\draw[draw=black,fill=black,very thick] (axis cs:0.06666667,0.09815204) rectangle (axis cs:0.1,0.09815204);
\draw[draw=black,fill=black,very thick] (axis cs:0.1,0.13719403) rectangle (axis cs:0.13333333,0.13719403);
\draw[draw=black,fill=black,very thick] (axis cs:0.13333333,0.17577803) rectangle (axis cs:0.16666667,0.17577803);
\draw[draw=black,fill=black,very thick] (axis cs:0.16666667,0.21171193) rectangle (axis cs:0.2,0.21171193);
\draw[draw=black,fill=black,very thick] (axis cs:0.2,0.24718025) rectangle (axis cs:0.23333333,0.24718025);
\draw[draw=black,fill=black,very thick] (axis cs:0.23333333,0.28208594) rectangle (axis cs:0.26666667,0.28208594);
\draw[draw=black,fill=black,very thick] (axis cs:0.26666667,0.31892765) rectangle (axis cs:0.3,0.31892765);
\draw[draw=black,fill=black,very thick] (axis cs:0.3,0.35578338) rectangle (axis cs:0.33333333,0.35578338);
\draw[draw=black,fill=black,very thick] (axis cs:0.33333333,0.39197403) rectangle (axis cs:0.36666667,0.39197403);
\draw[draw=black,fill=black,very thick] (axis cs:0.36666667,0.42737753) rectangle (axis cs:0.4,0.42737753);
\draw[draw=black,fill=black,very thick] (axis cs:0.4,0.46256584) rectangle (axis cs:0.43333333,0.46256584);
\draw[draw=black,fill=black,very thick] (axis cs:0.43333333,0.49912101) rectangle (axis cs:0.46666667,0.49912101);
\draw[draw=black,fill=black,very thick] (axis cs:0.46666667,0.5332863) rectangle (axis cs:0.5,0.5332863);
\draw[draw=black,fill=black,very thick] (axis cs:0.5,0.56833525) rectangle (axis cs:0.53333333,0.56833525);
\draw[draw=black,fill=black,very thick] (axis cs:0.53333333,0.60216177) rectangle (axis cs:0.56666667,0.60216177);
\draw[draw=black,fill=black,very thick] (axis cs:0.56666667,0.63365485) rectangle (axis cs:0.6,0.63365485);
\draw[draw=black,fill=black,very thick] (axis cs:0.6,0.66404522) rectangle (axis cs:0.63333333,0.66404522);
\draw[draw=black,fill=black,very thick] (axis cs:0.63333333,0.69178331) rectangle (axis cs:0.66666667,0.69178331);
\draw[draw=black,fill=black,very thick] (axis cs:0.66666667,0.71679316) rectangle (axis cs:0.7,0.71679316);
\draw[draw=black,fill=black,very thick] (axis cs:0.7,0.74203804) rectangle (axis cs:0.73333333,0.74203804);
\draw[draw=black,fill=black,very thick] (axis cs:0.73333333,0.76615841) rectangle (axis cs:0.76666667,0.76615841);
\draw[draw=black,fill=black,very thick] (axis cs:0.76666667,0.79166786) rectangle (axis cs:0.8,0.79166786);
\draw[draw=black,fill=black,very thick] (axis cs:0.8,0.81817617) rectangle (axis cs:0.83333333,0.81817617);
\draw[draw=black,fill=black,very thick] (axis cs:0.83333333,0.84613682) rectangle (axis cs:0.86666667,0.84613682);
\draw[draw=black,fill=black,very thick] (axis cs:0.86666667,0.88029274) rectangle (axis cs:0.9,0.88029274);
\draw[draw=black,fill=black,very thick] (axis cs:0.9,0.92348337) rectangle (axis cs:0.93333333,0.92348337);
\draw[draw=black,fill=black,very thick] (axis cs:0.93333333,0.95799198) rectangle (axis cs:0.96666667,0.95799198);
\draw[draw=black,fill=black,very thick] (axis cs:0.96666667,0.9839845) rectangle (axis cs:1,0.9839845);
\addplot [semithick, gray, dashed, forget plot]
table {%
0 0
1 1
};
\draw (axis cs:0.98,0.02) node[
  scale=1.0,
  anchor=south east,
  text=black,
  rotate=0.0
]{ECE=0.201};
\end{axis}

\end{tikzpicture}

%% file: figures/reliability_plots/argo_implicitO.tex
\begin{tikzpicture}

\definecolor{darkgray176}{RGB}{176,176,176}
\definecolor{gray}{RGB}{128,128,128}
\definecolor{lightgray204}{RGB}{204,204,204}
\definecolor{tomato2406060}{RGB}{240,60,60}

\begin{axis}[
legend cell align={left},
legend style={
  fill opacity=0.8,
  draw opacity=1,
  text opacity=1,
  at={(0.03,0.97)},
  anchor=north west,
  draw=lightgray204
},
tick align=outside,
tick pos=left,
x grid style={darkgray176},
xlabel={\footnotesize{Confidence}},
xmin=0, xmax=1,
xtick style={color=black},
y grid style={darkgray176},
ylabel={\footnotesize{Expected Accuracy}},
ymin=0, ymax=1,
ytick style={color=black},
scale=0.6,
title={\textsc{ImplicitO}},
]
\draw[draw=tomato2406060,fill=tomato2406060] (axis cs:-3.46944695195361e-18,0.000976049109) rectangle (axis cs:0.03333333,0.00205473);
\addlegendimage{ybar,ybar legend,draw=tomato2406060,fill=tomato2406060}
\addlegendentry{Gap}

\draw[draw=tomato2406060,fill=tomato2406060,opacity=0.110952888713037] (axis cs:0.03333333,0.0475311981) rectangle (axis cs:0.06666667,0.05695783);
\draw[draw=tomato2406060,fill=tomato2406060,opacity=0.105196953920218] (axis cs:0.06666667,0.0819221624) rectangle (axis cs:0.1,0.08951416);
\draw[draw=tomato2406060,fill=tomato2406060,opacity=0.103192629035567] (axis cs:0.1,0.115677164) rectangle (axis cs:0.13333333,0.1205137);
\draw[draw=tomato2406060,fill=tomato2406060,opacity=0.102174524024603] (axis cs:0.13333333,0.149232357) rectangle (axis cs:0.16666667,0.15162872);
\draw[draw=tomato2406060,fill=tomato2406060,opacity=0.101572132062443] (axis cs:0.16666667,0.18033952) rectangle (axis cs:0.2,0.182699133);
\draw[draw=tomato2406060,fill=tomato2406060,opacity=0.101172682978034] (axis cs:0.2,0.20818779) rectangle (axis cs:0.23333333,0.216145398);
\draw[draw=tomato2406060,fill=tomato2406060,opacity=0.100899728361355] (axis cs:0.23333333,0.23639722) rectangle (axis cs:0.26666667,0.249546626);
\draw[draw=tomato2406060,fill=tomato2406060,opacity=0.10069251926398] (axis cs:0.26666667,0.26499618) rectangle (axis cs:0.3,0.282936877);
\draw[draw=tomato2406060,fill=tomato2406060,opacity=0.100535003925123] (axis cs:0.3,0.29396766) rectangle (axis cs:0.33333333,0.31631503);
\draw[draw=tomato2406060,fill=tomato2406060,opacity=0.100411177969269] (axis cs:0.33333333,0.32412103) rectangle (axis cs:0.36666667,0.349683154);
\draw[draw=tomato2406060,fill=tomato2406060,opacity=0.100314606270498] (axis cs:0.36666667,0.35236143) rectangle (axis cs:0.4,0.383054698);
\draw[draw=tomato2406060,fill=tomato2406060,opacity=0.100236185706073] (axis cs:0.4,0.38094654) rectangle (axis cs:0.43333333,0.41643679);
\draw[draw=tomato2406060,fill=tomato2406060,opacity=0.100174666231924] (axis cs:0.43333333,0.40975701) rectangle (axis cs:0.46666667,0.449793239);
\draw[draw=tomato2406060,fill=tomato2406060,opacity=0.100124316290405] (axis cs:0.46666667,0.44084768) rectangle (axis cs:0.5,0.483143626);
\draw[draw=tomato2406060,fill=tomato2406060,opacity=0.100084456520936] (axis cs:0.5,0.47356516) rectangle (axis cs:0.53333333,0.516511611);
\draw[draw=tomato2406060,fill=tomato2406060,opacity=0.100055662757579] (axis cs:0.53333333,0.50354074) rectangle (axis cs:0.56666667,0.549870679);
\draw[draw=tomato2406060,fill=tomato2406060,opacity=0.100031825905688] (axis cs:0.56666667,0.53580265) rectangle (axis cs:0.6,0.583224607);
\draw[draw=tomato2406060,fill=tomato2406060,opacity=0.100013206069198] (axis cs:0.6,0.56881659) rectangle (axis cs:0.63333333,0.616602239);
\draw[draw=tomato2406060,fill=tomato2406060,opacity=0.100003616732368] (axis cs:0.63333333,0.60290959) rectangle (axis cs:0.66666667,0.649965629);
\draw[draw=tomato2406060,fill=tomato2406060,opacity=0.1] (axis cs:0.66666667,0.63778113) rectangle (axis cs:0.7,0.683340378);
\draw[draw=tomato2406060,fill=tomato2406060,opacity=0.100005124305088] (axis cs:0.7,0.66979915) rectangle (axis cs:0.73333333,0.716704975);
\draw[draw=tomato2406060,fill=tomato2406060,opacity=0.100019618532596] (axis cs:0.73333333,0.70386712) rectangle (axis cs:0.76666667,0.750119597);
\draw[draw=tomato2406060,fill=tomato2406060,opacity=0.100045577935625] (axis cs:0.76666667,0.73653656) rectangle (axis cs:0.8,0.783483344);
\draw[draw=tomato2406060,fill=tomato2406060,opacity=0.100089545802127] (axis cs:0.8,0.77368801) rectangle (axis cs:0.83333333,0.816891182);
\draw[draw=tomato2406060,fill=tomato2406060,opacity=0.100149674106506] (axis cs:0.83333333,0.81034971) rectangle (axis cs:0.86666667,0.850294074);
\draw[draw=tomato2406060,fill=tomato2406060,opacity=0.100229351410746] (axis cs:0.86666667,0.84866534) rectangle (axis cs:0.9,0.883655947);
\draw[draw=tomato2406060,fill=tomato2406060,opacity=0.100371972837629] (axis cs:0.9,0.88224244) rectangle (axis cs:0.93333333,0.917383016);
\draw[draw=tomato2406060,fill=tomato2406060,opacity=0.100867768025711] (axis cs:0.93333333,0.92116608) rectangle (axis cs:0.96666667,0.951701924);
\draw[draw=tomato2406060,fill=tomato2406060,opacity=0.105977297149768] (axis cs:0.96666667,0.97997054) rectangle (axis cs:1,0.99095892);
\draw[draw=black,fill=black,very thick] (axis cs:-3.46944695195361e-18,0.00205473) rectangle (axis cs:0.03333333,0.00205473);
\addlegendimage{ybar,ybar legend,draw=black,fill=black,very thick}
\addlegendentry{Accuracy}

\draw[draw=black,fill=black,very thick] (axis cs:0.03333333,0.05695783) rectangle (axis cs:0.06666667,0.05695783);
\draw[draw=black,fill=black,very thick] (axis cs:0.06666667,0.08951416) rectangle (axis cs:0.1,0.08951416);
\draw[draw=black,fill=black,very thick] (axis cs:0.1,0.1205137) rectangle (axis cs:0.13333333,0.1205137);
\draw[draw=black,fill=black,very thick] (axis cs:0.13333333,0.15162872) rectangle (axis cs:0.16666667,0.15162872);
\draw[draw=black,fill=black,very thick] (axis cs:0.16666667,0.18033952) rectangle (axis cs:0.2,0.18033952);
\draw[draw=black,fill=black,very thick] (axis cs:0.2,0.20818779) rectangle (axis cs:0.23333333,0.20818779);
\draw[draw=black,fill=black,very thick] (axis cs:0.23333333,0.23639722) rectangle (axis cs:0.26666667,0.23639722);
\draw[draw=black,fill=black,very thick] (axis cs:0.26666667,0.26499618) rectangle (axis cs:0.3,0.26499618);
\draw[draw=black,fill=black,very thick] (axis cs:0.3,0.29396766) rectangle (axis cs:0.33333333,0.29396766);
\draw[draw=black,fill=black,very thick] (axis cs:0.33333333,0.32412103) rectangle (axis cs:0.36666667,0.32412103);
\draw[draw=black,fill=black,very thick] (axis cs:0.36666667,0.35236143) rectangle (axis cs:0.4,0.35236143);
\draw[draw=black,fill=black,very thick] (axis cs:0.4,0.38094654) rectangle (axis cs:0.43333333,0.38094654);
\draw[draw=black,fill=black,very thick] (axis cs:0.43333333,0.40975701) rectangle (axis cs:0.46666667,0.40975701);
\draw[draw=black,fill=black,very thick] (axis cs:0.46666667,0.44084768) rectangle (axis cs:0.5,0.44084768);
\draw[draw=black,fill=black,very thick] (axis cs:0.5,0.47356516) rectangle (axis cs:0.53333333,0.47356516);
\draw[draw=black,fill=black,very thick] (axis cs:0.53333333,0.50354074) rectangle (axis cs:0.56666667,0.50354074);
\draw[draw=black,fill=black,very thick] (axis cs:0.56666667,0.53580265) rectangle (axis cs:0.6,0.53580265);
\draw[draw=black,fill=black,very thick] (axis cs:0.6,0.56881659) rectangle (axis cs:0.63333333,0.56881659);
\draw[draw=black,fill=black,very thick] (axis cs:0.63333333,0.60290959) rectangle (axis cs:0.66666667,0.60290959);
\draw[draw=black,fill=black,very thick] (axis cs:0.66666667,0.63778113) rectangle (axis cs:0.7,0.63778113);
\draw[draw=black,fill=black,very thick] (axis cs:0.7,0.66979915) rectangle (axis cs:0.73333333,0.66979915);
\draw[draw=black,fill=black,very thick] (axis cs:0.73333333,0.70386712) rectangle (axis cs:0.76666667,0.70386712);
\draw[draw=black,fill=black,very thick] (axis cs:0.76666667,0.73653656) rectangle (axis cs:0.8,0.73653656);
\draw[draw=black,fill=black,very thick] (axis cs:0.8,0.77368801) rectangle (axis cs:0.83333333,0.77368801);
\draw[draw=black,fill=black,very thick] (axis cs:0.83333333,0.81034971) rectangle (axis cs:0.86666667,0.81034971);
\draw[draw=black,fill=black,very thick] (axis cs:0.86666667,0.84866534) rectangle (axis cs:0.9,0.84866534);
\draw[draw=black,fill=black,very thick] (axis cs:0.9,0.88224244) rectangle (axis cs:0.93333333,0.88224244);
\draw[draw=black,fill=black,very thick] (axis cs:0.93333333,0.92116608) rectangle (axis cs:0.96666667,0.92116608);
\draw[draw=black,fill=black,very thick] (axis cs:0.96666667,0.97997054) rectangle (axis cs:1,0.97997054);
\addplot [semithick, gray, dashed, forget plot]
table {%
0 0
1 1
};
\draw (axis cs:0.98,0.02) node[
  scale=1.0,
  anchor=south east,
  text=black,
  rotate=0.0
]{ECE=0.193};
\end{axis}

\end{tikzpicture}

%% file: main.bbl
\begin{thebibliography}{10}\itemsep=-1pt

\bibitem{carion2020end}
Nicolas Carion, Francisco Massa, Gabriel Synnaeve, Nicolas Usunier, Alexander
  Kirillov, and Sergey Zagoruyko.
\newblock End-to-end object detection with transformers.
\newblock In {\em European conference on computer vision}, pages 213--229.
  Springer, 2020.

\bibitem{casas2019spatially}
Sergio Casas, Cole Gulino, Renjie Liao, and Raquel Urtasun.
\newblock Spatially-aware graph neural networks for relational behavior
  forecasting from sensor data.
\newblock {\em arXiv preprint}, 2019.

\bibitem{casas2020implicit}
Sergio Casas, Cole Gulino, Simon Suo, Katie Luo, Renjie Liao, and Raquel
  Urtasun.
\newblock Implicit latent variable model for scene-consistent motion
  forecasting.
\newblock In {\em ECCV}, 2020.

\bibitem{casas2018intentnet}
Sergio Casas, Wenjie Luo, and Raquel Urtasun.
\newblock Intentnet: Learning to predict intention from raw sensor data.
\newblock In {\em CoRL}, 2018.

\bibitem{casas2021mp3}
Sergio Casas, Abbas Sadat, and Raquel Urtasun.
\newblock Mp3: A unified model to map, perceive, predict and plan.
\newblock In {\em CVPR}, 2021.

\bibitem{chai2019multipath}
Yuning Chai, Benjamin Sapp, Mayank Bansal, and Dragomir Anguelov.
\newblock Multipath: Multiple probabilistic anchor trajectory hypotheses for
  behavior prediction.
\newblock In {\em CoRL}, 2019.

\bibitem{chen2019learning}
Zhiqin Chen and Hao Zhang.
\newblock Learning implicit fields for generative shape modeling.
\newblock In {\em CVPR}, 2019.

\bibitem{cui2021lookout}
Alexander Cui, Sergio Casas, Abbas Sadat, Renjie Liao, and Raquel Urtasun.
\newblock Lookout: Diverse multi-future prediction and planning for
  self-driving.
\newblock In {\em ICCV}, 2021.

\bibitem{cui2022gorela}
Alexander Cui, Sergio Casas, Kelvin Wong, Simon Suo, and Raquel Urtasun.
\newblock Gorela: Go relative for viewpoint-invariant motion forecasting.
\newblock {\em arXiv preprint arXiv:2211.02545}, 2022.

\bibitem{dai2017deformable}
Jifeng Dai, Haozhi Qi, Yuwen Xiong, Yi Li, Guodong Zhang, Han Hu, and Yichen
  Wei.
\newblock Deformable convolutional networks.
\newblock In {\em ICCV}, 2017.

\bibitem{gu2021densetnt}
Junru Gu, Chen Sun, and Hang Zhao.
\newblock Densetnt: End-to-end trajectory prediction from dense goal sets.
\newblock In {\em ICCV}, 2021.

\bibitem{guo2017on}
Chuan Guo, Geoff Pleiss, Yu Sun, and Kilian~Q Weinberger.
\newblock On calibration of modern neural networks.
\newblock In {\em ICML}, 2017.

\bibitem{resnet}
Kaiming He, Xiangyu Zang, Shauqing Ren, and Jian Sun.
\newblock Deep residual learning for image recognition.
\newblock In {\em CVPR}, 2016.

\bibitem{hu2021fiery}
Anthony Hu, Zak Murez, Nikhil Mohan, Sof{\'\i}a Dudas, Jeffrey Hawke, Vijay
  Badrinarayanan, Roberto Cipolla, and Alex Kendall.
\newblock Fiery: Future instance prediction in bird's-eye view from surround
  monocular cameras.
\newblock In {\em ICCV}, 2021.

\bibitem{hu2022st}
Shengchao Hu, Li Chen, Penghao Wu, Hongyang Li, Junchi Yan, and Dacheng Tao.
\newblock St-p3: End-to-end vision-based autonomous driving via
  spatial-temporal feature learning.
\newblock In {\em ECCV}, 2022.

\bibitem{ivanovic2020multimodal}
Boris Ivanovic, Karen Leung, Edward Schmerling, and Marco Pavone.
\newblock Multimodal deep generative models for trajectory prediction: A
  conditional variational autoencoder approach.
\newblock {\em RA-L}, 2020.

\bibitem{ivanovic2018the}
Boris Ivanovic and Marco Pavone.
\newblock The trajectron: Probabilistic multi-agent trajectory modeling with
  dynamic spatiotemporal graphs.
\newblock In {\em ICCV}, 2019.

\bibitem{kim2022stopnet}
Jinkyu Kim, Reza Mahjourian, Scott Ettinger, Mayank Bansal, Brandyn White, Ben
  Sapp, and Dragomir Anguelov.
\newblock Stopnet: Scalable trajectory and occupancy prediction for urban
  autonomous driving.
\newblock In {\em ICRA}, 2022.

\bibitem{lang2019pointpillars}
Alex~H Lang, Sourabh Vora, Holger Caesar, Lubing Zhou, Jiong Yang, and Oscar
  Beijbom.
\newblock Pointpillars: Fast encoders for object detection from point clouds.
\newblock In {\em CVPR}, 2019.

\bibitem{CTRA}
St{\'e}phanie Lef{\`e}vre, Dizan Vasquez, and Christian Laugier.
\newblock A survey on motion prediction and risk assessment for intelligent
  vehicles.
\newblock {\em ROBOMECH}, 2014.

\bibitem{lanegcn}
Ming Liang, Bin Yang, Rui Hu, Yun Chen, Renjie Liao, Song Feng, and Raquel
  Urtasun.
\newblock Learning lane graph representations for motion forecasting.
\newblock In {\em ECCV}, 2020.

\bibitem{Liang_2020_CVPR}
Ming Liang, Bin Yang, Wenyuan Zeng, Yun Chen, Rui Hu, Sergio Casas, and Raquel
  Urtasun.
\newblock Pnpnet: End-to-end perception and prediction with tracking in the
  loop.
\newblock In {\em CVPR}, 2020.

\bibitem{lin2017feature}
Tsung-Yi Lin, Piotr Doll{\'a}r, Ross Girshick, Kaiming He, Bharath Hariharan,
  and Serge Belongie.
\newblock Feature pyramid networks for object detection.
\newblock In {\em CVPR}, 2017.

\bibitem{loshchilov2017decoupled}
Ilya Loshchilov and Frank Hutter.
\newblock Decoupled weight decay regularization.
\newblock {\em arXiv preprint arXiv:1711.05101}, 2017.

\bibitem{luo2021safety}
Katie Luo, Sergio Casas, Renjie Liao, Xinchen Yan, Yuwen Xiong, Wenyuan Zeng,
  and Raquel Urtasun.
\newblock Safety-oriented pedestrian motion and scene occupancy forecasting.
\newblock 2021.

\bibitem{luo2018faf}
Wenjie Luo, Bin Yang, and Raquel Urtasun.
\newblock Fast and furious: Real time end-to-end 3d detection, tracking and
  motion forecasting with a single convolutional net.
\newblock In {\em CVPR}, 2018.

\bibitem{mahjourian2022occupancy}
Reza Mahjourian, Jinkyu Kim, Yuning Chai, Mingxing Tan, Ben Sapp, and Dragomir
  Anguelov.
\newblock Occupancy flow fields for motion forecasting in autonomous driving.
\newblock {\em RA-L}, 2022.

\bibitem{mescheder2019occupancy}
Lars Mescheder, Michael Oechsle, Michael Niemeyer, Sebastian Nowozin, and
  Andreas Geiger.
\newblock Occupancy networks: Learning 3d reconstruction in function space.
\newblock In {\em CVPR}, 2019.

\bibitem{Park2019deepsdf}
Jeong~Joon Park, Peter Florence, Julian Straub, Richard Newcombe, and Steven
  Lovegrove.
\newblock Deepsdf: Learning continuous signed distance functions for shape
  representation.
\newblock In {\em CVPR}, 2019.

\bibitem{peng2020convolutional}
Songyou Peng, Michael Niemeyer, Lars Mescheder, Marc Pollefeys, and Andreas
  Geiger.
\newblock Convolutional occupancy networks.
\newblock In {\em ECCV}, 2020.

\bibitem{phan2020covernet}
Tung Phan-Minh, Elena~Corina Grigore, Freddy~A Boulton, Oscar Beijbom, and
  Eric~M Wolff.
\newblock Covernet: Multimodal behavior prediction using trajectory sets.
\newblock In {\em CVPR}, 2020.

\bibitem{philion2020lift}
Jonah Philion and Sanja Fidler.
\newblock Lift, splat, shoot: Encoding images from arbitrary camera rigs by
  implicitly unprojecting to 3d.
\newblock In {\em ECCV}, 2020.

\bibitem{sadat2020perceive}
Abbas Sadat, Sergio Casas, Mengye Ren, Xinyu Wu, Pranaab Dhawan, and Raquel
  Urtasun.
\newblock Perceive, predict, and plan: Safe motion planning through
  interpretable semantic representations.
\newblock In {\em ECCV}, 2020.

\bibitem{sadat2019joint}
Abbas Sadat, Mengye Ren, Andrei Pokrovsky, Yen-Chen Lin, Ersin Yumer, and
  Raquel Urtasun.
\newblock Jointly learnable behavior and trajectory planning for self-driving
  vehicles.
\newblock In {\em IROS}, 2018.

\bibitem{salzmann2020trajectron}
Tim Salzmann, Boris Ivanovic, Punarjay Chakravarty, and Marco Pavone.
\newblock Trajectron++: Dynamically-feasible trajectory forecasting with
  heterogeneous data.
\newblock In {\em ECCV}, 2020.

\bibitem{sharma2018beyond}
Sarthak Sharma, Junaid~Ahmed Ansari, J~Krishna Murthy, and K~Madhava Krishna.
\newblock Beyond pixels: Leveraging geometry and shape cues for online
  multi-object tracking.
\newblock In {\em ICRA}, 2018.

\bibitem{waymo_open_dataset}
Pei Sun, Henrik Kretzschmar, Xerxes Dotiwalla, Aurelien Chouard, Vijaysai
  Patnaik, Paul Tsui, James Guo, Yin Zhou, Yuning Chai, Benjamin Caine, et~al.
\newblock Scalability in perception for autonomous driving: Waymo open dataset.
\newblock In {\em CVPR}, 2020.

\bibitem{tang2019multiple}
Charlie Tang and Russ~R Salakhutdinov.
\newblock Multiple futures prediction.
\newblock In {\em Advances in Neural Information Processing Systems}, 2019.

\bibitem{trabelsi2021drowned}
Ameni Trabelsi, Ross~J Beveridge, and Nathaniel Blanchard.
\newblock Drowned out by the noise: Evidence for tracking-free motion
  prediction.
\newblock {\em arXiv preprint arXiv:2104.08368}, 2021.

\bibitem{Vaswani2017Attention}
Ashish Vaswani, Noam Shazeer, Niki Parmar, Jakob Uszkoreit, Llion Jones,
  Aidan~N Gomez, {\L}ukasz Kaiser, and Illia Polosukhin.
\newblock Attention is all you need.
\newblock {\em NeurIPS}, 2017.

\bibitem{weng20203d}
Xinshuo Weng, Jianren Wang, David Held, and Kris Kitani.
\newblock 3d multi-object tracking: A baseline and new evaluation metrics.
\newblock In {\em IROS}, 2020.

\bibitem{weng2020ptp}
Xinshuo Weng, Ye Yuan, and Kris Kitani.
\newblock Ptp: Parallelized tracking and prediction with graph neural networks
  and diversity sampling.
\newblock {\em RA-L}, 2021.

\bibitem{Argoverse2}
Benjamin Wilson, William Qi, Tanmay Agarwal, John Lambert, Jagjeet Singh,
  Siddhesh Khandelwal, Bowen Pan, Ratnesh Kumar, Andrew Hartnett,
  Jhony~Kaesemodel Pontes, Deva Ramanan, Peter Carr, and James Hays.
\newblock Argoverse 2: Next generation datasets for self-driving perception and
  forecasting.
\newblock In {\em NeurIPS}, 2021.

\bibitem{wu2019detectron2}
Yuxin Wu, Alexander Kirillov, Francisco Massa, Wan-Yen Lo, and Ross Girshick.
\newblock Detectron2.
\newblock \url{https://github.com/facebookresearch/detectron2}, 2019.

\bibitem{yang2018hdnet}
Bin Yang, Ming Liang, and Raquel Urtasun.
\newblock Hdnet: Exploiting hd maps for 3d object detection.
\newblock In {\em CoRL}, 2018.

\bibitem{yang2018pixor}
Bin Yang, Wenjie Luo, and Raquel Urtasun.
\newblock Pixor: Real-time 3d object detection from point clouds.
\newblock In {\em CVPR}, 2018.

\bibitem{zeng2019end}
Wenyuan Zeng, Wenjie Luo, Simon Suo, Abbas Sadat, Bin Yang, Sergio Casas, and
  Raquel Urtasun.
\newblock End-to-end interpretable neural motion planner.
\newblock In {\em CVPR}, 2019.

\bibitem{zhao2019multi}
Tianyang Zhao, Yifei Xu, Mathew Monfort, Wongun Choi, Chris Baker, Yibiao Zhao,
  Yizhou Wang, and Ying~Nian Wu.
\newblock Multi-agent tensor fusion for contextual trajectory prediction.
\newblock In {\em CVPR}, 2019.

\bibitem{zhu2020deformable}
Xizhou Zhu, Weijie Su, Lewei Lu, Bin Li, Xiaogang Wang, and Jifeng Dai.
\newblock Deformable detr: Deformable transformers for end-to-end object
  detection.
\newblock {\em arXiv}, 2020.

\end{thebibliography}
